\documentclass{bmvc2k}

\usepackage{amsmath}
\usepackage{amssymb}
\usepackage{graphics} 
\usepackage{multicol,multirow}
\usepackage{subfigure}
\usepackage{multirow}

\graphicspath{{./figs/}}
\DeclareGraphicsExtensions{.pdf,.jpg,.png}

\newcommand{\figref}[1]{Fig.~\ref{#1}}
\newcommand{\tabref}[1]{Table~\ref{#1}}

\title{Dual Pyramid Generative Adversarial Networks for Semantic Image Synthesis}

\addauthor{Shijie Li}{lsj@uni-bonn.de}{1}
\addauthor{Ming-Ming Cheng}{cmm@nankai.edu.cn}{2}
\addauthor{Juergen Gall}{gall@iai.uni-bonn.de}{1}

\addinstitution{
 Computer Vision Group\\
 University of Bonn\\
 Bonn, Germany
}
\addinstitution{
 Media Computing Lab.\\
 Nankai University,\\
 Tianjin, China
}

\runninghead{Li, Cheng, Gall}{Dual Pyramid Generative Adversarial Networks}

\def\ie{\emph{i.e}\bmvaOneDot}
\def\eg{\emph{e.g}\bmvaOneDot}

\begin{document}

\maketitle

\begin{abstract}

The goal of semantic image synthesis is to generate photo-realistic images from semantic label maps. It is highly relevant for tasks like content generation and image editing. Current state-of-the-art approaches, however, still struggle to generate realistic objects in images at various scales. In particular, small objects tend to fade away and large objects are often generated as collages of patches. In order to address this issue, we propose a Dual Pyramid Generative Adversarial Network (DP-GAN) that learns the conditioning of spatially-adaptive normalization blocks at all scales jointly, such that scale information is bi-directionally used, and it unifies supervision at different scales. Our qualitative and quantitative results show that the proposed approach generates images where small and large objects look more realistic compared to images generated by state-of-the-art methods. 
\end{abstract}


\section{Introduction}



\begin{figure*}[t]
\centering
\subfigure[Label]{
\begin{minipage}[t]{0.195\linewidth}
\centering
\includegraphics[width=\linewidth,height=1.3cm]{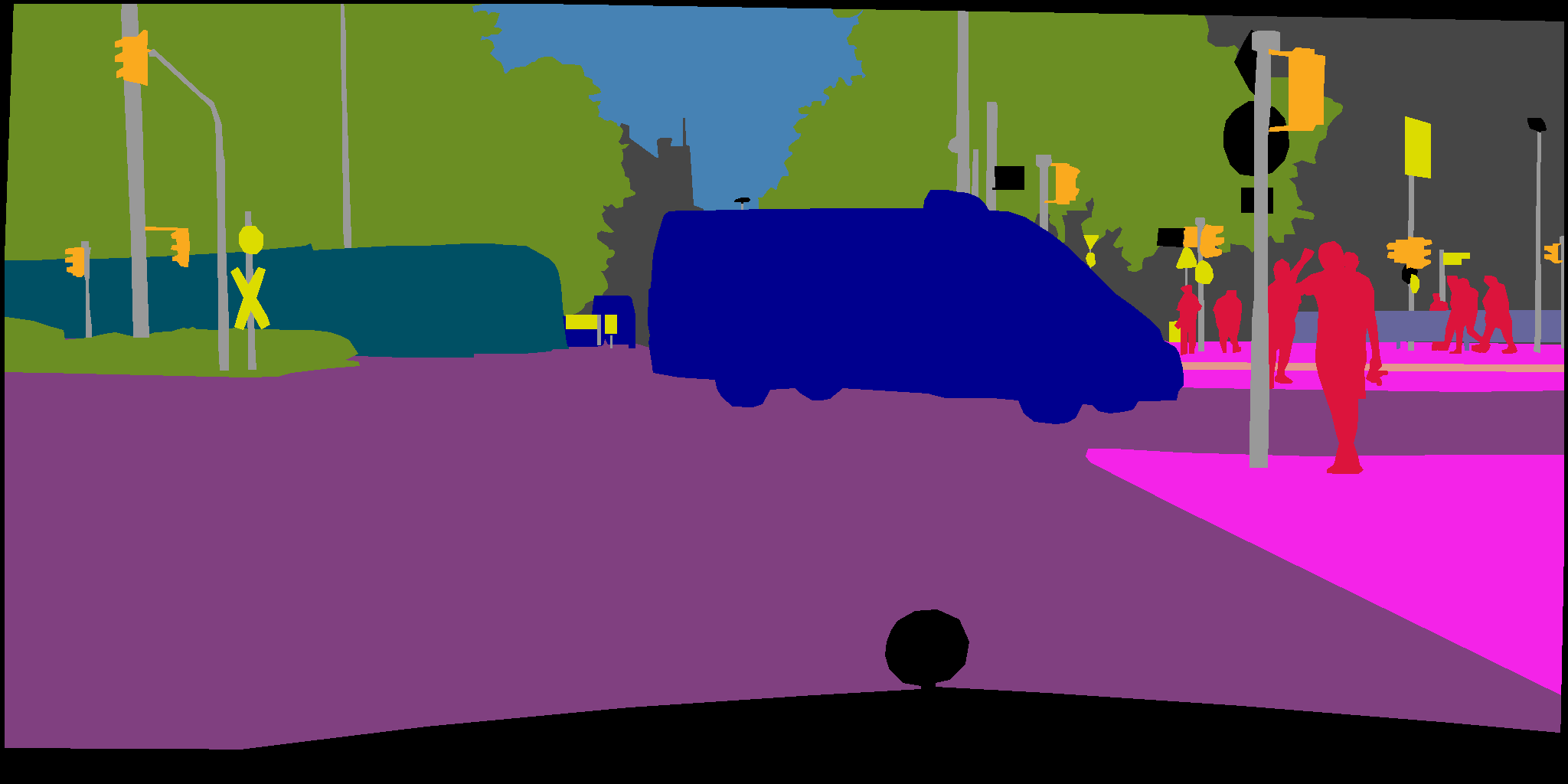}\\
\includegraphics[width=\linewidth,height=1.3cm]{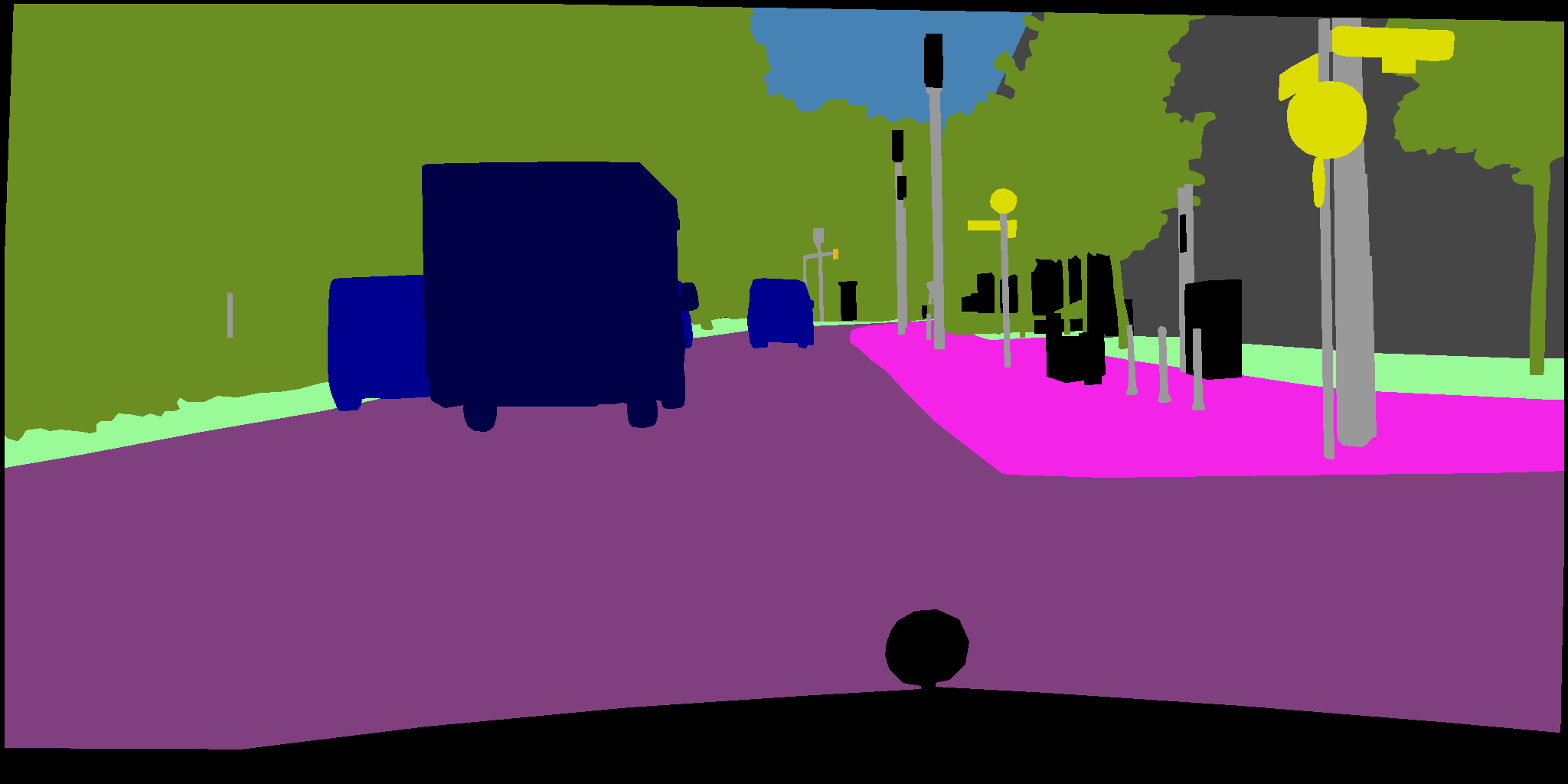}\\
\includegraphics[width=\linewidth,height=1.3cm]{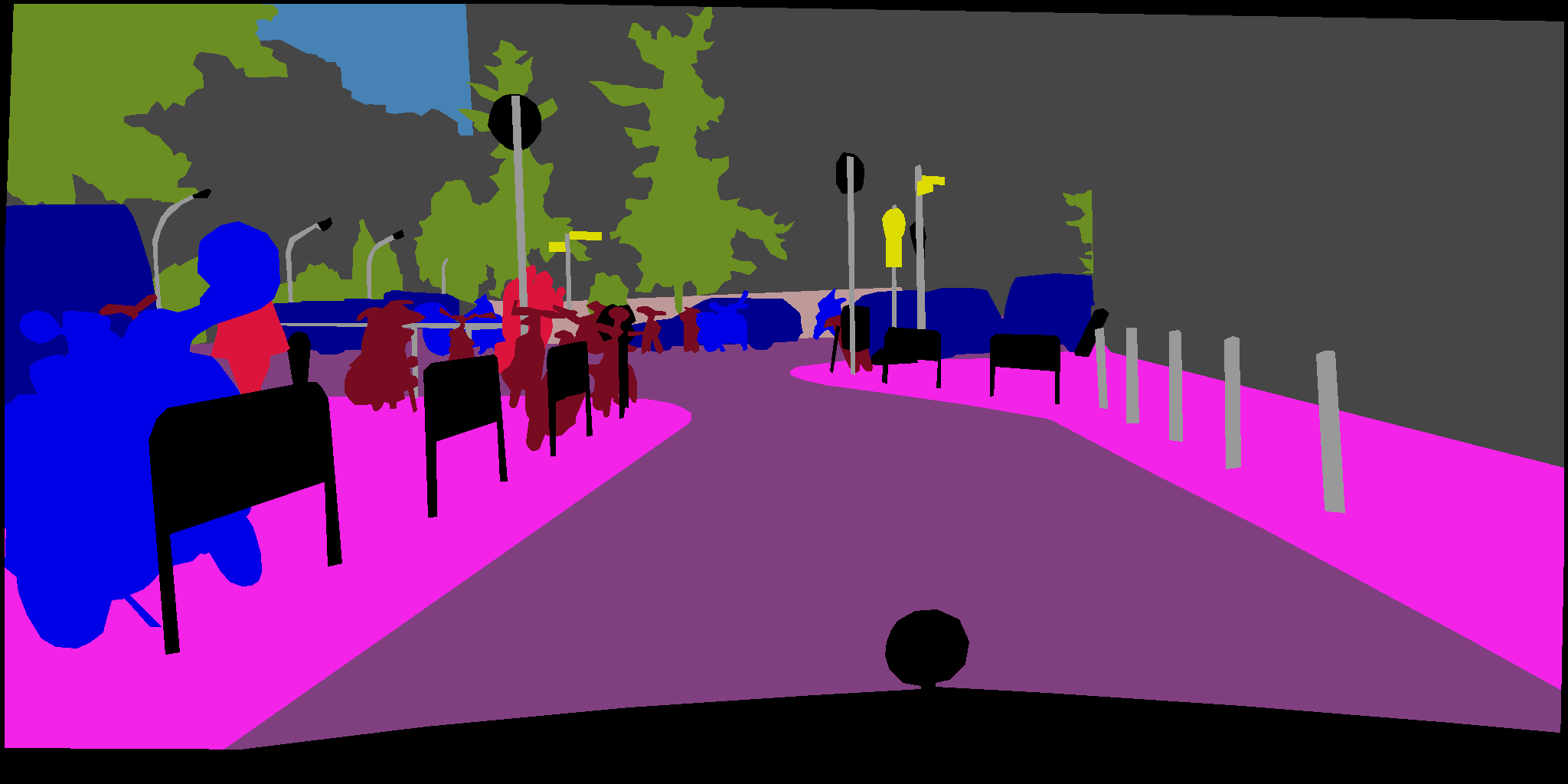} \\
\end{minipage}
}  \hspace*{-3mm}
\subfigure[Ground Truth]{
\begin{minipage}[t]{0.195\linewidth}
\centering
\includegraphics[width=\linewidth,height=1.3cm]{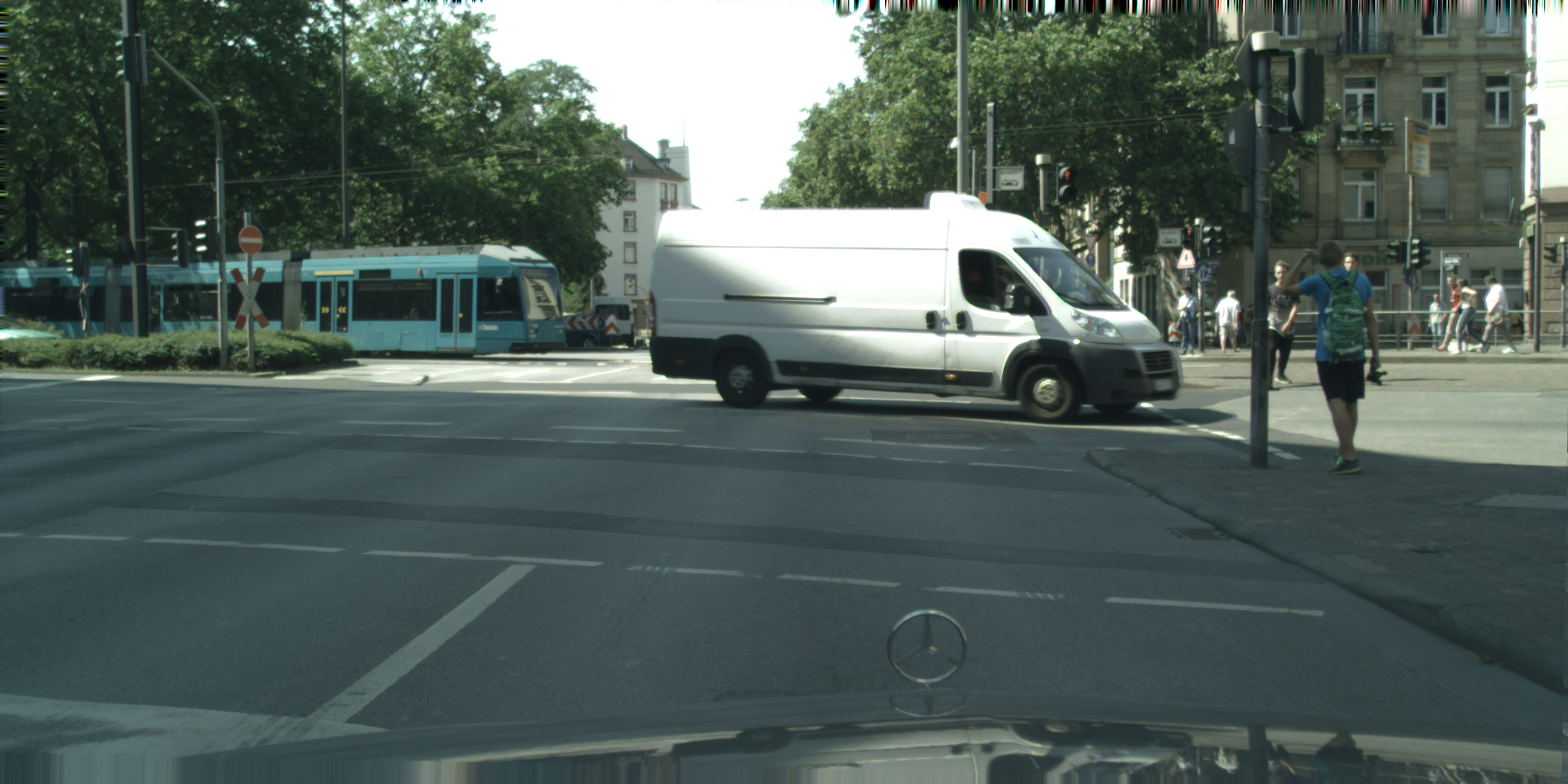}\\
\includegraphics[width=\linewidth,height=1.3cm]{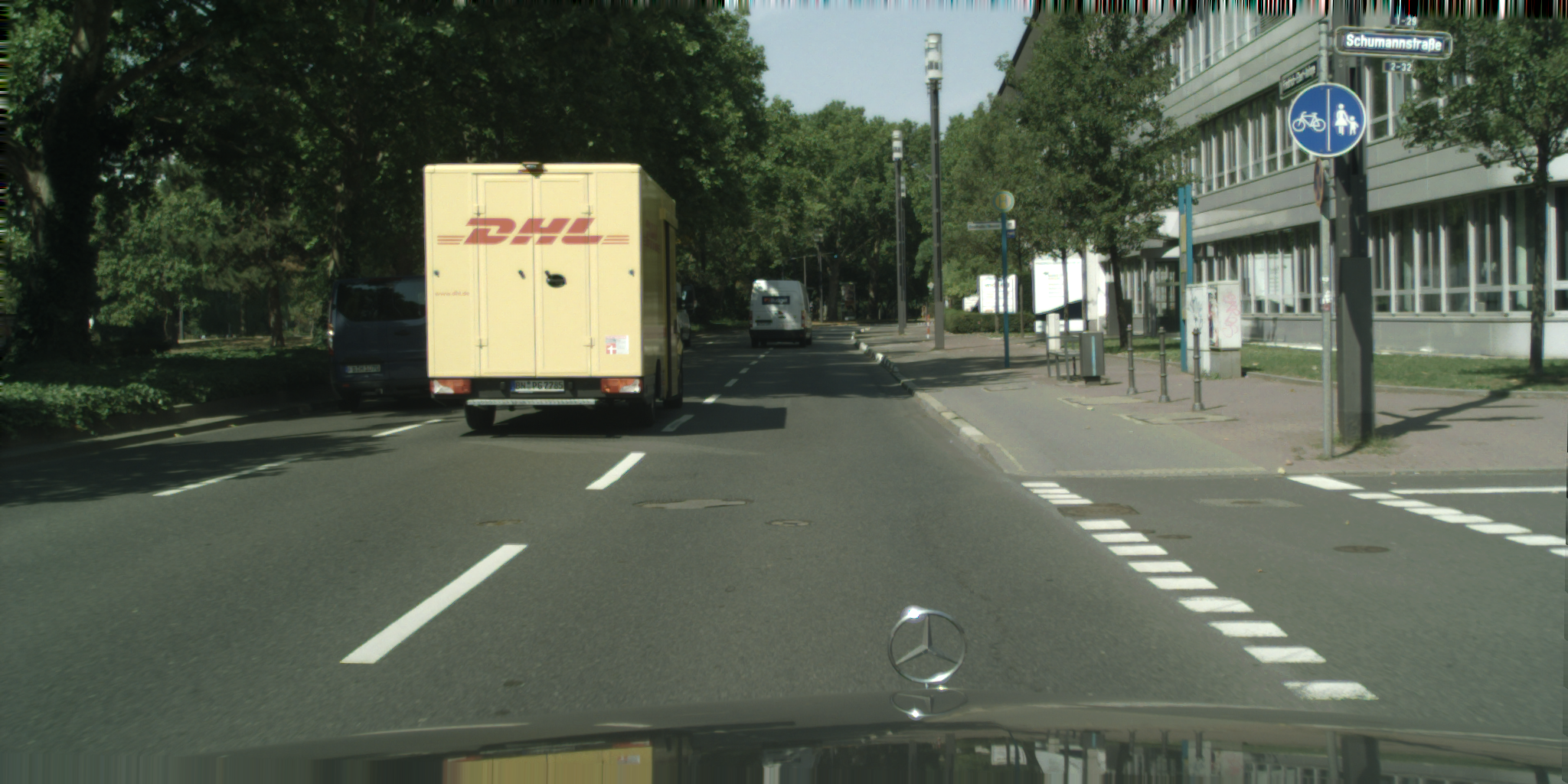}\\
\includegraphics[width=\linewidth,height=1.3cm]{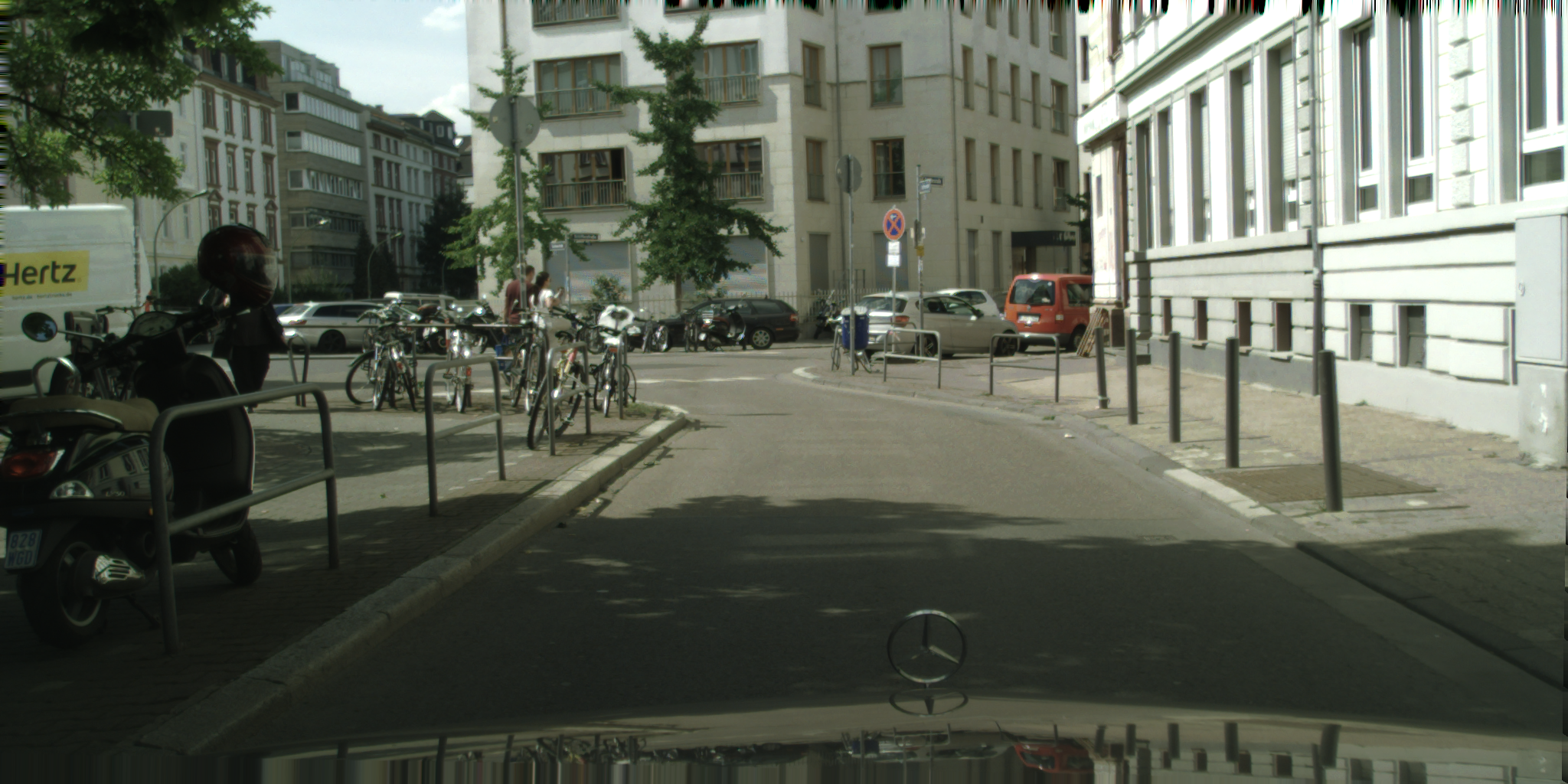} \\
\end{minipage}
}  \hspace*{-3mm}
\subfigure[SPADE]{
\begin{minipage}[t]{0.195\linewidth}
\centering
\includegraphics[width=\linewidth,height=1.3cm]{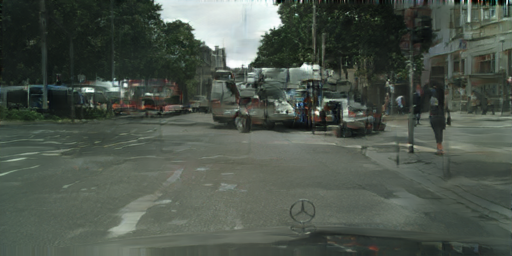}\\
\includegraphics[width=\linewidth,height=1.3cm]{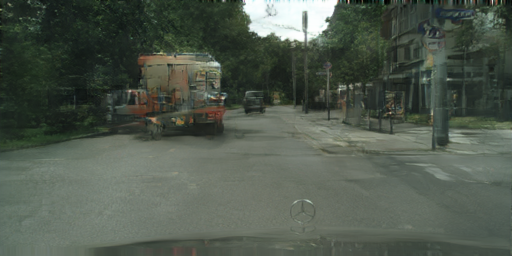}\\
\includegraphics[width=\linewidth,height=1.3cm]{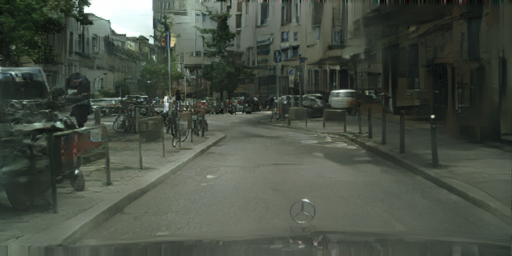} \\
\end{minipage}
} \hspace*{-3mm}
\subfigure[OASIS]{
\begin{minipage}[t]{0.195\linewidth}
\centering
\includegraphics[width=\linewidth,height=1.3cm]{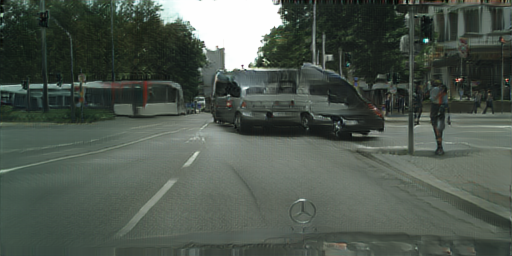}\\
\includegraphics[width=\linewidth,height=1.3cm]{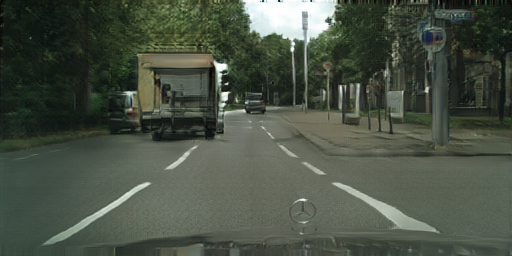}\\
\includegraphics[width=\linewidth,height=1.3cm]{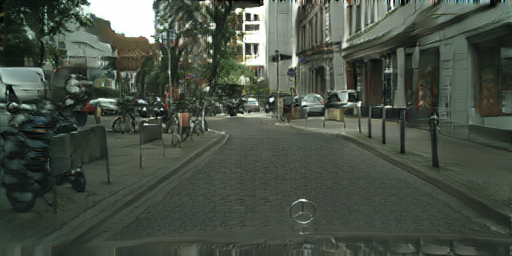} \\
\end{minipage}
} \hspace*{-3mm}
\subfigure[DP-GAN]{
\begin{minipage}[t]{0.195\linewidth}
\centering
\includegraphics[width=\linewidth,height=1.3cm]{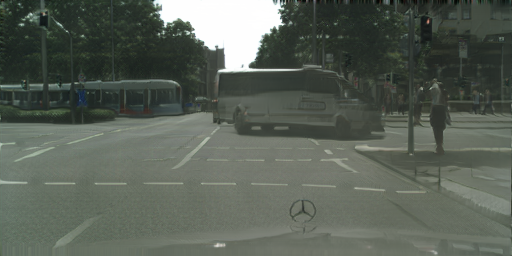}\\
\includegraphics[width=\linewidth,height=1.3cm]{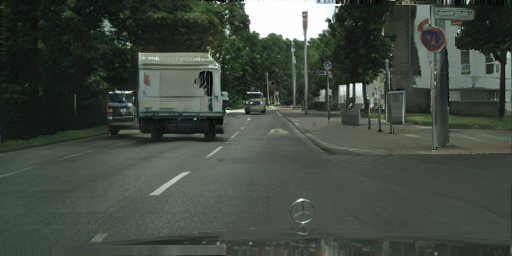}\\
\includegraphics[width=\linewidth,height=1.3cm]{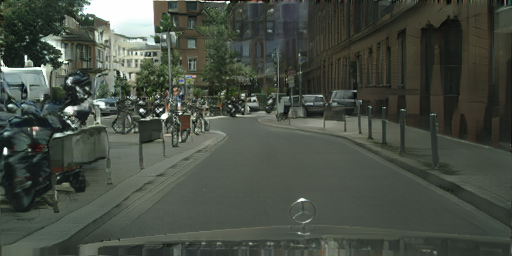} \\
\end{minipage}
} 
\caption{Given a label map (a) as input, current approaches like SPADE (c) or OASIS (d) struggle to generate realistic looking objects. In particular large objects like trucks look highly unrealistic. Instead of generating a consistent appearance of an object, the methods generate a collage of patches. For instance,  when zooming in on the truck in the top row of (d), it can be seen that the truck is a combination of patches from the side and back view of a car or van and four wheels are visible. They also struggle to generate other objects like traffic signs (row 2) or bollards (row 3). This is due to the handling of scale in the generator and discriminator, which is addressed by our approach (e). 
}
\vspace{-4mm}
\label{fig:motivation}
\end{figure*}

Generative adversarial networks \cite{goodfellow2014generative} have become very successful in generating simple photo-realistic images like faces or animals \cite{karras2018progressive,karras2019style,brock2018large}.
However, it is still very challenging to generate images of complex scenes that contain many different objects at various scales. This is also the case when the semantic layout is provided, \eg for content generation or image editing \cite{isola2017image,chen2017photographic,wang2018high}. This task is also known as semantic image synthesis and requires to generate photo-realistic images for given semantic input layouts as illustrated in \figref{fig:motivation}.  

Previous methods for semantic image synthesis usually generate images by an encoder-decoder architecture that takes the semantic label map as input.
However, as shown in \cite{park2019semantic}, the flat segmentation maps will fade away due to the usage of common normalization layers, like instance normalization \cite{ulyanov2016instance}.
To address this issue, spatially-adaptive normalization (SPADE) \cite{park2019semantic} has been proposed, which propagates semantic information throughout the network.
This is achieved by modulating the normalized activation in a spatially-adaptive manner, conditioned on the input segmentation mask. SPADE \cite{park2019semantic}, however, uses as many other works a classification network as discriminator that takes the label map as input along with the image and makes a global decision whether an image is real or not. Since a global discriminator does not enforce the generator to learn a precise alignment with
the input semantic label maps, OASIS \cite{schonfeld_sushko_iclr2021,schonfeld_sushko_ijcv2022} proposed a segmentation-based discriminator to address this issue. 
While the generated images of OASIS are less fuzzy than the objects generated by SPADE as shown in \figref{fig:motivation}, there is still a huge difference between the generated and real images. While areas like road, vegetation, or sky are well generated by OASIS, the approach struggles to generate realistic objects at small and large scales. For instance, large objects like trucks, buses, or trams are generated as collages of patches as shown in the first and second row of \figref{fig:motivation} and small objects like traffic signs or bollards fade away and are barely recognizable as shown in the second and third row of \figref{fig:motivation}. 

\begin{figure*}[t]
    \centering
    \resizebox{0.8\linewidth}{!}{
    \includegraphics{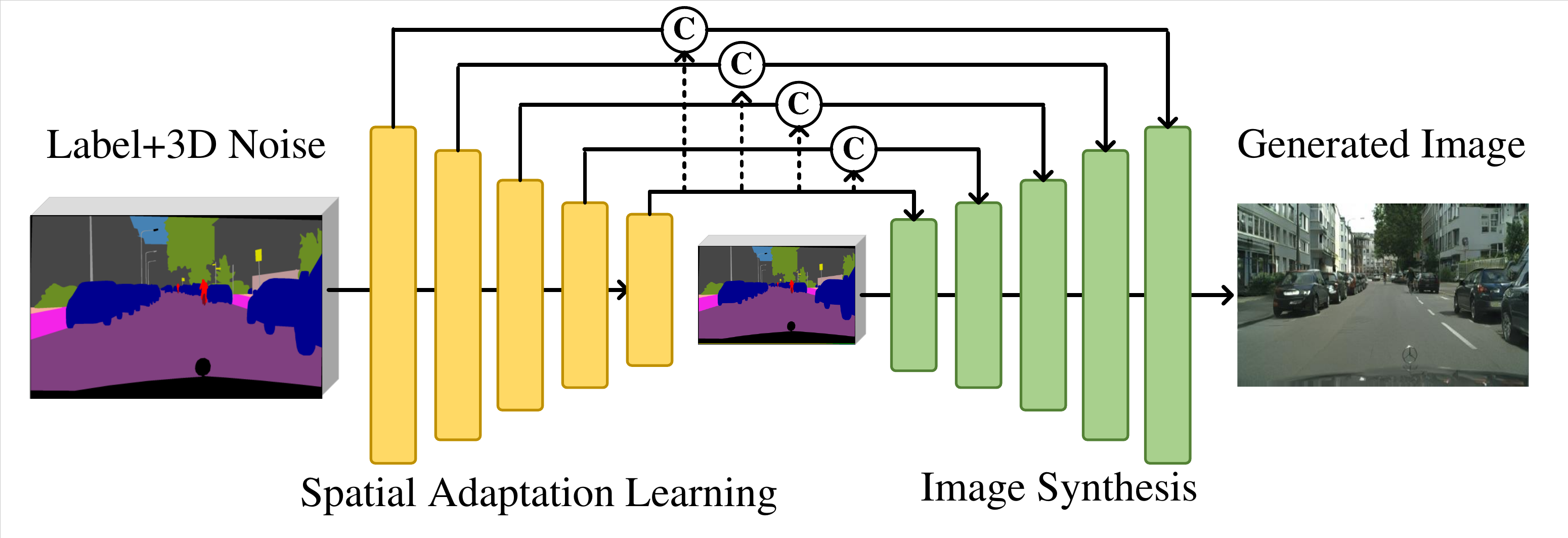}}
    \caption{The dual pyramid generator takes a semantic label map concatenated with a 3D noise tensor as input and generates an image. In contrast to previous works that use only one pyramid (green) consisting of spatially-adaptive normalization (SPADE) blocks with upsampling to generate an image from a downsampled version of the input, we propose a network with two pyramids. The yellow pyramid learns the conditioning of the SPADE blocks in the green pyramid, starting from the full resolution. The scale information is in this way used bi-directionally. In addition, the output of the layer with lowest resolution in the yellow pyramid is upsampled and concatenated (denoted by \textcircled{c}) with the features of the previous layers.}
    \label{fig:gen}
    \vspace{-5mm}
\end{figure*}

In this work, we address this problem and aim to generate images with realistic looking objects at various scales from semantic image maps. This will be achieved by properly dealing with different object sizes in the generator and discriminator. While previous works \cite{park2019semantic,schonfeld_sushko_iclr2021} condition each spatially-adaptive normalization block statically on the resized label map, our generator adapts the conditioning for each scale by a second pyramid as shown in \figref{fig:gen}. In this way, the conditioning is adaptive to the size of the object and artifacts of combining patches of different scales are avoided. This improves the quality of the generated objects both for small objects like signs as well as large objects like trucks as shown in \figref{fig:motivation}.          

For very thin structures like bollards, however, the objects tend to fade away when the objects are getting smaller as shown in \figref{fig:gendis}. While our generator is able to generate such objects, previously proposed classification-based \cite{park2019semantic} or segmentation-based \cite{schonfeld_sushko_iclr2021} discriminators do not handle such small details in the label map, and thus do not encourage the generator to generate small objects. While classification-based discriminators ignore such details completely, the segmentation-based discriminator proposed in \cite{schonfeld_sushko_iclr2021} does not handle scale variations well and extending the discriminator to different scales even deteriorates the quality of the generated images. 
To address this issue, we introduce supervision at different scales. 
While a segmentation-based discriminator provides pixel-wise supervision, we enhance the network by two different types of multi-scale supervision. 
The first one is a multi-scale feature matching loss \cite{johnson2016perceptual}. 
It encourages the generator to generate images that align with the input semantic maps at all scales.
The second one is multi-scale adversarial supervision at patch level.
The loss ensures that important scale information is preserved  within the discriminator such that the generator is enforced to generate more realistic objects at various scales. 
We show that the unification of the three types of supervision is essential to generate realistic objects that align well with the semantic maps.

The contribution of this work is thus two-fold:
1) We propose a dual pyramid generator for semantic image synthesis which adapts the conditioning to the size of the objects. 
2) We propose to unify supervision at pixel, patch, and feature level to enforce the generator to generate realistic objects that are well aligned with the semantic maps.

We evaluate our approach on datasets with complex indoor and outdoor scenes where our approach generates realistic images and outperforms the state-of-the-art by a large margin. The source code is available at \href{https://github.com/sj-li/DP_GAN}{https://github.com/sj-li/DP\_GAN}.

\section{Related Work}

\textbf{Generative Adversarial Networks.} 
Approaches for semantic image synthesis rely on generative adversarial networks (GAN) \cite{goodfellow2014generative}.
The adversarial training enables the generator to produce photo-realistic images from semantic label maps.
While there have been some drawbacks in early works \cite{goodfellow2014generative} which limited their applications, several works have addressed these problems gradually. Since adversarial training is usually unstable, \cite{arjovsky2017wasserstein,salimans2016improved,mao2017least} increase the training stability by regularizers and different loss functions.
\cite{radford2015unsupervised} proposed some design guidelines on the architectural topology of convolutional GANs to increase the robustness during training.

\textbf{Semantic Image Synthesis.} 
Pix2pix \cite{isola2017image} has been the first work that uses a conditional GAN \cite{mirza2014conditional} for semantic image synthesis.
Based on it, Pix2pixHD \cite{wang2018high} aims at generating high-resolution images. This is achieved by a multi-resolution approach.
These works, however, suffer from the problem that flat segmentation maps fade away due to the usage of common normalization layers.
This is solved by SPADE \cite{park2019semantic} which modulates the activations of the normalization layers by a learned affine transformation.
In contrast to SPADE, CC-FPSE \cite{liu2019learning} utilizes the recent progress of dynamic filter networks and learns the convolution weight in the generator directly. A semantics-embedding discriminator is further utilized to improve fine details and boost the semantic alignment.
While previous approaches used a classification-based discriminator, OASIS \cite{schonfeld_sushko_iclr2021} proposed a segmentation-based discriminator, which substantially improved the alignment between generated images and semantic label maps.
Apart from GAN-based methods, some other methods have been also proposed for semantic image synthesis \cite{chen2017photographic,qi2018semi,wang2018high,huang2018multimodal,jiang2020tsit,lee2018diverse}.


\begin{figure*}[t]
    \centering
    \resizebox{0.8\linewidth}{!}{
    \includegraphics{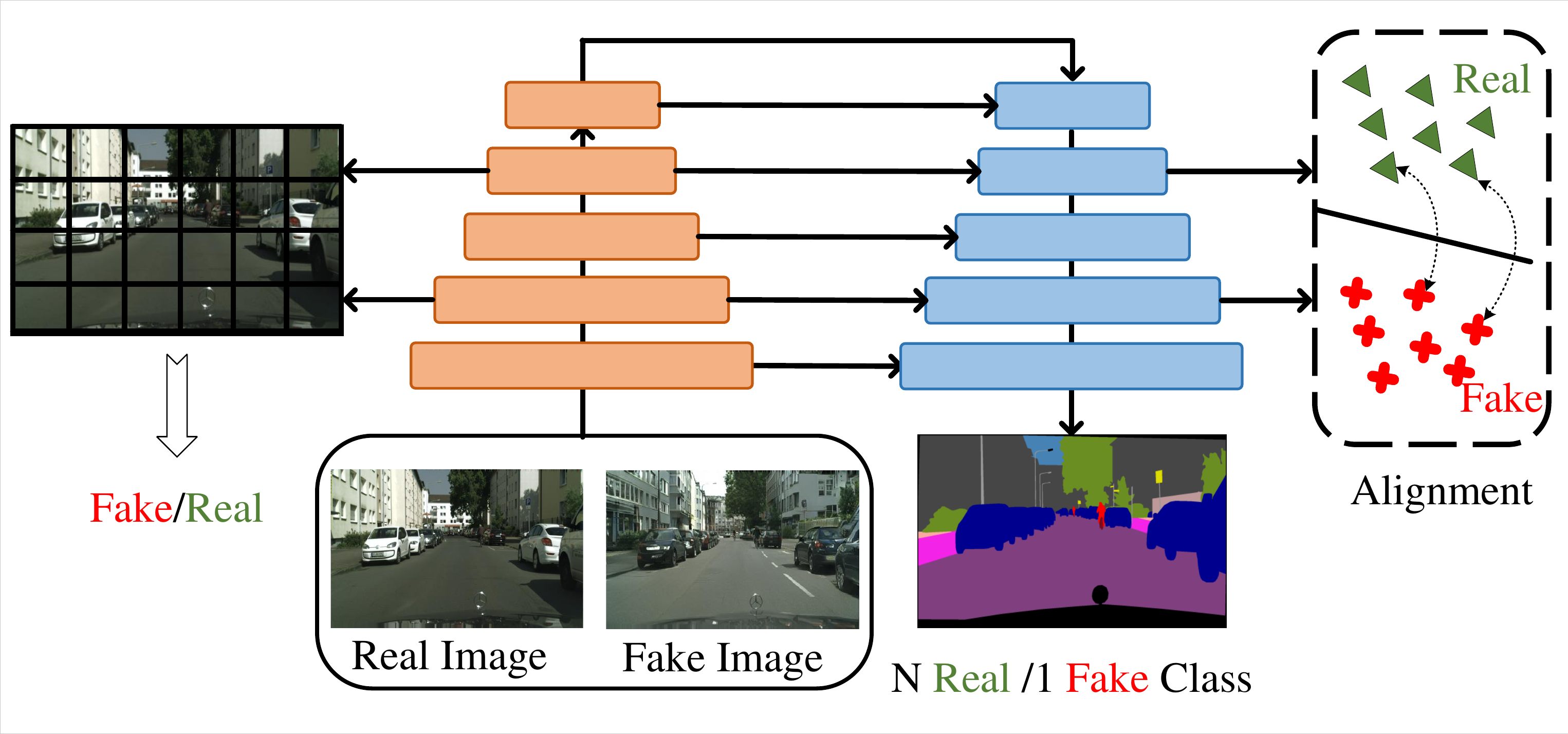}}
    \caption{The proposed discriminator exploits supervision at different scales including a loss at patch level (left), feature level (top right), and pixel level (bottom right). }
    \label{fig:dis}
    \vspace{-5mm}
\end{figure*}

\section{Methodology}


In this section, we discuss the generator and discriminator, which are shown in \figref{fig:gen} and \figref{fig:dis}, respectively. 

\subsection{Dual-Pyramid Generator}
The proposed dual pyramid generator is shown in \figref{fig:gen}. It consists of two pyramids, one for image synthesis and another one for spatial adaptation learning. The input to the generator is a channel-wise concatenation of a label map $\mathbf{l}$ and a 3D noise tensor $\mathbf{z}$ as in \cite{schonfeld_sushko_iclr2021}. The noise tensor can be used to generate different plausible images for the same label map as we show in Fig.~\ref{fig:qvis_multi_modal}. 

The image synthesis pyramid is built from ResNet blocks \cite{he2016deep} with spatially-adaptive normalization (SPADE) \cite{park2019semantic}. At layer $i$, SPADE performs a pixel-wise normalization conditioned on the input $\mathbf{l}$:
\begin{equation}
    \gamma^i_{x,y,c}(\mathbf{l}^i)\frac{h^i_{x,y,c,n}-\mu^i_c}{\sigma^i_c} + \beta^i_{x,y,c}(\mathbf{l}^i)
\end{equation}
where $h^i_{x,y,c,n}$ is the input to the normalization layer, and $n$ denotes the training sample in the batch, $c$ the channel, and $x$ and $y$ the pixel location for the resolution at layer $i$. $\mu^i_c$ and $\sigma^i_c$ are the mean and standard deviation of $h^i_{x,y,c,n}$ in channel $c$, \ie computed over $n$, $x$, and $y$. 

In \cite{park2019semantic,schonfeld_sushko_iclr2021}, the tensors $\gamma^i_{x,y,c}(\mathbf{l}^i)$ and $\beta^i_{x,y,c}(\mathbf{l}^i)$ depend on the input $\mathbf{l}$ that is downsampled to the resolution of the layer $i$. In this way, $\gamma^i$ and $\beta^i$ do not contain any information of small objects in the early layers of the generator for image synthesis due to the low resolution of $\mathbf{l}^i$. As shown in \figref{fig:motivation}, small objects like traffic signs thus do not look realistic and fade. Large objects like trucks also do not look realistic since $\mathbf{l}^i$ just provides redundant information at different scales, which results in collages of patches instead of consistent objects.     

We address this issue by learning $\gamma^i$ and $\beta^i$ for each scale jointly instead of just scaling $\mathbf{l}$. Since $\gamma^i$ and $\beta^i$ are learned, any relevant information of objects including scale information can already be encoded at early layers where the resolution is still low. This is achieved by the spatial adaptation learning pyramid shown in \figref{fig:gen}, which comprises blocks consisting of a convolutional layer, ReLU, and a batch normalization layer. We denote the output of each layer in the spatial adaptation learning pyramid as $\alpha^i$. To match the layer index of the image synthesis pyramid, we use an inverse index $i$ for the spatial adaptation learning pyramid, \ie $\alpha^0$ is the output of the last layer of the spatial adaptation learning pyramid,  which has the lowest resolution. For $i>0$, we concatenate $\alpha^i$ with an upsampled version of $\alpha^0$. The upsampling ($\mathrm{Up}$) is conducted once to match the resolution of $\alpha^0$ to the resolution of $\alpha^i$. Finally, we apply a convolution to obtain $\gamma^i$ and $\beta^i$:
\begin{equation}
\begin{split}
     \gamma^i &= \mathrm{Conv}(\alpha^i \oplus \mathrm{Up}(\alpha^0))\\
     \beta^i &= \mathrm{Conv}(\alpha^i \oplus \mathrm{Up}(\alpha^0)).   
\end{split}
\end{equation}
Since the learned modulation, \ie$\gamma^i$ and $\beta^i$, goes from the high resolution
to the low resolution, but the image generator goes in the
other direction from the low resolution to the high resolution,
we term it bi-directional. By learning the spatially-adaptive normalization for all scales jointly, our network generates not only a realistic background, but also realistic objects as shown in \figref{fig:motivation}.

\subsection{Scale-Enhancement Discriminator}\label{sec:disc}

Semantic image synthesis requires that the generated images are well aligned with the given semantic label map. Since commonly used classification-based discriminators \cite{park2019semantic} do not provide strong supervision for the alignment, \cite{schonfeld_sushko_iclr2021} proposed a segmentation-based discriminator, which provides semantic supervision at the pixel level. Using a loss only at the pixel level, however, does not ensure realistic structures at all scales, as \figref{fig:motivation} shows. To solve this issue, we propose a discriminator that provides supervision at the pixel, patch, and feature level as illustrated in \figref{fig:dis}.

As in \cite{schonfeld_sushko_iclr2021}, we use an encoder-decoder architecture consisting of ResNet blocks \cite{he2016deep}. The network predicts for each pixel the probabilities of $N+1$ classes where $N$ is the number of semantic classes and an additional `fake' class is added.
During training, the ground-truth label for each pixel is defined by the input label map for a real image and by the `fake' class for a generated image. 
Since the classes are imbalanced, a weighted $N+1$-class cross-entropy loss is used:
\begin{equation}
\begin{split}
        \mathcal{L}_{pixel} = -\mathbf{E}_{(\mathbf{x}, \mathbf{l})}\left[\sum_{c=1}^N\alpha_c \sum_{x,y}^{H\times W}\mathbf{l}_{x,y,c} \log D(\mathbf{x})_{x,y,c}\right] 
        - \mathbf{E}_{(\mathbf{z},\mathbf{l})}\left[\sum_{x,y}^{H\times W}\log D(G(\mathbf{z},\mathbf{l}))_{x,y,c=N+1}\right]
\end{split}
    \label{eq:l_d_seg}
\end{equation}
where $\mathbf{x}$ is a real image, $\mathbf{l}$ is the semantic label map, and $(\mathbf{z}, \mathbf{l})$ is the concatenation of the 3D noise tensor and the label map.
The weight $\alpha_c$ is modeled as inverse of the per-pixel class frequency to give rare classes more weight:
\begin{equation}
    \alpha_c = E_{\mathbf{l}}\left[\frac{H\times W}{\sum_{x,y}^{H\times W} \mathbf{l}_{x,y,c}}\right].
\end{equation}

Since the pixel-wise loss is insufficient as shown in \figref{fig:motivation}, we add two additional loss functions. The first one is a multi-scale adversarial patch-based loss \cite{miyato2018spectral}, which is applied to feature maps of lower resolution:   
\begin{equation}
\begin{split}
    \mathcal{L}^i_{ms} = &-\mathbf{E}_{\mathbf{x}}\left[\min(-1 + D^i_{p}(\psi^i(\mathbf{x})), 0)\right]  - \mathbf{E}_{(\mathbf{z}, \mathbf{l})}\left[\min(-1-D^i_{p}(\psi^i(G(\mathbf{z},\mathbf{l}))), 0)\right] 
\end{split}
\label{eq:ms}
\end{equation}
where $\psi^i(\mathbf{x})$ and $\psi^i(G(\mathbf{z},\mathbf{l}))$ are the feature maps at the layer $i$ of the encoder of the discriminator for a real or generated image. $D^i_{p}$ is a convolutional network consisting of blocks with convolutional layer, ReLU, and a batch
normalization layer. It makes the prediction for each feature which corresponds to patches of different scales in the original image. The loss $\mathcal{L}_{ms}$ is then averaged over all layers. In our implementation, we apply it after the  fourth and sixth block. 

The second loss is based on a multi-scale feature matching loss \cite{johnson2016perceptual} that computes the L2 distance between the features of a real image and a corresponding generated image. While the features are extracted at the layers of the decoder of the discriminator, the loss is used for training the generator $G(\mathbf{z},\mathbf{l})$ and added in \eqref{eq:G}. The purpose of the loss is to encourage the generator to generate images that are close to the corresponding real images in the feature space of the decoder of the discriminator. For a single layer $i$, the loss is defined by        
\begin{equation}
\begin{split}
    \mathcal{L}^i_{fm} = \mathbf{E}_{(\mathbf{x}, \mathbf{l}, \mathbf{z})} \left[ \frac{\sum_{x,y}^{H^i \times W^i}\left\Vert \phi^i(\mathbf{x})_{x,y} - \phi^i(G(\mathbf{z},\mathbf{l}))_{x,y} \right\Vert^2_2}{C^i \times H^i \times W^i}\right], 
\end{split}
\label{eq:fm}
\end{equation}
where $\phi^i$ is the feature vector of size $C^i$ and $H^i \times W^i$ corresponds to the resolution at layer $i$. As $\mathcal{L}_{ms}$, the loss $\mathcal{L}_{fm}$ is averaged over all layers. In our implementation, we apply it after all blocks except for the first one. 

In our experiments, we show that applying $\mathcal{L}_{ms}$ to the encoder of the discriminator and $\mathcal{L}_{fm}$ to the decoder performs best. In this way, $\mathcal{L}_{ms}$ ensures that scale information is not lost in the encoder, which improves the quality of generated objects at all scales, and $\mathcal{L}_{fm}$ ensures a better alignment of the generated images with real images. 

\subsection{Training}
The generator and discriminator are trained jointly. For the generator, we use the loss: 
\begin{equation}\label{eq:G}
\begin{split}
    \mathcal{L}_G = &-\mathbf{E}_{(\mathbf{z}, \mathbf{l})}\left[\sum_{c=1}^N\alpha_c \sum_{x,y}^{H\times W}\mathbf{l}_{x,y,c} \log D(G(\mathbf{z},\mathbf{l}))_{x,y,c}\right] \\
    &- \frac{1}{L} \sum_{i=1}^L \mathbf{E}_{(\mathbf{z}, \mathbf{l})} \left[\min(-1 + D^i_{p}(\psi^i(G(\mathbf{z}, \mathbf{l}))), 0)\right] + \mathcal{L}_{fm}   
\end{split}
\end{equation}
where each term corresponds to supervision at pixel, patch, and feature level. Furthermore, 
we include the LabelMix regularization proposed in \cite{schonfeld_sushko_iclr2021} for a fair comparison:
\begin{equation}
\begin{split}
    \mathcal{L}_{LM} = &||D_{logits}(\mathrm{LabelMix}(\mathbf{x}, \hat{\mathbf{x}}, M)) - \mathrm{LabelMix}(D_{logits}(\mathbf{x}), D_{logits}(\hat{\mathbf{x}}), M)||_2^2
\end{split}
\end{equation}
where $D_{logits}$ are the logits before the last softmax and $M$ is a binary mask for mixing real and generated images $(\mathbf{x}, \hat{\mathbf{x}})$ by 
$\mathrm{LabelMix}(\mathbf{x}, \hat{\mathbf{x}},M) = M \odot \mathbf{x} + (1-M) \odot \hat{\mathbf{x}}$. The entire loss is thus given by 
\begin{equation}
     \mathcal{L} = \mathcal{L}_G + \mathcal{L}_{pixel} + \mathcal{L}_{ms} + \lambda_{LM} \mathcal{L}_{LM}.
\end{equation}
More details regarding the network are provided in the supplemental material.


\begin{table}[t]
    \centering
    \setlength{\tabcolsep}{2.5mm}
    \resizebox{0.8\linewidth}{!}{
    \begin{tabular}{c|cc|cc|cc}
        \multirow{2}{*}{Methods}  &  \multicolumn{2}{c|}{Cityscapes} & \multicolumn{2}{c|}{ADE20K} & \multicolumn{2}{c}{ADE20K-Outdoor} \\
        \cline{2-7}
        & FID \textcolor{blue}{$\downarrow$}  & mIoU \textcolor{red}{$\uparrow$} & FID \textcolor{blue}{$\downarrow$} & mIoU \textcolor{red}{$\uparrow$} & FID \textcolor{blue}{$\downarrow$}  & mIoU \textcolor{red}{$\uparrow$} \\
        \hline
        CRN \cite{chen2017photographic} & 104.7  & 52.4  & 73.3  & 22.4 & 99.0 & 16.5 \\
        pix2pixHD \cite{wang2018high} & 95.0 &  58.3 &  81.8 & 20.3 & 97.8 & 17.4 \\
        SPADE \cite{park2019semantic} & 71.8 & 62.3 &  33.9 & 38.5 & 63.3 & 30.8\\
        DAGAN \cite{tang2020dual} & 60.3 &  66.1 & 31.9 & 40.5 & N/A & N/A\\
        LGGAN \cite{tang2020local} & 57.7 &  68.4 & 31.6 & 41.6 & N/A & N/A\\
        CC-FPSE \cite{liu2019learning} & 54.3  & 65.5  & 31.7  & 43.7 & N/A & N/A\\
        SIMS \cite{qi2018semi} & 49.7 &  47.2  & N/A & N/A & 67.7 & 13.1\\
        OASIS \cite{schonfeld_sushko_iclr2021} & 47.7  & 69.3 & 28.3 & 48.8 & 48.6 & \textbf{40.4} \\ 
        \hline
        DP-GAN & \textbf{44.1} & \textbf{73.6} & \textbf{26.1} & \textbf{52.7} & \textbf{45.8} & \textbf{40.4} \\
    \end{tabular}}
    \vspace{1mm}
    \caption{Comparison to state-of-the-art methods on different datasets. Lower FID and higher mIoU are better. Bold denotes the best performance.}
    \label{tab:comparasion_stoa}
\end{table}

\begin{table*}[t]
    \centering
    \resizebox{\linewidth}{!}{%
    \begin{tabular}{c|ccccccccccccccccccc|c}
       & \rotatebox{90}{road} & \rotatebox{90}{swalk} & \rotatebox{90}{build} & \rotatebox{90}{wall} & \rotatebox{90}{fence} & \rotatebox{90}{\underline{pole}} & \rotatebox{90}{\underline{tlight}} & \rotatebox{90}{\underline{sign}} & \rotatebox{90}{veg} & \rotatebox{90}{terrain} & \rotatebox{90}{sky} & \rotatebox{90}{\underline{person}} & \rotatebox{90}{\underline{rider}} & \rotatebox{90}{\underline{car}} & \rotatebox{90}{\underline{truck}} & \rotatebox{90}{\underline{bus}} & \rotatebox{90}{\underline{train}} & \rotatebox{90}{\underline{mbike}} & \rotatebox{90}{\underline{bike}} & \rotatebox{90}{obj-mIoU} \\
       \hline
       SPADE \cite{park2019semantic} & 97.5 & 80.8 & 88.5 & 54.3 & 50.6 & 40.4 & 39.0 & 41.9 & 88.7 & 69.1 & 92.0 & 66.2 & 41.5 & 89.1 & 64.6 & 73.2 & 42.1 & 29.7 & 61.5 & 53.6 \\ 
       DAGAN \cite{tang2020dual} & 97.4 & 80.0 & 89.0 & 60.1 & 53.7 & 41.2 & 39.4 & 46.5 & 88.9 & 65.9 & 92.5 & 66.8 & 45.8 & 89.9 & 71.2 & 75.4 & 57.0 & 25.8 & 60.9 & 56.4 \\ 
       CC-FPSE \cite{liu2019learning} & 97.7 & 82.8 & \textbf{89.8} & 56.1 & 61.3 & 42.3 & 41.8 & 50.4 & \textbf{89.6} & 69.3 & 92.5 & 68.5 & 48.3 & 90.2 & 69.7 & 74.3 & 45.4 & 43.4 & 65.0 & 58.1 \\ 
       LGGAN \cite{tang2020local} & \textbf{97.8} & \textbf{83.1} & 89.7 & 59.8 & 56.0 & 42.5 & 42.8 & 50.5 & 89.5 & 70.0 & \textbf{92.7} & \textbf{69.0} & 48.6 & \textbf{90.6} & 72.2 & 80.2 & 52.4 & 38.8 & 64.0 & 59.2 \\ 
       OASIS \cite{schonfeld_sushko_iclr2021} & 96.9 & 79.2 & 85.1 & 70.3 & 64.2 & 41.6 & 50.7 & 49.9 & 85.0 & \textbf{74.8} & 92.0 & 64.9 & 54.0 & 88.4 & 65.6 & 79.9 & 63.4 & 53.9 & 63.7 & 61.5 \\ 
       \hline
       DP-GAN & 97.5 & 81.9 & 87.2 & \textbf{71.4} & \textbf{72.7} & \textbf{46.9} & \textbf{55.5} & \textbf{60.3} & 87.3 & 72.9 & 92.4 & 67.4 & \textbf{55.5} & 89.9 & \textbf{81.5} & \textbf{83.1} & \textbf{73.9} & \textbf{55.3} & \textbf{66.9} & \textbf{66.9} \\ 
       \hline
    \end{tabular}}
    \caption{Per-class IoU for Cityscapes. Classes corresponding to objects, \ie the object, human, and vehicle group of Cityscapes, are underlined.
    obj-mIoU is mIoU only for object classes. mIoU for all classes is reported in \tabref{tab:comparasion_stoa}.}
    \vspace{-5mm}
    \label{tab:miou}
\end{table*}

\begin{table}[t]
    \centering
    \setlength{\tabcolsep}{2.5mm}
    \resizebox{0.5\linewidth}{!}{
    \begin{tabular}{c|c|c|c}
         &  OASIS~\cite{schonfeld_sushko_iclr2021} & DP-GAN \\
        \hline
        Cityscapes & 48.1$\pm$0.18  & \textbf{44.6$\pm$0.17}   \\ 
        \hline
        ADE20K &  29.0$\pm$0.03 & \textbf{26.7$\pm$0.09}   \\
    \end{tabular}}
    \vspace{1mm}
    \caption{The mean and variance of FID computed over 5 runs. In this experiment, we change the noise tensor $z$ to generate diverse images as shown in \figref{fig:qvis_multi_modal}. 
    }
    \label{tab:multi_modal}
\end{table}

\begin{figure*}[t]
\centering
\subfigure[Label]{
\begin{minipage}[t]{0.15\linewidth}
\centering
\includegraphics[width=\linewidth,height=2.0cm]{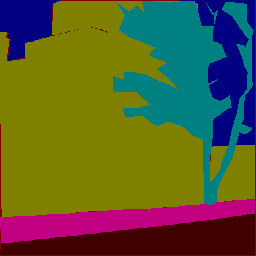}\\
\includegraphics[width=\linewidth,height=2.0cm]{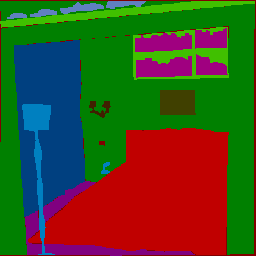}
\end{minipage}
} \hspace{-3mm}
\subfigure[Ground Truth]{
\begin{minipage}[t]{0.15\linewidth}
\centering
\includegraphics[width=\linewidth,height=2.0cm]{figs/vis_multi_modal/gt/ADE_val_00000023.jpg}\\
\includegraphics[width=\linewidth,height=2.0cm]{figs/vis_multi_modal/gt/ADE_val_00000013.jpg}
\end{minipage}
} \hspace{-3mm}
\subfigure[$z_1$]{
\begin{minipage}[t]{0.15\linewidth}
\centering
\includegraphics[width=\linewidth,height=2.0cm]{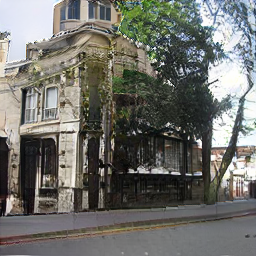}\\
\includegraphics[width=\linewidth,height=2.0cm]{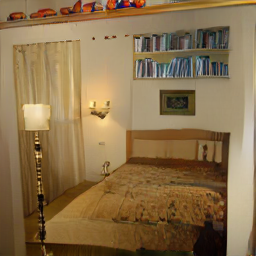}
\end{minipage}
} \hspace{-3mm}
\subfigure[$z_2$]{
\begin{minipage}[t]{0.15\linewidth}
\centering
\includegraphics[width=\linewidth,height=2.0cm]{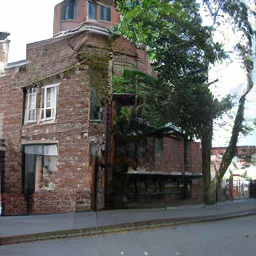}\\
\includegraphics[width=\linewidth,height=2.0cm]{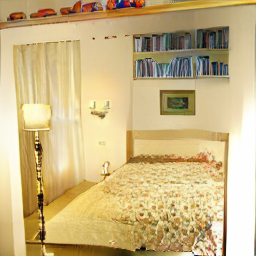}
\end{minipage}
} \hspace{-3mm}
\subfigure[$z_3$]{
\begin{minipage}[t]{0.15\linewidth}
\centering
\includegraphics[width=\linewidth,height=2.0cm]{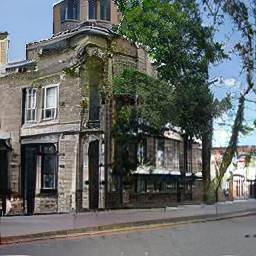}\\
\includegraphics[width=\linewidth,height=2.0cm]{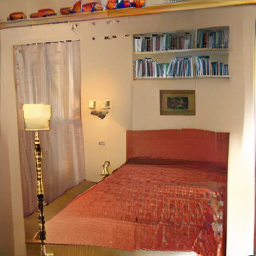}
\end{minipage}
} \hspace{-3mm}
\subfigure[$z_4$]{
\begin{minipage}[t]{0.15\linewidth}
\centering
\includegraphics[width=\linewidth,height=2.0cm]{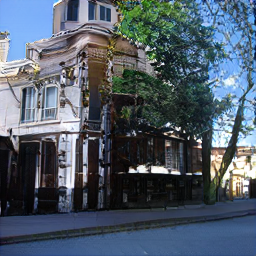}\\
\includegraphics[width=\linewidth,height=2.0cm]{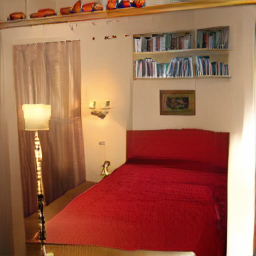}
\end{minipage}
}
\caption{Images generated with four different random noise tensors $z$.}
\label{fig:qvis_multi_modal}
\vspace{-5mm}
\end{figure*}

\section{Experiments}
\label{sec:exp}

\textbf{Datasets.} We evaluate our method on the two challenging datasets Cityscapes \cite{cordts2016cityscapes}
and ADE20K \cite{zhou2017scene}
, including the ADE20K-outdoor subset. We follow the protocol used in \cite{park2019semantic,schonfeld_sushko_iclr2021}. The Cityscapes dataset comprises complex street images recorded by a driving car. It contains 2975 training images and 500 validation images with 35 classes, which are mapped to 19 classes for evaluating the semantic alignment. While the resolution of the images is $1024\times2048$, we resized them to $256\times512$ as in previous works for a fair comparison. The ADE20K dataset is a large-scale dataset aiming at scene understanding. It contains 20210 training images and 2000 validation images with 150 classes. The images are resized to $256\times256$. The ADE20K-outdoor dataset is a subset of the ADE20K dataset and contains only outdoor scenes. It contains 9649 training images and 943 validation images.

\begin{figure*}[t]
\centering
\subfigure[Label]{
\begin{minipage}[t]{0.195\linewidth}
\centering
\includegraphics[width=\linewidth,height=1.3cm]{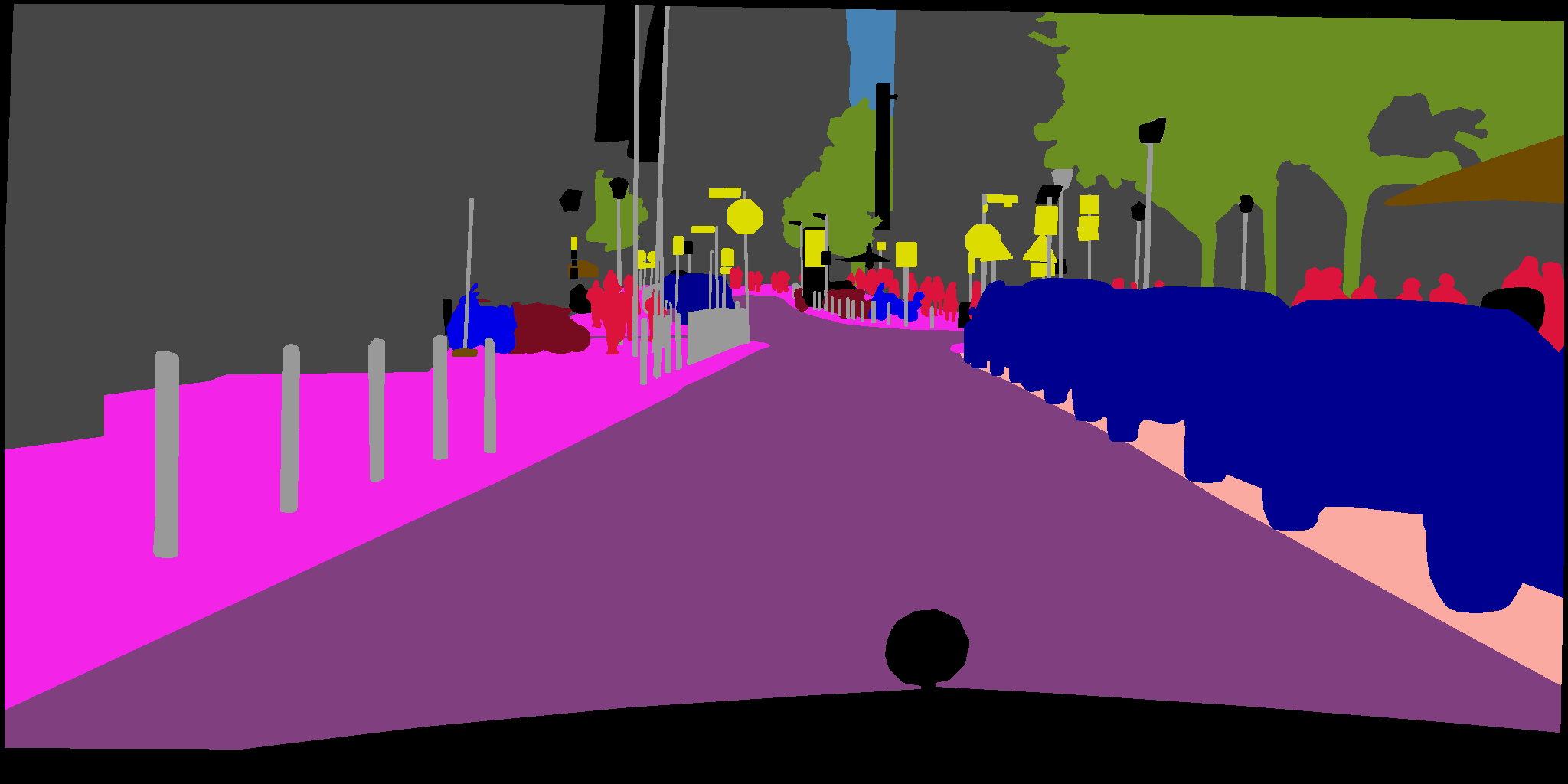}
\includegraphics[width=\linewidth,height=1.3cm]{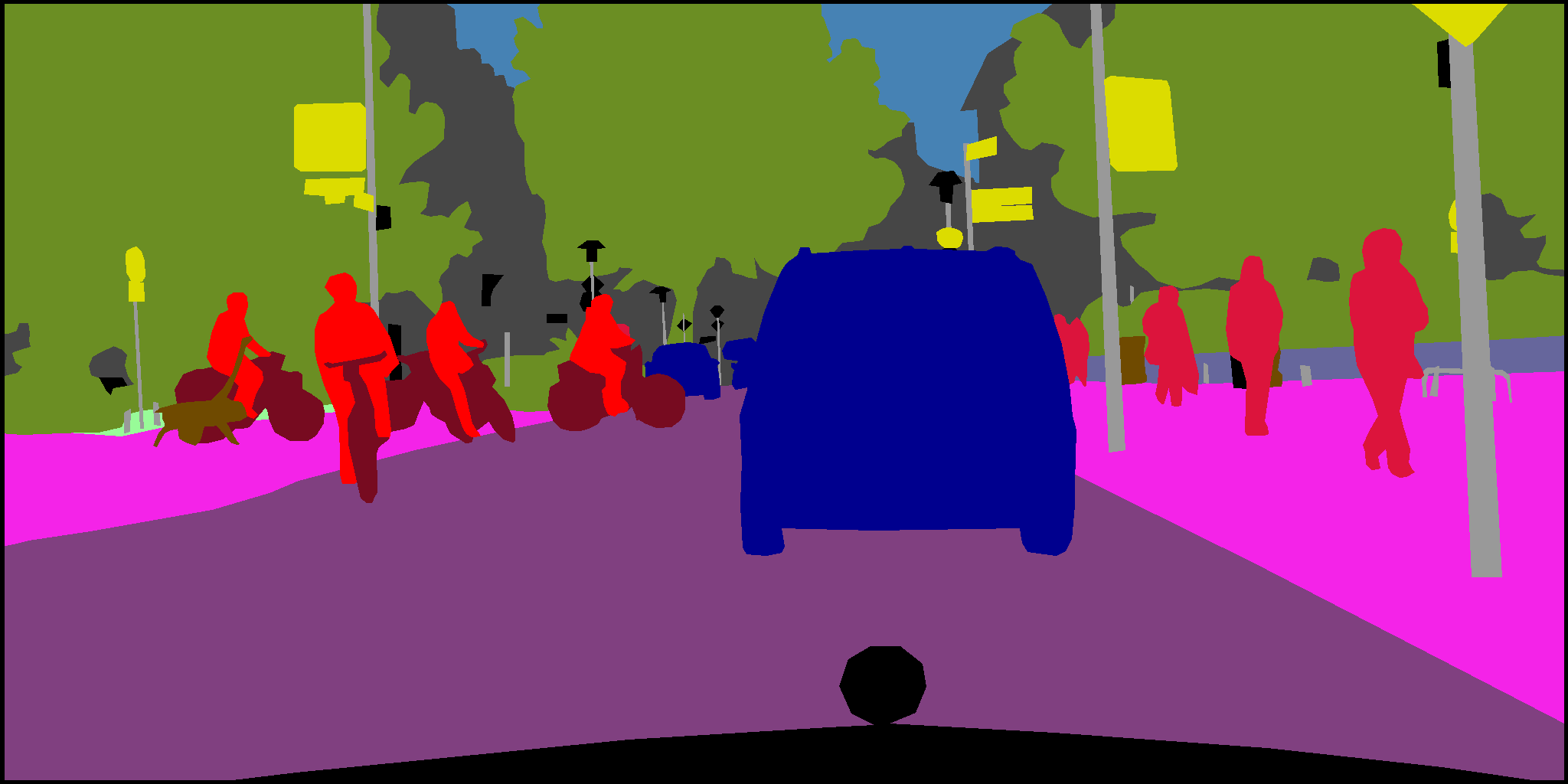}
\end{minipage}
}  \hspace*{-3mm}
\subfigure[OA-OA]{
\begin{minipage}[t]{0.195\linewidth}
\centering
\includegraphics[width=\linewidth,height=1.3cm]{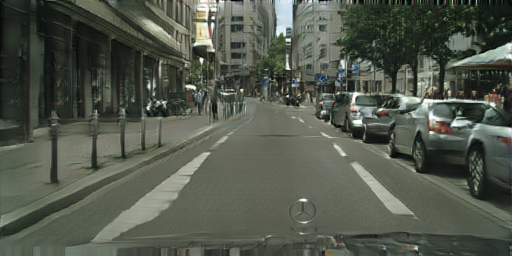}
\includegraphics[width=\linewidth,height=1.3cm]{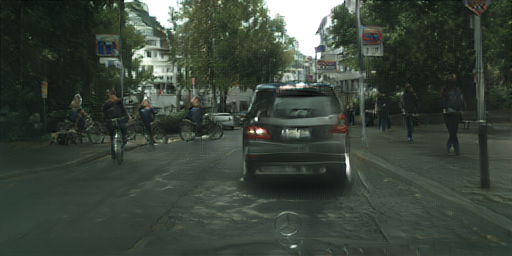}
\end{minipage}
}  \hspace*{-3mm}
\subfigure[DP-OA]{
\begin{minipage}[t]{0.195\linewidth}
\centering
\includegraphics[width=\linewidth,height=1.3cm]{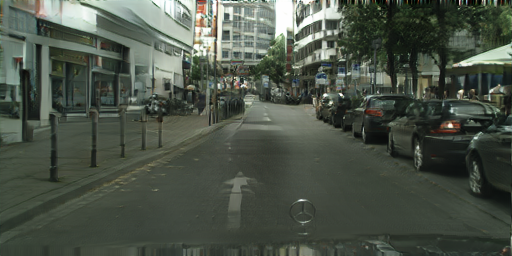}
\includegraphics[width=\linewidth,height=1.3cm]{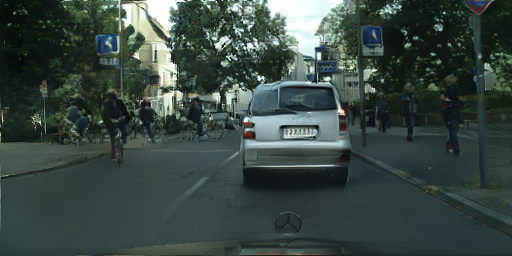}
\end{minipage}
} \hspace*{-3mm}
\subfigure[OA-DP]{
\begin{minipage}[t]{0.195\linewidth}
\centering
\includegraphics[width=\linewidth,height=1.3cm]{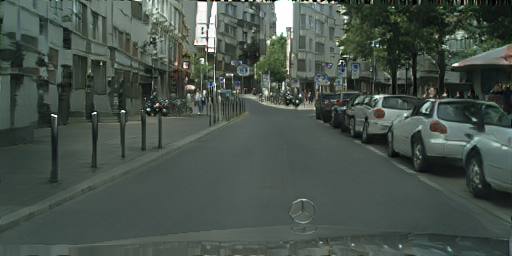}
\includegraphics[width=\linewidth,height=1.3cm]{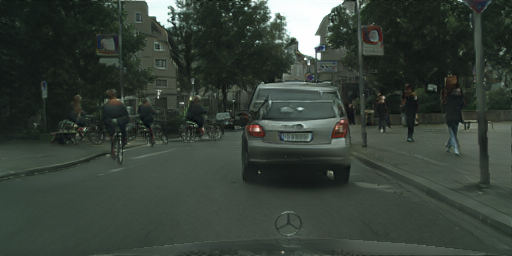}
\end{minipage}
} \hspace{-3mm}
\subfigure[DP-DP]{
\begin{minipage}[t]{0.195\linewidth}
\centering
\includegraphics[width=\linewidth,height=1.3cm]{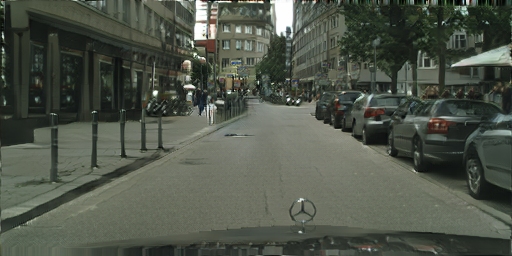}
\includegraphics[width=\linewidth,height=1.3cm]{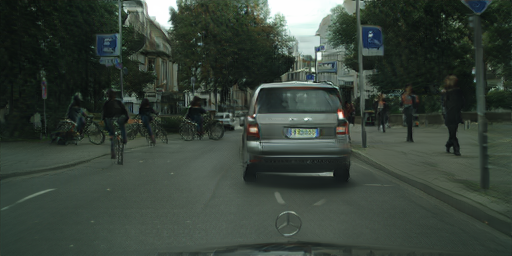}
\end{minipage}
} 
\caption{Impact of the proposed generator and discriminator. DP or OA denote if the generator or discriminator from OASIS (OA) or our approach (DP) are used, \eg DP-OA means that our generator is combined with the discriminator from OASIS. First row: Given the input label map (a), the image generated by OASIS (b) looks reasonable but when zooming in we observe that the parking cars on the right hand side contain strong artifacts and the five bollards on the left hand side fade away. Using the proposed generator (c) improves the quality of the image and the parking cars look realistic. Small objects like the bollards, however, still fade away. If we use only our discriminator (d), the bollards are clearly visible but the building on the left hand side looks unrealistic. Only our full model (e) generates an image of high fidelity. The parking cars on the right hand side and the building and bollards on the left hand side look realistic. Second row: Although there is only one large car in the label map (a), the generator of OASIS overlays a small car over a large car (d), which becomes visible when zooming in. Whereas the proposed generator generates only one car (c) and the quality of the car is further improved by the proposed discriminator (e).       
}
\label{fig:gendis}
\end{figure*}

\begin{figure*}[t]
\centering
\subfigure[Label]{
\begin{minipage}[t]{0.19\linewidth}
\centering
\includegraphics[width=\linewidth,height=1.8cm]{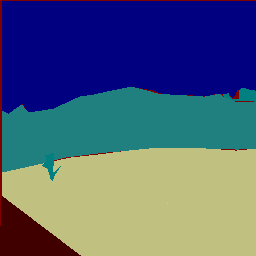}\\
\includegraphics[width=\linewidth,height=1.8cm]{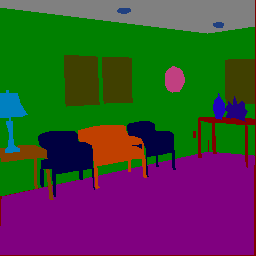}\\
\includegraphics[width=\linewidth,height=1.8cm]{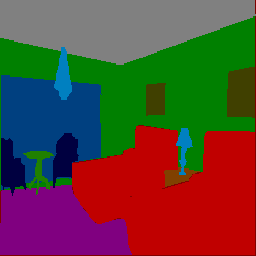}\\
\end{minipage}
} \hspace*{-3mm}
\subfigure[Ground Truth]{
\begin{minipage}[t]{0.19\linewidth}
\centering
\includegraphics[width=\linewidth,height=1.8cm]{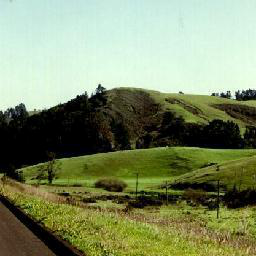}\\
\includegraphics[width=\linewidth,height=1.8cm]{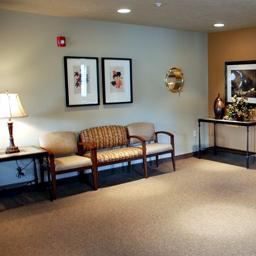}\\
\includegraphics[width=\linewidth,height=1.8cm]{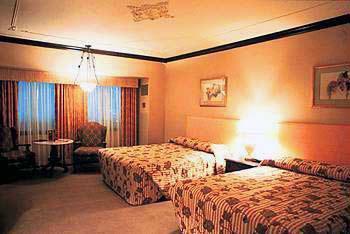}\\
\end{minipage}
} \hspace*{-3mm}
\subfigure[SPADE]{
\begin{minipage}[t]{0.19\linewidth}
\centering
\includegraphics[width=\linewidth,height=1.8cm]{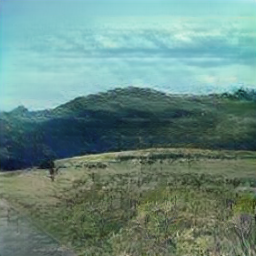}\\
\includegraphics[width=\linewidth,height=1.8cm]{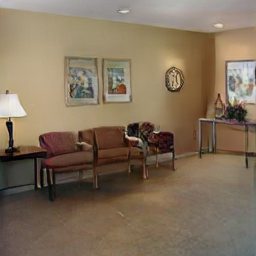}\\
\includegraphics[width=\linewidth,height=1.8cm]{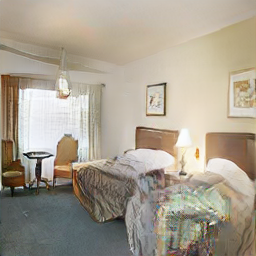}\\
\end{minipage}
} \hspace*{-3mm}
\subfigure[OASIS]{
\begin{minipage}[t]{0.19\linewidth}
\centering
\includegraphics[width=\linewidth,height=1.8cm]{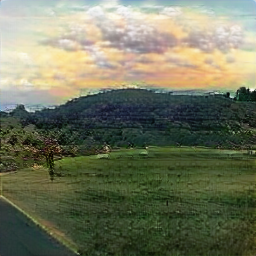}\\
\includegraphics[width=\linewidth,height=1.8cm]{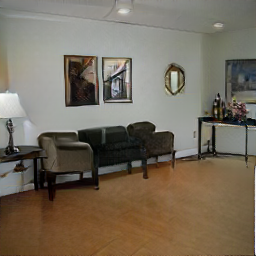}\\
\includegraphics[width=\linewidth,height=1.8cm]{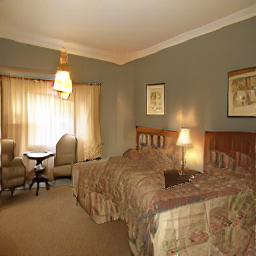}\\
\end{minipage}
} \hspace*{-3mm}
\subfigure[DP-GAN]{
\begin{minipage}[t]{0.19\linewidth}
\centering
\includegraphics[width=\linewidth,height=1.8cm]{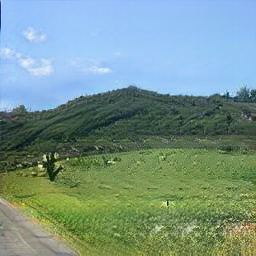}\\
\includegraphics[width=\linewidth,height=1.8cm]{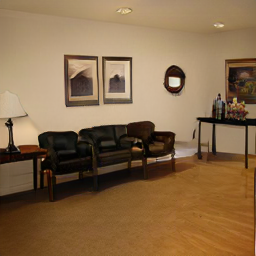}\\
\includegraphics[width=\linewidth,height=1.8cm]{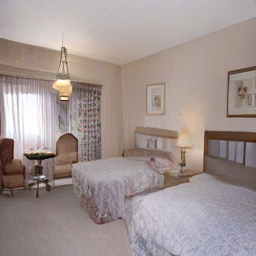}\\
\end{minipage}
} \hspace*{-3mm}
\caption{Qualitative results for ADE20K. Our approach performs well for indoor and outdoor images. SPADE and OASIS struggle to generate multiple instances of objects, like one of the chairs in row 2 is either of poor quality or in the wrong direction, and the two beds in row 3 are connected. In contrast, our approach generates in all cases realistic images.     
}
\label{fig:qvis_ade20k}
\vspace{-5mm}
\end{figure*}

\textbf{Evaluation Metrics.}
We use Fr\'{e}chet Inception Distance (FID) \cite{heusel2017gans} and mean  Intersection-over-Union (mIoU) for evaluation as previous works \cite{park2019semantic,schonfeld_sushko_iclr2021}. mIoU is used to measure the semantic consistency between generated and real images via a pre-trained semantic segmentation network. As in \cite{park2019semantic,schonfeld_sushko_iclr2021}, UperNet101 \cite{xiao2018unified,xiao20upernet} is used for ADE20K and DRN \cite{yu2017dilated,fyu2019drn} for Cityscapes.
FID \cite{heusel2017gans} is widely used to measure the quality of generated images and we use the implementation from \cite{schonfeld_sushko_iclr2021}.

\subsection{Comparison to State-of-the-Art}
We compare our method to state-of-the-art methods in \tabref{tab:comparasion_stoa}. Our method (DP-GAN) achieves the best performance, measured by FID and mIoU, on all datasets. On Cityscapes, our method outperforms previous state-of-the-art methods by -3.6 FID and +4.3 mIoU. On the ADE20K dataset, our approach achieves an improvement by -2.2 FID and +3.9 mIoU. Only on the ADE20K-outdoor dataset, the mIoU of our approach is the same as OASIS \cite{schonfeld_sushko_iclr2021}. This is not surprising since the outdoor images contain less objects as shown in \figref{fig:qvis_ade20k} and our goal is to generate photo-realistic objects in complex and cluttered scenes. Nevertheless, our approach achieves -2.8 FID compared to OASIS and thus generates more realistic images. In order to analyze the effect on object classes and other semantic classes more in detail, we report the per-class IoU for Cityscapes in \tabref{tab:miou}. Except for the class fence, the IoU for non-object classes is similar to other approaches, but our approach achieves a much higher IoU for object classes. In particular for the small object classes pole, traffic light, and sign and the large object classes truck, bus, and train, the improvements in IoU are by a very large margin. For all object classes, the mIoU is improved by +5.4, which is higher than the improvement for all classes.

We show some qualitative results in \figref{fig:motivation} and \ref{fig:qvis_ade20k}. We observe that objects are often not well generated by SPADE and the clear boundaries of semantic classes often disappear in particular for Cityscapes. The quality of the images that are generated by OASIS is much better, but a common artifact is the merging of multiple instances like parking cars or two beds that are close. Our approach does not suffer from this issue and generates realistic looking objects. 

While in the previous experiments we used the same noise tensor $z$ in all experiments, our approach can generate diverse images for the same input label map by generating different random noise tensors $z$ as shown in \figref{fig:qvis_multi_modal}. For a quantitative comparison with OASIS, which generates diverse images in the same way, we report mean and variance of FID over 5 runs with different noise tensors. The results are reported in \tabref{tab:multi_modal}. 

\begin{table*}[h]
    \centering
    \label{tab:ablation}
    \subfigure[Gen / Dis\label{tab:ablationa}]{
    \begin{minipage}[t]{0.32\linewidth}
    \centering
    \setlength{\tabcolsep}{1mm}
    \resizebox{\linewidth}{!}{
    \begin{tabular}{c|c|c|c|c}
    \hline
    Gen & Dis & FID & mIoU & obj-mIoU\\
    \hline
    OA & OA & 47.7 & 69.3 & 61.5\\
    OA & DP & 47.9 & \textbf{74.0} & \textbf{67.4}\\
    DP & OA & 45.4 & 69.9 & 62.0 \\
    \hline
    DP & DP & \textbf{44.1} & 73.6 & 66.9\\
    \hline    
    \end{tabular} 
    }
    \vspace{-2mm}
    \label{tab:abltionA}
    \end{minipage}}
    \subfigure[$\mathcal{L}_{ms}$ in \eqref{eq:ms}\label{tab:ablationb}]{
    \begin{minipage}[t]{0.32\linewidth}
    \centering
        \setlength{\tabcolsep}{1mm}
    \resizebox{\linewidth}{!}{
    \begin{tabular}{c|c|c|c|c}
    \hline
    Enc & Dec & FID & mIoU & obj-mIoU\\
    \hline
    & & 49.2 & 67.9 & 59.5\\
    & \checkmark & 44.5 & 72.1 & 64.4 \\
    \checkmark & \checkmark & 44.3 & 72.8 & 66.4\\ 
    \hline
    \checkmark & & \textbf{44.1} & \textbf{73.6} & \textbf{66.9} \\
    \hline    
    \end{tabular}}
    \vspace{-2mm}
    \label{tab:abltionB}
    \end{minipage}}
    \subfigure[$\mathcal{L}_{fm}$ in \eqref{eq:fm}\label{tab:ablationc}]{
    \begin{minipage}[t]{0.32\linewidth}
    \centering
        \setlength{\tabcolsep}{1mm}
    \resizebox{\linewidth}{!}{
    \begin{tabular}{c|c|c|c|c}
    \hline
    Enc & Dec & FID & mIoU & obj-mIoU\\
    \hline
    & & \textbf{44.1} & 69.9 & 62.1\\
    \checkmark &  & 44.4 & 69.2 & 60.8\\
    \checkmark & \checkmark & 45.0 & \textbf{73.8} & 66.8 \\
    \hline
    & \checkmark & \textbf{44.1} & 73.6 & \textbf{66.9} \\
    \hline    
    \end{tabular}}
    \vspace{-2mm}
    \label{tab:abltionC}
    \end{minipage}}
    \vspace{2mm}
    \caption{Ablation study. `OA' denotes OASIS \cite{schonfeld_sushko_iclr2021} whereas `DP' denotes our method (DP-GAN). Using `OA' as generator corresponds to a generator with a single pyramid.}
    \label{tab:abltion}
    \vspace{-4mm}
\end{table*}

\subsection{Ablation Study}
We analyze the impact of the proposed generator and discriminator in Table~\ref{tab:abltion}(a). To this end, we combine the proposed generator and discriminator with the generator and discriminator of OASIS \cite{schonfeld_sushko_iclr2021}. Replacing the discriminator of OASIS by our discriminator substantially improves mIoU by +4.7. For the object classes, mIoU even improves by +5.9. FID, however, does not significantly change. In contrast, replacing the generator of OASIS with our generator substantially decreases FID by -2.3, but increases mIoU only slightly. This shows that the combination of the proposed generator and discriminator is essential to achieve a low FID and a high mIoU. \figref{fig:gendis} illustrates the effect of the generator and discriminator on the generated images. 

Next, we evaluate the impact of multi-scale adversarial supervision ($\mathcal{L}_{ms}$) in Table~\ref{tab:abltion}(b). The first row shows the result without the loss $\mathcal{L}_{ms}$. For the other rows, the loss is either added to the encoder or decoder of the discriminator or to both. Adding $\mathcal{L}_{ms}$ improves FID and mIoU. This shows that adding $\mathcal{L}_{ms}$ to a segmentation-based discriminator improves the quality of the generated images substantially. The differences of adding the loss to the encoder, decoder, or both are smaller, but using it only for the encoder performs best.           

Table~\ref{tab:abltion}(c) evaluates the impact of using feature alignment ($\mathcal{L}_{fm}$). 
While it does not improve the FID, the mIoU improves by +3.7 if the feature matching loss is added to the decoder of the discriminator. In contrast, adding the feature matching loss to the encoder decreases the performance. This result is interesting since it has been reported in \cite{schonfeld_sushko_iclr2021} that adding a feature matching loss to a segmentation-based discriminator decreases the performance. Our results, however, show that it is important where the feature matching loss is added. For the decoder, it actually increases the performance. For more qualitative results and ablation studies, we refer to the supplemental material.



\section{Conclusion}
\label{sec:conclusion}

In this paper, we have addressed the problem that previous methods for semantic image synthesis struggle to generate photo-realistic objects at different scales in complex scenes. To this end, we proposed a dual pyramid architecture for the generator that adapts the conditioning depending on the object size. As a second contribution, we proposed to unify multi-scale supervision within a single framework. The proposed approach outperforms the state-of-the-art for semantic image synthesis by a large margin. While the quality of the generated images has been substantially improved compared to the state-of-the-art, the qualitative results contain still some artifacts. Although semantic image synthesis is important for image generation and editing tasks and applications like autonomous driving in order to generate additional training data for rare situations, such technology can be potentially misused for disinformation and propaganda.  

\paragraph{Acknowledgement}
The work has been supported by the Deutsche Forschungsgemeinschaft (DFG, German Research Foundation) under Germany’s Excellence Strategy - EXC 2070 - 390732324 and GA 1927/5-2 (FOR 2535 Anticipating Human Behavior).


\bibliography{egbib}
\end{document}


\maketitle

\begin{table*}[t]
    \centering
    \label{tab:ablation}
    \begin{minipage}[t]{0.54\linewidth}
    \centering
    \setlength{\tabcolsep}{2mm}
    \resizebox{\linewidth}{!}{
    \begin{tabular}{c|c|ccc}
    \hline
        & all & large & medium & small \\
        \hline
        SPADE & 33.4 & 84.3 & 63.3 & 36.8 \\
        OASIS & 35.0 & 98.6 & 79.9 & 37.6 \\
        \hline
        DP-GAN & 24.5 & 77.6 & 57.5 & 27.1 \\
        \hline
    \end{tabular}}
    \vspace{2mm}
    \caption{FID for cropped objects from the Cityscapes dataset.}
    \label{tab:ms_object}
    \end{minipage}
    \begin{minipage}[t]{0.45\linewidth}
    \centering
    \setlength{\tabcolsep}{2mm}
    \resizebox{\linewidth}{!}{
    \begin{tabular}{c|ccc}
    \hline
        & $0.5\times$ & $1.0\times$ & $2.0\times$ \\
        \hline
        SPADE & 99.9 &  71.8 & 145.9 \\
        OASIS & 91.8 &  47.7 & 115.5 \\
        \hline
        DP-GAN & 85.0 &  44.1 & 96.1 \\
        \hline
    \end{tabular}}
    \vspace{2mm}
    \caption{FID for different image resolutions of the Cityscapes dataset.}
    \label{tab:ms_scene}
    \end{minipage}
\end{table*}

\begin{table}[t]
    \centering
    \begin{tabular}{c|c|c|c}
    \hline
         & w/o $\mathcal{L}_{LM}$ & w/o cat & Ours\\         
         \hline
         FID & 46.8 & 44.8 & \textbf{44.1}\\
         mIoU & 70.9 & 71.2 & \textbf{73.6} \\
         \hline
    \end{tabular}
    \vspace{2mm}
    \caption{Ablation study on Cityscapes. ``w/o $LM$" means without LabelMix regularization $\mathcal{L}_{LM}$ in (9). ``w/o cat" means without concatenating feature map $\alpha_0$ in (2).}
    \label{tab:added}
\end{table}

\begin{table}[t]
    \centering
    \begin{tabular}{cccc}
    \hline
     & SPADE & OASIS & DP-GAN \\
    \hline
    Preferred image (\%) & 13.6 & 24.2 & 62.2 \\
    \hline
    \end{tabular}
    \vspace{2mm}
    \caption{User study. In 62.2\% of the cases, the users considered the image generated by DP-GAN more realistic than the images generated by SPADE or OASIS.}
    \label{tab:up}
\end{table}

\begin{table}[t]
    \centering
    \renewcommand{\tabcolsep}{3.0mm}
    \begin{tabular}{c|c|c}
         & \#Parameters (M) & Inference time (ms) \\
         \hline
        LGGAN  & 111.1 & 222.7 \\
        DAGAN  & 93.1 & 63.8\\
        CC-FPSE  & 128.1 & 130.1 \\
        SPADE  & 93.0 & 70.8\\
        OASIS  & 71.1 & 60.5\\
        \hline
        Ours & \textbf{69.5} & \textbf{51.8} \\
    \end{tabular}
        \caption{Comparison of number of parameters and inference time.}
    \label{tab:efficiency}
\end{table}

\begin{table}[!h]
    \centering
\resizebox{0.8\linewidth}{!}{
    \begin{tabular}{c|cc|cc}
Operation & Input & Size & Output & Size \\ \hline \hline
\multirow{2}{*}{Concatenate} & $z$ & (64,256,256) & \multirow{2}{*}{$z$\_$y$} & \multirow{2}{*}{(64+$N$,256,256)} \\
& $y$ & ($N$,256,256) \\ \hline
ConvBlock & $z\_y$ &  (64+$N$,256,256) &  $s$ &  (32, 256, 256) \\ \hline
ConvBlock & $s$ &  (32, 256, 256) &  $s_5$ &  (64, 256, 256) \\ \hline
ConvBlock & $s_5$ &  (64, 256, 256) &  $s_4$ &  (64, 128, 128) \\ \hline
ConvBlock & $s_4$ &  (64, 128, 128) &  $s_3$ &  (64, 64, 64) \\ \hline
ConvBlock & $s_3$ &  (64, 64, 64) &  $s_2$ &  (64, 32, 32) \\ \hline
ConvBlock & $s_2$ &  (64, 32, 32) &  $s_1$ &  (64, 16, 16) \\ \hline
ConvBlock & $s_1$ &  (64, 16, 16) &  $s_0$ &  (64, 8, 8) \\ \hline
    \end{tabular}}
    \caption{Spatial adaptation learning pyramid of the generator. $N$ refers to the number of semantic classes, $z$ is noise sampled from a unit Gaussian, $y$ is the label map, and ConvBlock denotes Conv2d-BatchNorm2d-ReLU block.}
    \label{tab:generator_sap}
\end{table}

\begin{table}[t]
    \centering
    \resizebox{\linewidth}{!}{
    \begin{tabular}{c|cc|cc|c}
        Operation & Input & Size & Output & Size & Sup \\ \hline \hline
        ResBlock-Down & image & (3,256,256) & $down_1$ & (128,128,128) \\ \hline
        ResBlock-Down & $down_1$ & (128,128,128) & $down_2$ &         (128,64,64) \\ \hline
        ResBlock-Down & $down_2$ & (128,64,64) & $down_3$   &         (256,32,32) \\ \hline
        ResBlock-Down & $down_3$ & (256,32,32) & $down_4$   &         (256,16,16) & $\mathcal{L}_{patch}$\\ \hline
        ResBlock-Down & $down_4$ & (256,16,16) & $down_5$   & (512,8,8)         \\ \hline
        ResBlock-Down & $down_5$ & (512,8,8)   & $down_6$   & (512,4,4) & $\mathcal{L}_{patch}$ \\ \hline
        ResBlock-Up   & $down_6$ & (512,4,4)   & $up_1$ & (512,8,8) &  \\ \hline
        ResBlock-Up  & cat($up_1$, $down_5$) & (1024,8,8) & $up_2$ & (256,16,16)& $\mathcal{L}_{feature}$\\ \hline
        ResBlock-Up   & cat($up_2$, $down_4$) & (512,16,16) & $up_3$ & (256,32,32) & $\mathcal{L}_{feature}$\\ \hline
        ResBlock-Up   & cat($up_3$, $down_3$) &  (512,32,32) & $up_4$ & (128,64,64) & $\mathcal{L}_{feature}$\\ \hline
        ResBlock-Up   & cat($up_4$, $down_2$) & (256,64,64) & $up_5$ & (128,128,128) & $\mathcal{L}_{feature}$ \\ \hline
        ResBlock-Up   & cat($up_5$, $down_1$) & (256,128,128) & $up_6$ & (64,256,256) & $\mathcal{L}_{feature}$ \\ \hline
        Conv2D & $up_6$ & (64,256,256) & out & ($N$+1,256,256) & $\mathcal{L}_{pixel}$ \\ \hline
    \end{tabular}}
    \caption{Discriminator. $N$ refers to the number of semantic classes. $\mathcal{L}_{pixel}$, $\mathcal{L}_{patch}$, $\mathcal{L}_{feature}$ correspond to pixel, patch and feature supervision, respectively.}
    \label{tab:discriminator}
\end{table}

\section{Additional Ablation Studies}
\begin{figure*}[b]
\centering
\subfigure[Large]{
\begin{minipage}[t]{\linewidth}
\includegraphics[width=0.16\linewidth,height=1.5cm]{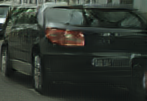}\vspace{-1mm}
\includegraphics[width=0.16\linewidth,height=1.5cm]{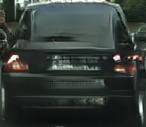}\vspace{-1mm}
\includegraphics[width=0.16\linewidth,height=1.5cm]{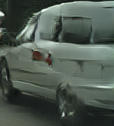} 
\includegraphics[width=0.001\linewidth, height=1.5cm]{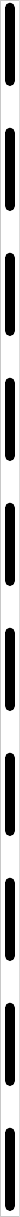}
\includegraphics[width=0.16\linewidth,height=1.5cm]{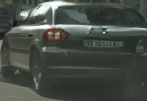}\vspace{-1mm} 
\includegraphics[width=0.16\linewidth,height=1.5cm]{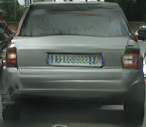} \vspace{-1mm}
\includegraphics[width=0.16\linewidth,height=1.5cm]{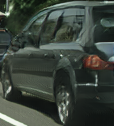}
\end{minipage}
} \\ \vspace{-2mm}
\subfigure[Medium]{
\begin{minipage}[t]{\linewidth}
\includegraphics[width=0.117\linewidth,height=1.4cm]{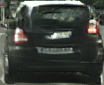}\vspace{-1mm}
\includegraphics[width=0.117\linewidth,height=1.4cm]{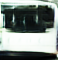}\vspace{-1mm}
\includegraphics[width=0.117\linewidth,height=1.4cm]{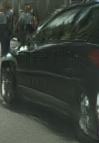}\vspace{-1mm}
\includegraphics[width=0.117\linewidth,height=1.4cm]{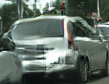}
\includegraphics[width=0.001\linewidth, height=1.5cm]{figs/line.pdf}
\includegraphics[width=0.117\linewidth,height=1.4cm]{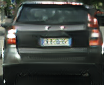}\vspace{-1mm}
\includegraphics[width=0.117\linewidth,height=1.4cm]{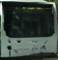}\vspace{-1mm}
\includegraphics[width=0.117\linewidth,height=1.4cm]{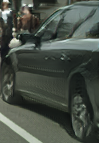}\vspace{-1mm}
\includegraphics[width=0.117\linewidth,height=1.4cm]{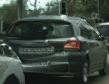}\vspace{-1mm}
\centering
\end{minipage}} \\ \vspace{-2mm}
\subfigure[Small]{
\begin{minipage}[t]{\linewidth}
\includegraphics[width=0.093\linewidth,height=1.3cm]{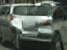}\vspace{-1mm}
\includegraphics[width=0.093\linewidth,height=1.3cm]{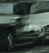}\vspace{-1mm}
\includegraphics[width=0.093\linewidth,height=1.3cm]{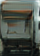}\vspace{-1mm}
\includegraphics[width=0.093\linewidth,height=1.3cm]{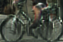}
\includegraphics[width=0.093\linewidth,height=1.3cm]{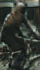}\vspace{-1mm}
\includegraphics[width=0.001\linewidth, height=1.5cm]{figs/line.pdf}
\includegraphics[width=0.093\linewidth,height=1.3cm]{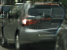}\vspace{-1mm}
\includegraphics[width=0.093\linewidth,height=1.3cm]{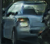}\vspace{-1mm}
\includegraphics[width=0.093\linewidth,height=1.3cm]{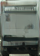}\vspace{-1mm}
\includegraphics[width=0.093\linewidth,height=1.3cm]{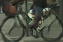}\vspace{-1mm}
\includegraphics[width=0.093\linewidth,height=1.3cm]{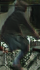}\vspace{-1mm}
\end{minipage}}
\caption{Generated objects of different sizes. The objects on the left hand side are generated by OASIS \cite{schonfeld_sushko_iclr2021} whereas the objects on the right hand side are generated by our method. Our method generates more realistic objects for all sizes.}
\label{fig:multi_scale}
\vspace{-5mm}
\end{figure*}

\paragraph{Quality of the generated objects}
In order to analyze the quality of the generated objects, we crop the instances of the objects from the generated images. The bounding boxes are obtained by the label maps. \figref{fig:multi_scale} shows examples of such crops. From these examples, we can see that our method generates more realistic objects at all sizes. We evaluate the quality of the objects by computing the FID score for the cropped objects instead of the images. The results are shown in \tabref{tab:ms_object}. The FID score for SPADE is slightly lower than for OASIS although the generated objects do not look better. Nevertheless, the FID score of our approach is much lower.   
We furthermore split the objects into large, medium, and small objects based on the size of the label mask. There are 69 large objects ($>$ 10000 pixels), 378 medium objects, and 3190 small objects ($<$ 2500 pixels). Our approach performs well for objects of all sizes. Note that the FID scores between the columns are not comparable since each set (all, large, medium, small) contains a different number of objects. 
We furthermore report the FID score on Cityscapes if we increase or decrease the image resolution by factor 2 in \tabref{tab:ms_scene}. Our approach also performs better for different image resolutions.

\paragraph{Impact of feature map concatenation and LabelMix}
Finally, \tabref{tab:added} shows
the results when we do not concatenate (w/o cat) the feature map $\alpha_0$ in (2). We can see that both FID and mIoU are worse.  
For completeness, we also report the impact of $\mathcal{L}_{LM}$. Without $\mathcal{L}_{LM}$ (w/o $\mathcal{L}_{LM}$) FID and mIoU are worse, which confirms the effectiveness of the LabelMix regularization.    

\paragraph{User study}
We further evaluated the quality of the generated images by a user study. To this end, we showed the participants three images in randomized order that have been generated for the same label map by SPADE, OASIS, and DP-GAN, respectively. The label map itself was not shown. The participants needed then to select the image among the three images which looked most realistic. For the user study, we used all 500 label maps of the validation set of Cityscapes. The label maps were equally assigned to 10 participants such that each participant rated 50 label maps. The results of the user study are shown in \tabref{tab:up}. The participants selected in more than 60\% of the cases, the image that has been generated by our method.

\paragraph{Efficiency Analysis}
We report the model size and time for inference in \tabref{tab:efficiency} and compare it to other approaches. The inference time has been measured for a single TITAN RTX. Compared to OASIS, the inference time is reduced by more than 14\%.
This shows that our method outperforms the state-of-the-art not only in terms of image generation quality, but it is also more efficient.

\paragraph{Multi-Modal Image Synthesis}

\begin{figure*}[t]
\centering
\subfigure[Label]{
\begin{minipage}[t]{0.15\linewidth}
\centering
\includegraphics[width=\linewidth,height=2.0cm]{figs/vis_multi_modal/label/ADE_val_00000023.png}\\
\includegraphics[width=\linewidth,height=2.0cm]{figs/vis_multi_modal/label/ADE_val_00000013.png}\\
\includegraphics[width=\linewidth,height=2.0cm]{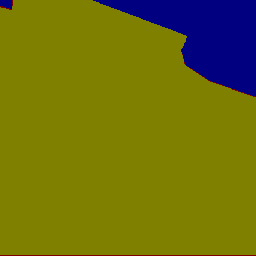}\\
\includegraphics[width=\linewidth,height=2.0cm]{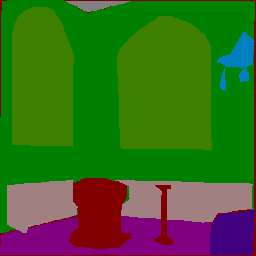}\\
\end{minipage}
} \hspace{-3mm}
\subfigure[GT]{
\begin{minipage}[t]{0.15\linewidth}
\centering
\includegraphics[width=\linewidth,height=2.0cm]{figs/vis_multi_modal/gt/ADE_val_00000023.jpg}\\
\includegraphics[width=\linewidth,height=2.0cm]{figs/vis_multi_modal/gt/ADE_val_00000013.jpg}\\
\includegraphics[width=\linewidth,height=2.0cm]{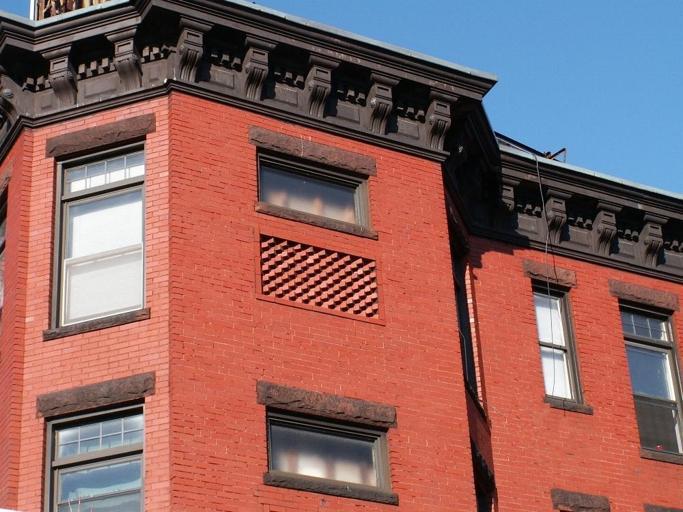}\\
\includegraphics[width=\linewidth,height=2.0cm]{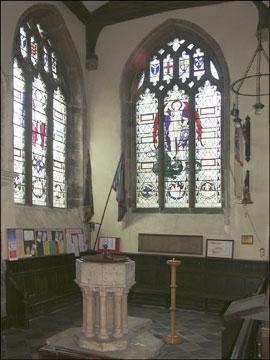}\\
\end{minipage}
} \hspace{-3mm}
\subfigure[1]{
\begin{minipage}[t]{0.15\linewidth}
\centering
\includegraphics[width=\linewidth,height=2.0cm]{figs/vis_multi_modal/4/ADE_val_00000023.png}\\
\includegraphics[width=\linewidth,height=2.0cm]{figs/vis_multi_modal/5/ADE_val_00000013.png}\\
\includegraphics[width=\linewidth,height=2.0cm]{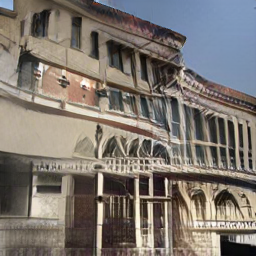}\\
\includegraphics[width=\linewidth,height=2.0cm]{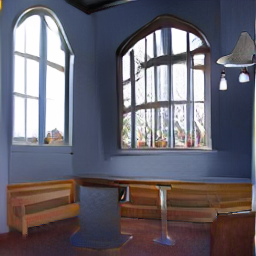}\\
\end{minipage}
} \hspace{-3mm}
\subfigure[2]{
\begin{minipage}[t]{0.15\linewidth}
\centering
\includegraphics[width=\linewidth,height=2.0cm]{figs/vis_multi_modal/3/ADE_val_00000023.png}\\
\includegraphics[width=\linewidth,height=2.0cm]{figs/vis_multi_modal/6/ADE_val_00000013.png}\\
\includegraphics[width=\linewidth,height=2.0cm]{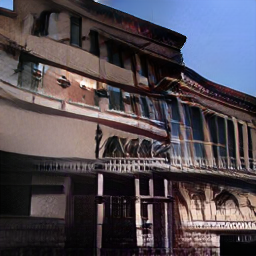}\\
\includegraphics[width=\linewidth,height=2.0cm]{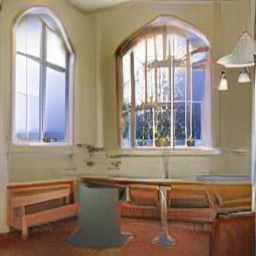}\\
\end{minipage}
} \hspace{-3mm}
\subfigure[3]{
\begin{minipage}[t]{0.15\linewidth}
\centering
\includegraphics[width=\linewidth,height=2.0cm]{figs/vis_multi_modal/2/ADE_val_00000023.png}\\
\includegraphics[width=\linewidth,height=2.0cm]{figs/vis_multi_modal/7/ADE_val_00000013.png}\\
\includegraphics[width=\linewidth,height=2.0cm]{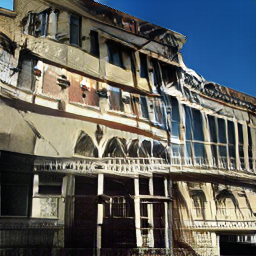}\\
\includegraphics[width=\linewidth,height=2.0cm]{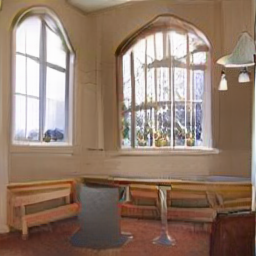}\\
\end{minipage}
} \hspace{-3mm}
\subfigure[4]{
\begin{minipage}[t]{0.15\linewidth}
\centering
\includegraphics[width=\linewidth,height=2.0cm]{figs/vis_multi_modal/1/ADE_val_00000023.png}\\
\includegraphics[width=\linewidth,height=2.0cm]{figs/vis_multi_modal/1/ADE_val_00000013.png}\\
\includegraphics[width=\linewidth,height=2.0cm]{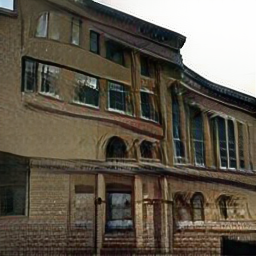}\\
\includegraphics[width=\linewidth,height=2.0cm]{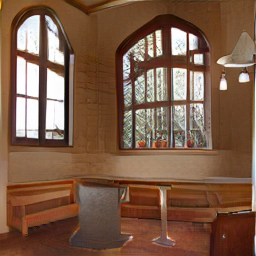}\\
\end{minipage}
}
\caption{Images generated with four different random noise tensors $z$.}
\label{fig:qvis_multi_modal}
\vspace{-5mm}
\end{figure*}

\figref{fig:qvis_multi_modal} shows examples where we generate four images from the same label map with different styles.
This is achieved by randomly sampling a 3D noise tensor. Our method can produce diverse high quality images for both indoor and outdoor scenarios. The color, texture, and illumination vary, but the semantic structure is maintained which is desired.

\section{Details of Architecture and Training}

The details of the two pyramids of the generator are shown in \tabref{tab:generator_sap} and \tabref{tab:generator_isp}, respectively. The discriminator is shown in \tabref{tab:discriminator}. It is a U-Net architecture built from ResNet blocks.
For training, we use the Adam optimizer \cite{kingma2014adam} with momenta $\beta = (0,0.999)$. The 
learning rates for the generator and discriminator are set to 0.0001 and 0.0004, respectively. 
Our method also uses an exponential moving average \cite{brock2018large} for the generator weights with $0.9999$ decay. As in \cite{schonfeld_sushko_iclr2021}, we use $\lambda_{LM}=5$.
All experiments have been conducted on a single TITAN RTX with a fix random seed. The source code is available at \href{https://github.com/sj-li/DP_GAN}{https://github.com/sj-li/DP\_GAN}.



\begin{table}[t]
    \centering
    \renewcommand{\tabcolsep}{3.0mm}
\resizebox{\linewidth}{!}{
    \begin{tabular}{c|cc|cc}
Operation & Input & Size & Output & Size \\ \hline \hline
\multirow{2}{*}{Concatenate} & $z$ & (64,256,256) & \multirow{2}{*}{$z$\_$y$} & \multirow{2}{*}{(64+$N$,256,256)} \\
& $y$ & ($N$,256,256) \\ \hline
Conv2D & $interp$($z$\_$y$) &  (64+$N$,8,8) &  $up_0$ &  (1024,8,8) \\ \hline
\multirow{2}{*}{SPADE-ResBlock} & $up_0$ &  (1024,8,8) & \multirow{2}{*}{$up_1$} & \multirow{2}{*}{(1024,16,16)} \\ & $s_0$ & (64,8,8) & \\ \hline
\multirow{2}{*}{SPADE-ResBlock} & $up_1$ & (1024,16,16) & \multirow{2}{*}{$up_2$} & \multirow{2}{*}{(512,32,32)} \\ 
& $s_1$ & (64,16,16) & \\ \hline
\multirow{2}{*}{SPADE-ResBlock} & $up_2$ & (512,32,32) & \multirow{2}{*}{$up_3$} & \multirow{2}{*}{(256,64,64)} \\ 
& $s_2$ & (64,32,32) \\ \hline
\multirow{2}{*}{SPADE-ResBlock} & $up_3$ & (256,64,64) &
\multirow{2}{*}{$up_4$} & \multirow{2}{*}{(128,128,128)} \\ 
& $s_3$ & (64,64,64) & \\ \hline
\multirow{2}{*}{SPADE-ResBlock} & $up_4$ &  (128,128,128) &
\multirow{2}{*}{$up_5$} & \multirow{2}{*}{(64,256,256)} \\ 
& $s_4$ & (64,128,128) \\ \hline
Conv2D, LeakyRelu, TanH & $up_5$ & (64,256,256)&  $x$ & (3,256,256) \\ \hline
    \end{tabular}}
    \caption{Image synthesis pyramid of the generator. $N$ refers to the number of semantic classes, $z$ is noise sampled from a unit Gaussian, $y$ is the label map, and $interp$ interpolates a given input to the appropriate spatial
dimensions of the current layer. $s_0$-$s_5$ are from the spatial adaptation learning pyramid shown in \tabref{tab:generator_sap}.}
    \label{tab:generator_isp}
\end{table}



\begin{figure*}[htbp]
\centering
\subfigure[Label]{
\begin{minipage}[t]{0.13\linewidth}
\centering
\includegraphics[width=\linewidth]{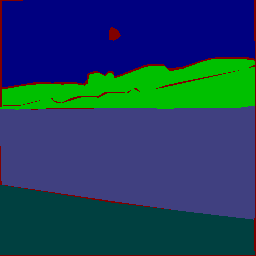}\\
\includegraphics[width=\linewidth]{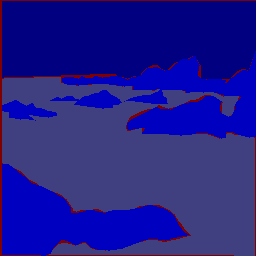}\\
\includegraphics[width=\linewidth]{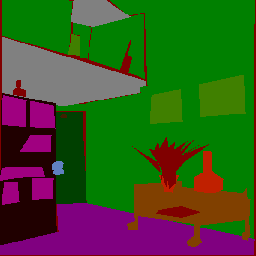}\\
\includegraphics[width=\linewidth]{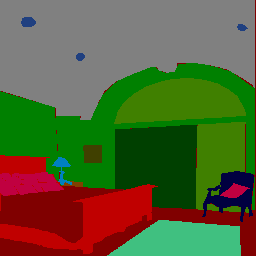}\\
\includegraphics[width=\linewidth]{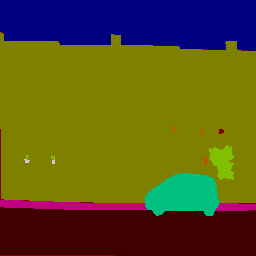}\\
\includegraphics[width=\linewidth]{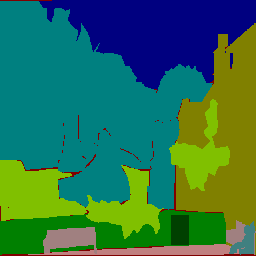}\\
\includegraphics[width=\linewidth]{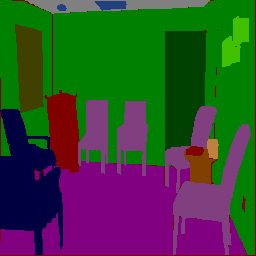}\\
\includegraphics[width=\linewidth]{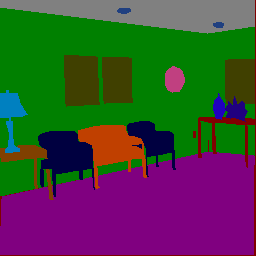}\\
\includegraphics[width=\linewidth]{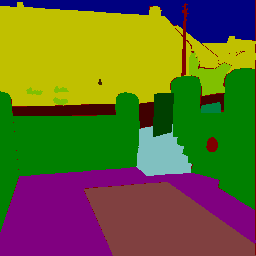}\\
\end{minipage}
}  \hspace{-3mm}
\subfigure[CC-FPSE]{
\begin{minipage}[t]{0.13\linewidth}
\centering
\includegraphics[width=\linewidth]{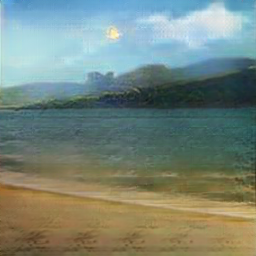}\\
\includegraphics[width=\linewidth]{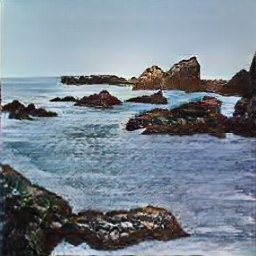}\\
\includegraphics[width=\linewidth]{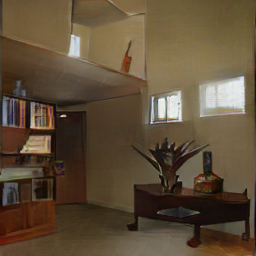}\\
\includegraphics[width=\linewidth]{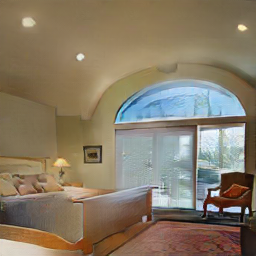}\\
\includegraphics[width=\linewidth]{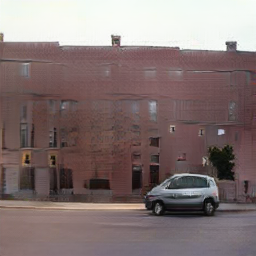}\\
\includegraphics[width=\linewidth]{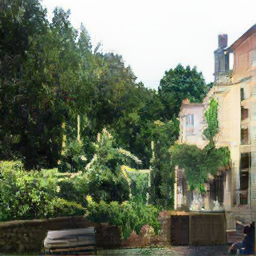}\\
\includegraphics[width=\linewidth]{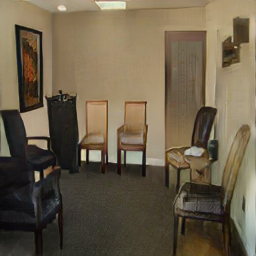}\\
\includegraphics[width=\linewidth]{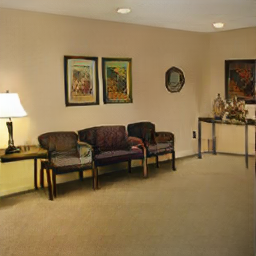}\\
\includegraphics[width=\linewidth]{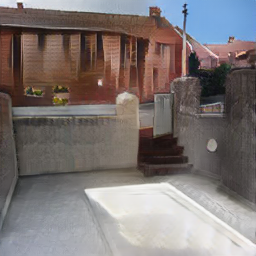}\\
\end{minipage}
} \hspace{-3mm}
\subfigure[DAGAN]{
\begin{minipage}[t]{0.13\linewidth}
\centering
\includegraphics[width=\linewidth]{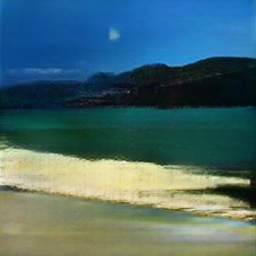}\\
\includegraphics[width=\linewidth]{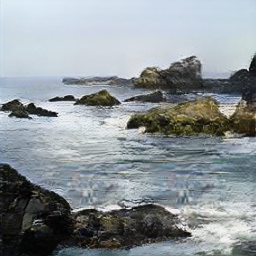}\\
\includegraphics[width=\linewidth]{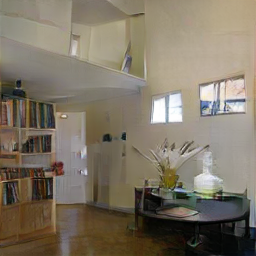}\\
\includegraphics[width=\linewidth]{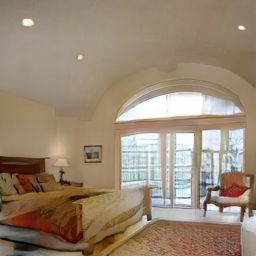}\\
\includegraphics[width=\linewidth]{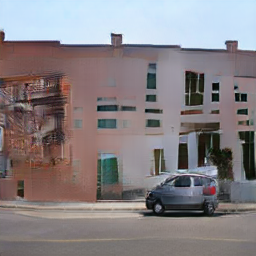}\\
\includegraphics[width=\linewidth]{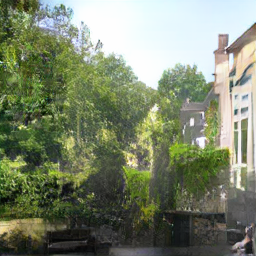}\\
\includegraphics[width=\linewidth]{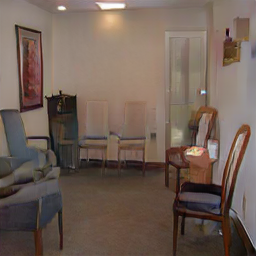}\\
\includegraphics[width=\linewidth]{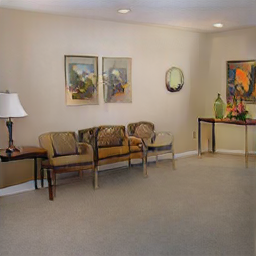}\\
\includegraphics[width=\linewidth]{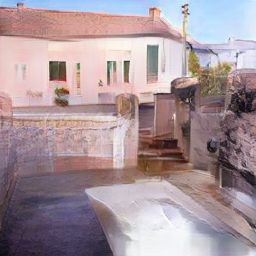}\\
\end{minipage}
} \hspace{-3mm}
\subfigure[LGGAN]{
\begin{minipage}[t]{0.13\linewidth}
\centering
\includegraphics[width=\linewidth]{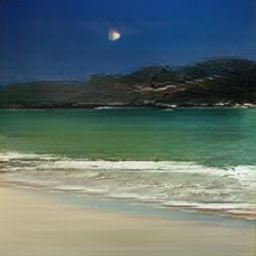}\\
\includegraphics[width=\linewidth]{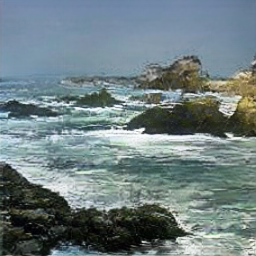}\\
\includegraphics[width=\linewidth]{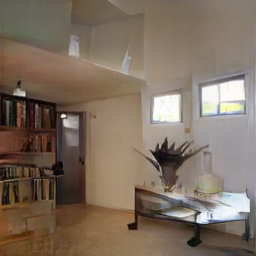}\\
\includegraphics[width=\linewidth]{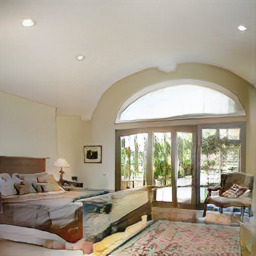}\\
\includegraphics[width=\linewidth]{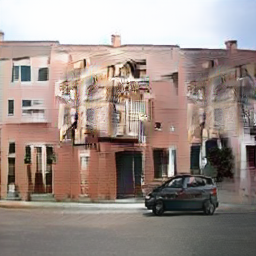}\\
\includegraphics[width=\linewidth]{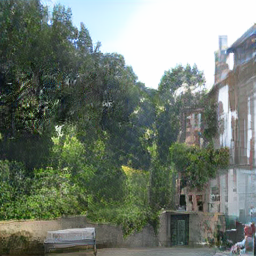}\\
\includegraphics[width=\linewidth]{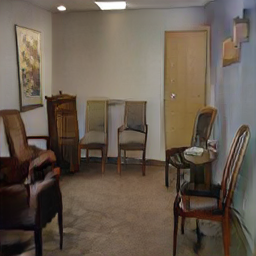}\\
\includegraphics[width=\linewidth]{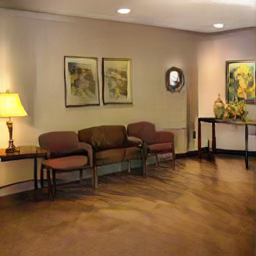}\\
\includegraphics[width=\linewidth]{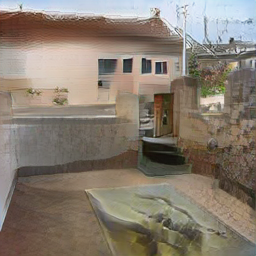}\\
\end{minipage}
} \hspace{-3mm}
\subfigure[SPADE]{
\begin{minipage}[t]{0.13\linewidth}
\centering
\includegraphics[width=\linewidth]{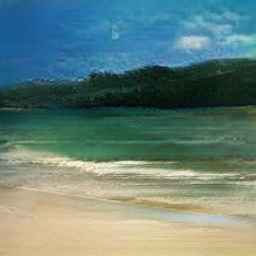}\\
\includegraphics[width=\linewidth]{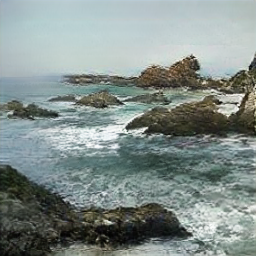}\\
\includegraphics[width=\linewidth]{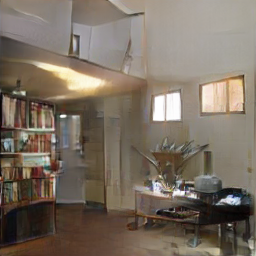}\\
\includegraphics[width=\linewidth]{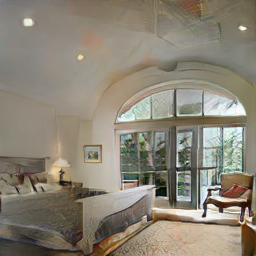}\\
\includegraphics[width=\linewidth]{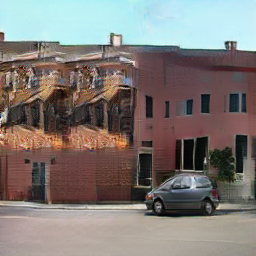}\\
\includegraphics[width=\linewidth]{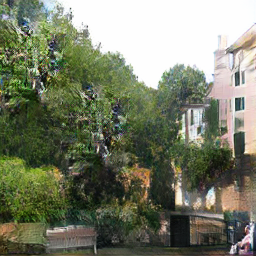}\\
\includegraphics[width=\linewidth]{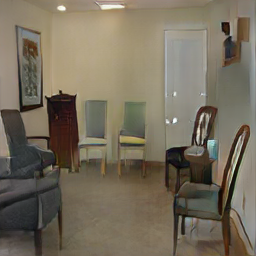}\\
\includegraphics[width=\linewidth]{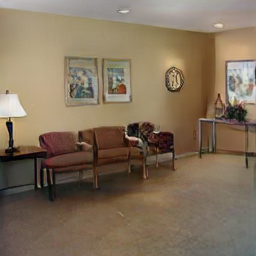}\\
\includegraphics[width=\linewidth]{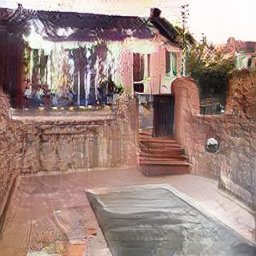}\\
\end{minipage}
} \hspace{-3mm}
\subfigure[OASIS]{
\begin{minipage}[t]{0.13\linewidth}
\centering
\includegraphics[width=\linewidth]{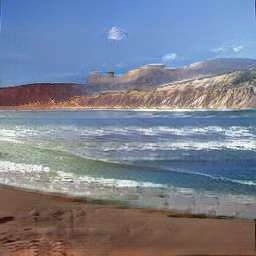}\\
\includegraphics[width=\linewidth]{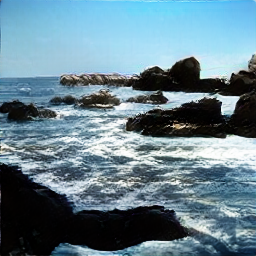}\\
\includegraphics[width=\linewidth]{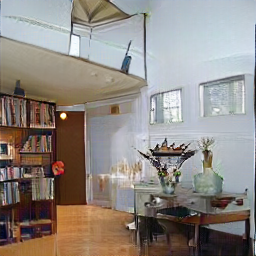}\\
\includegraphics[width=\linewidth]{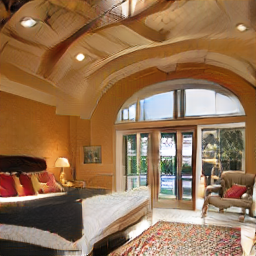}\\
\includegraphics[width=\linewidth]{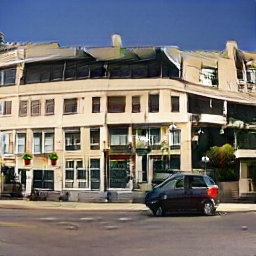}\\
\includegraphics[width=\linewidth]{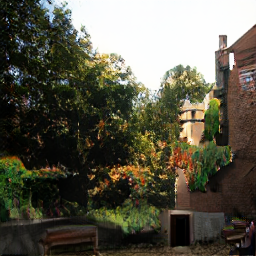}\\
\includegraphics[width=\linewidth]{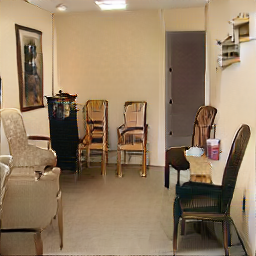}\\
\includegraphics[width=\linewidth]{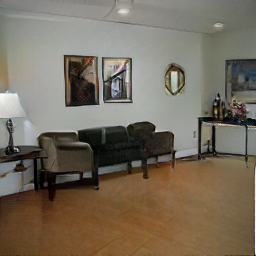}\\
\includegraphics[width=\linewidth]{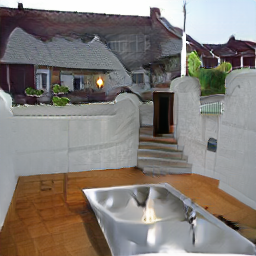}\\
\end{minipage}
} \hspace{-3mm}
\subfigure[Ours]{
\begin{minipage}[t]{0.13\linewidth}
\centering
\includegraphics[width=\linewidth]{figs/vis_all/ade20k/Ours/ADE_val_00001616.png}\\
\includegraphics[width=\linewidth]{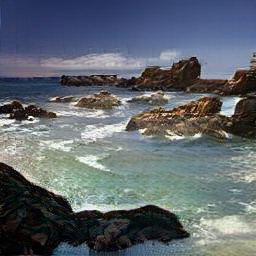}\\
\includegraphics[width=\linewidth]{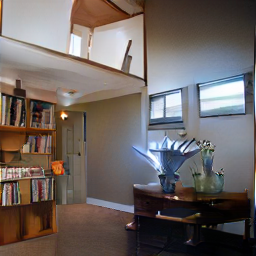}\\
\includegraphics[width=\linewidth]{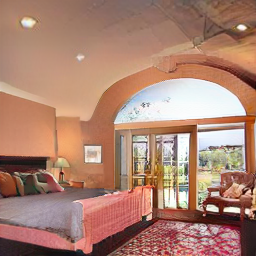}\\
\includegraphics[width=\linewidth]{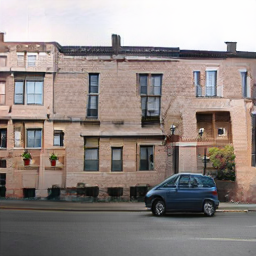}\\
\includegraphics[width=\linewidth]{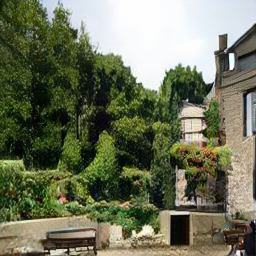}\\
\includegraphics[width=\linewidth]{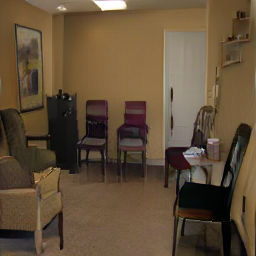}\\
\includegraphics[width=\linewidth]{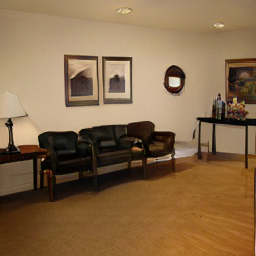}\\
\includegraphics[width=\linewidth]{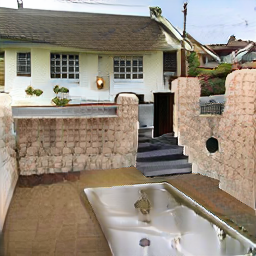}\\
\end{minipage}
} 
\caption{Qualitative results for ADE20K.}
\label{fig:qvis_cityscapes}
\end{figure*}

\begin{figure*}[htbp]
\centering
\subfigure[Label]{
\begin{minipage}[t]{0.13\linewidth}
\centering
\includegraphics[width=\linewidth]{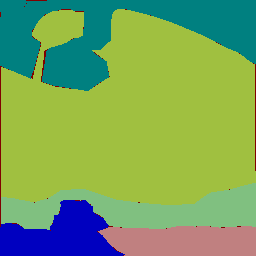}\\
\includegraphics[width=\linewidth]{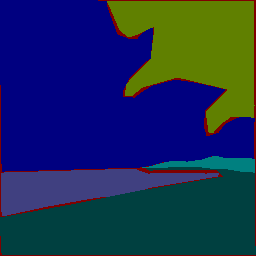}\\
\includegraphics[width=\linewidth]{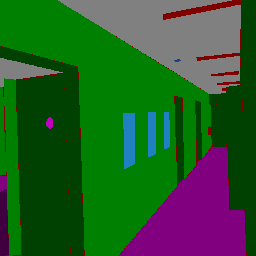}\\
\includegraphics[width=\linewidth]{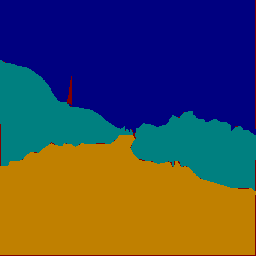}\\
\includegraphics[width=\linewidth]{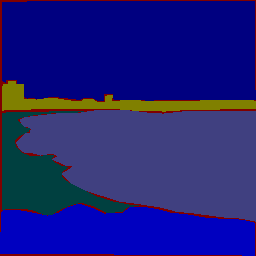}\\
\includegraphics[width=\linewidth]{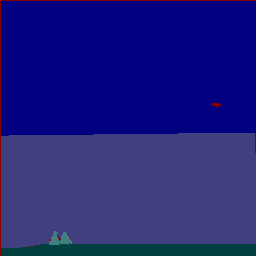}\\
\includegraphics[width=\linewidth]{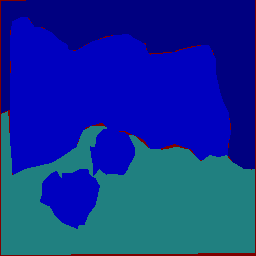}\\
\includegraphics[width=\linewidth]{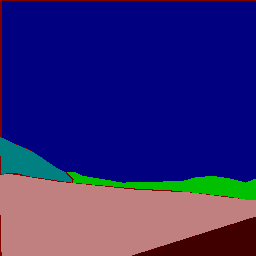}\\
\includegraphics[width=\linewidth]{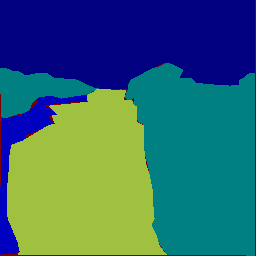}\\
\end{minipage}
}  \hspace{-3mm}
\subfigure[CC-FPSE]{
\begin{minipage}[t]{0.13\linewidth}
\centering
\includegraphics[width=\linewidth]{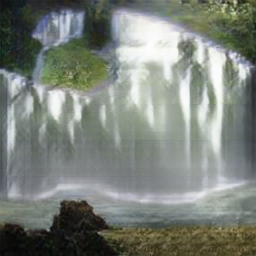}\\
\includegraphics[width=\linewidth]{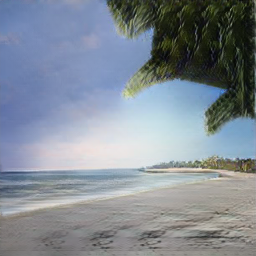}\\
\includegraphics[width=\linewidth]{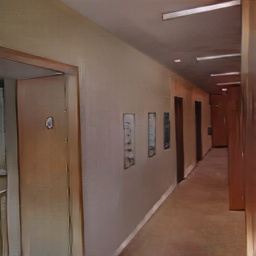}\\
\includegraphics[width=\linewidth]{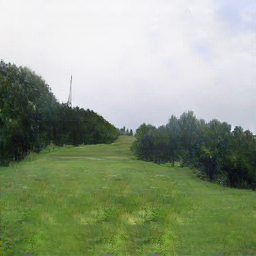}\\
\includegraphics[width=\linewidth]{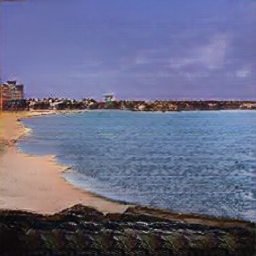}\\
\includegraphics[width=\linewidth]{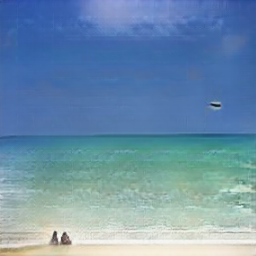}\\
\includegraphics[width=\linewidth]{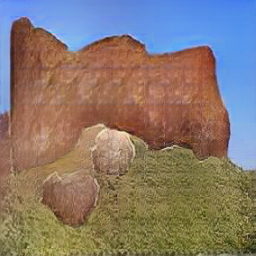}\\
\includegraphics[width=\linewidth]{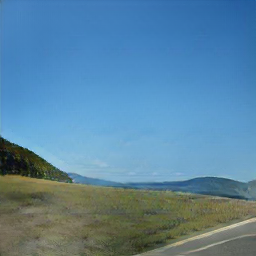}\\
\includegraphics[width=\linewidth]{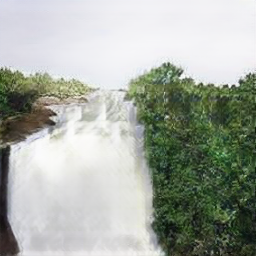}\\
\end{minipage}
} \hspace{-3mm}
\subfigure[DAGAN]{
\begin{minipage}[t]{0.13\linewidth}
\centering
\includegraphics[width=\linewidth]{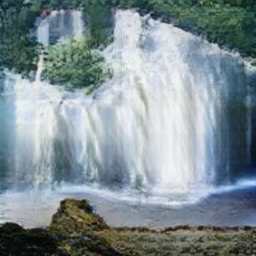}\\
\includegraphics[width=\linewidth]{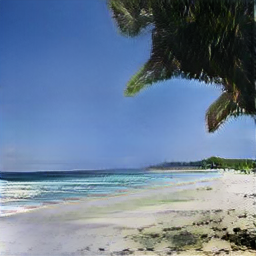}\\
\includegraphics[width=\linewidth]{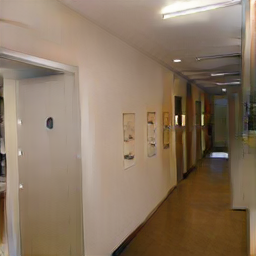}\\
\includegraphics[width=\linewidth]{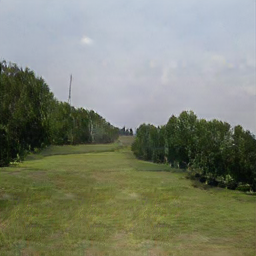}\\
\includegraphics[width=\linewidth]{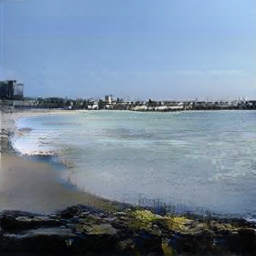}\\
\includegraphics[width=\linewidth]{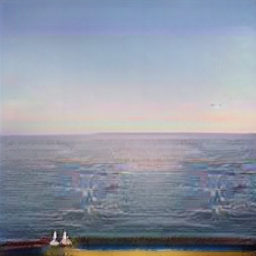}\\
\includegraphics[width=\linewidth]{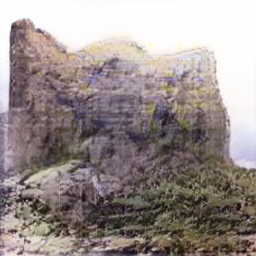}\\
\includegraphics[width=\linewidth]{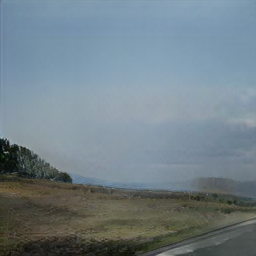}\\
\includegraphics[width=\linewidth]{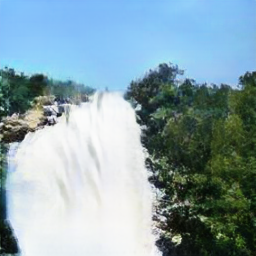}\\
\end{minipage}
} \hspace{-3mm}
\subfigure[LGGAN]{
\begin{minipage}[t]{0.13\linewidth}
\centering
\includegraphics[width=\linewidth]{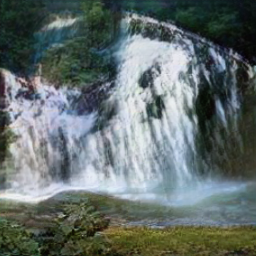}\\
\includegraphics[width=\linewidth]{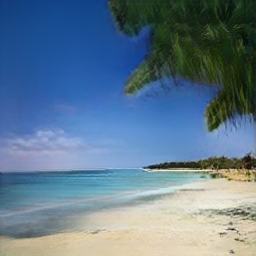}\\
\includegraphics[width=\linewidth]{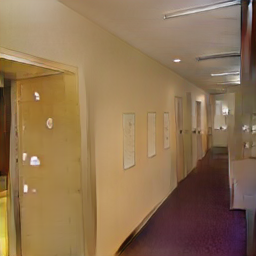}\\
\includegraphics[width=\linewidth]{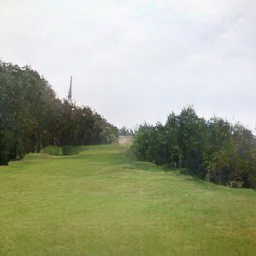}\\
\includegraphics[width=\linewidth]{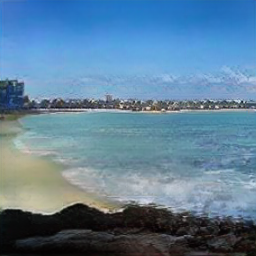}\\
\includegraphics[width=\linewidth]{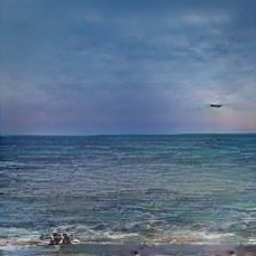}\\
\includegraphics[width=\linewidth]{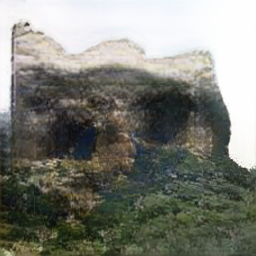}\\
\includegraphics[width=\linewidth]{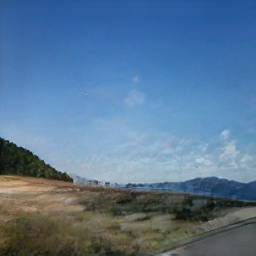}\\
\includegraphics[width=\linewidth]{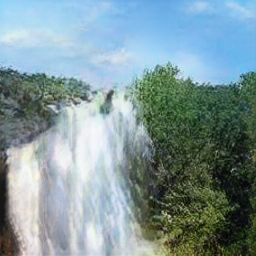}\\
\end{minipage}
} \hspace{-3mm}
\subfigure[SPADE]{
\begin{minipage}[t]{0.13\linewidth}
\centering
\includegraphics[width=\linewidth]{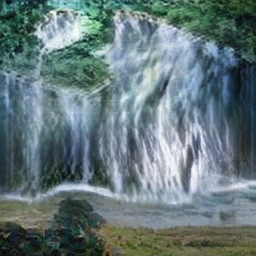}\\
\includegraphics[width=\linewidth]{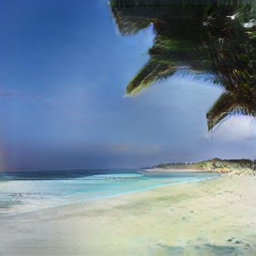}\\
\includegraphics[width=\linewidth]{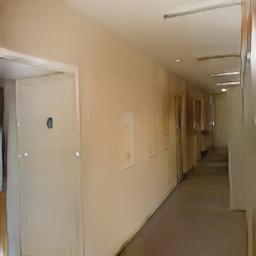}\\
\includegraphics[width=\linewidth]{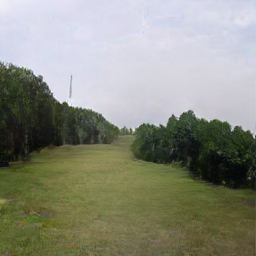}\\
\includegraphics[width=\linewidth]{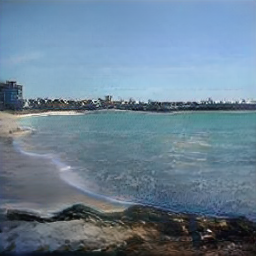}\\
\includegraphics[width=\linewidth]{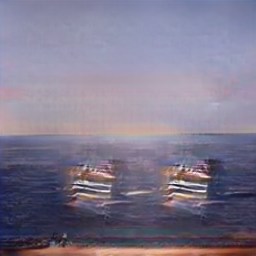}\\
\includegraphics[width=\linewidth]{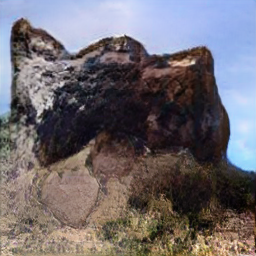}\\
\includegraphics[width=\linewidth]{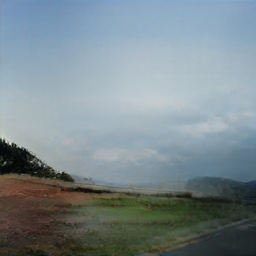}\\
\includegraphics[width=\linewidth]{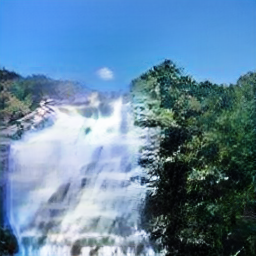}\\
\end{minipage}
} \hspace{-3mm}
\subfigure[OASIS]{
\begin{minipage}[t]{0.13\linewidth}
\centering
\includegraphics[width=\linewidth]{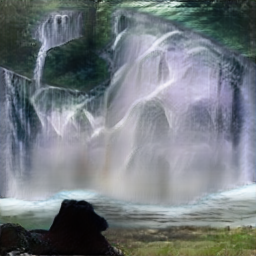}\\
\includegraphics[width=\linewidth]{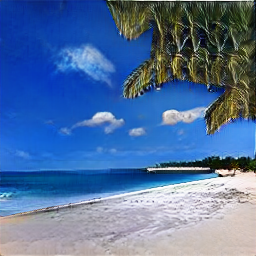}\\
\includegraphics[width=\linewidth]{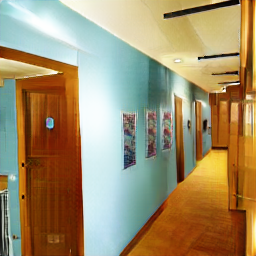}\\
\includegraphics[width=\linewidth]{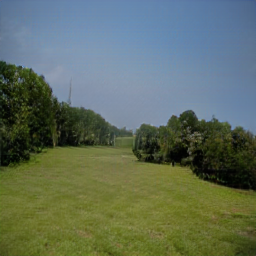}\\
\includegraphics[width=\linewidth]{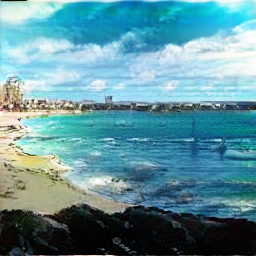}\\
\includegraphics[width=\linewidth]{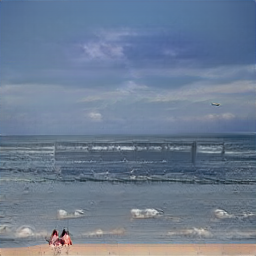}\\
\includegraphics[width=\linewidth]{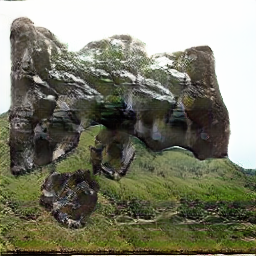}\\
\includegraphics[width=\linewidth]{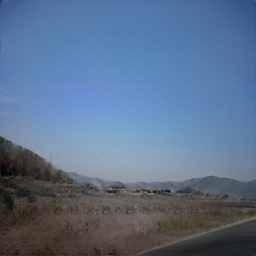}\\
\includegraphics[width=\linewidth]{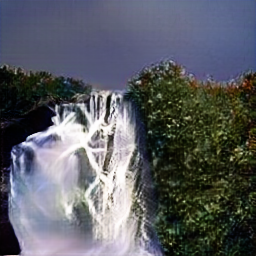}\\
\end{minipage}
} \hspace{-3mm}
\subfigure[Ours]{
\begin{minipage}[t]{0.13\linewidth}
\centering
\includegraphics[width=\linewidth]{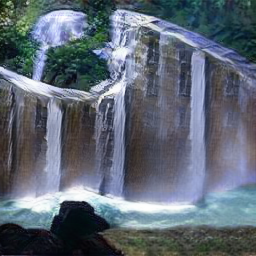}\\
\includegraphics[width=\linewidth]{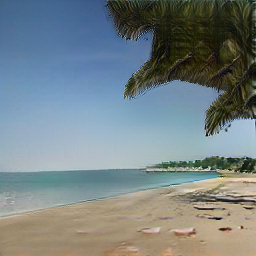}\\
\includegraphics[width=\linewidth]{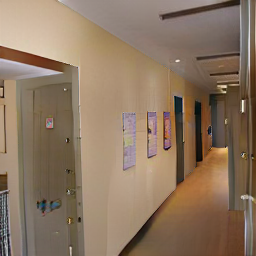}\\
\includegraphics[width=\linewidth]{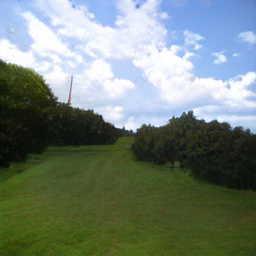}\\
\includegraphics[width=\linewidth]{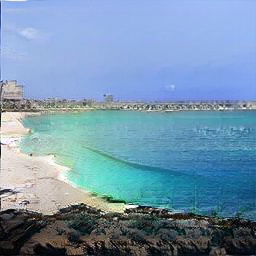}\\
\includegraphics[width=\linewidth]{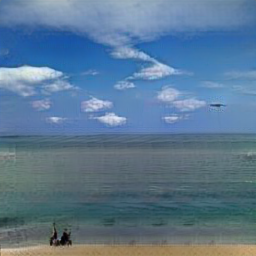}\\
\includegraphics[width=\linewidth]{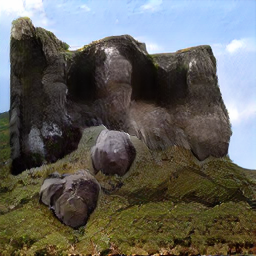}\\
\includegraphics[width=\linewidth]{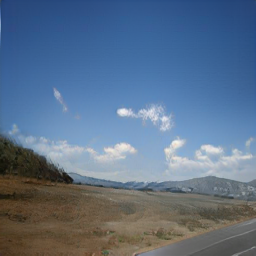}\\
\includegraphics[width=\linewidth]{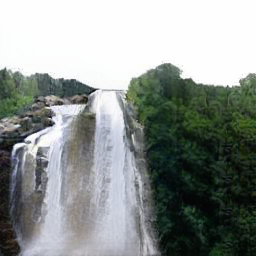}\\
\end{minipage}
} 
\caption{Qualitative results for ADE20K.}
\label{fig:qvis_cityscapes}
\end{figure*}

\begin{figure*}[htbp]
\centering
\subfigure[Label]{
\begin{minipage}[t]{0.13\linewidth}
\centering
\includegraphics[width=\linewidth]{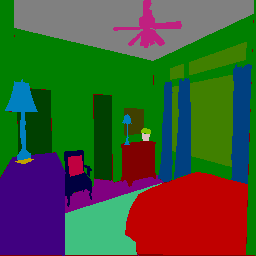}\\
\includegraphics[width=\linewidth]{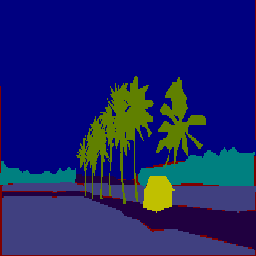}\\
\includegraphics[width=\linewidth]{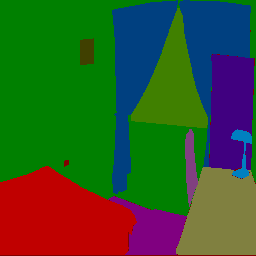}\\
\includegraphics[width=\linewidth]{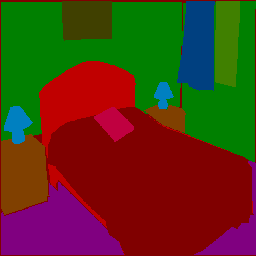}\\
\includegraphics[width=\linewidth]{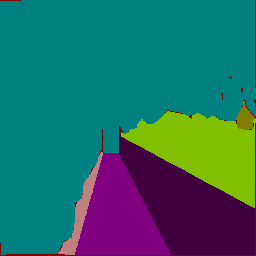}\\
\includegraphics[width=\linewidth]{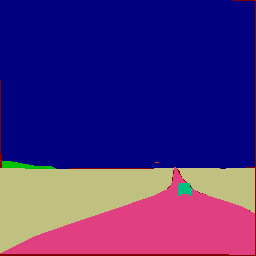}\\
\includegraphics[width=\linewidth]{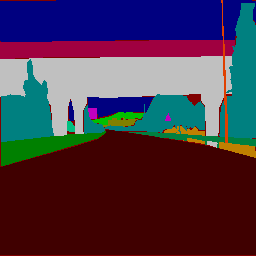}\\
\includegraphics[width=\linewidth]{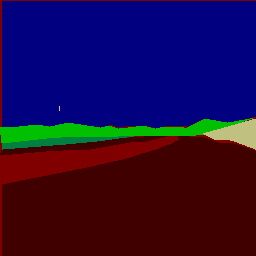}\\
\includegraphics[width=\linewidth]{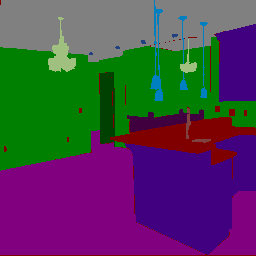}\\
\end{minipage}
}  \hspace{-3mm}
\subfigure[CC-FPSE]{
\begin{minipage}[t]{0.13\linewidth}
\centering
\includegraphics[width=\linewidth]{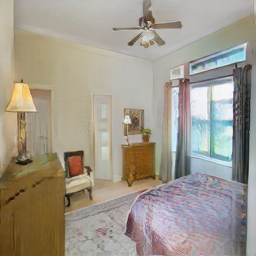}\\
\includegraphics[width=\linewidth]{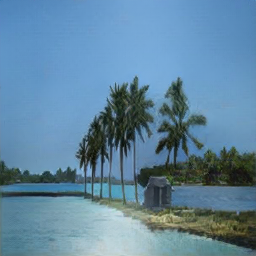}\\
\includegraphics[width=\linewidth]{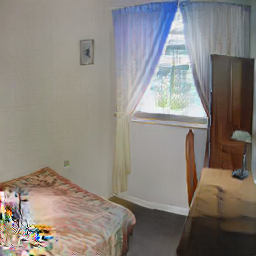}\\
\includegraphics[width=\linewidth]{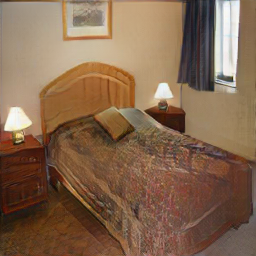}\\
\includegraphics[width=\linewidth]{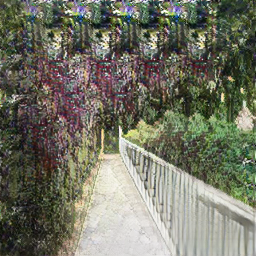}\\
\includegraphics[width=\linewidth]{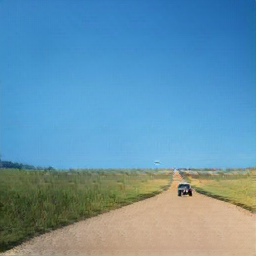}\\
\includegraphics[width=\linewidth]{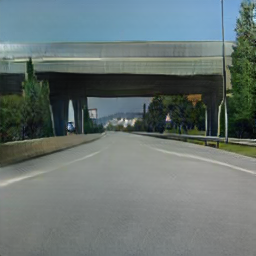}\\
\includegraphics[width=\linewidth]{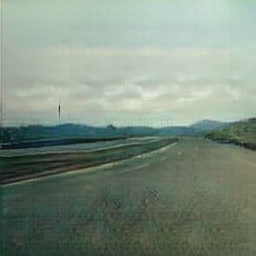}\\
\includegraphics[width=\linewidth]{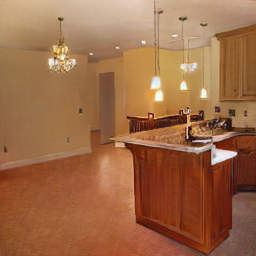}\\
\end{minipage}
} \hspace{-3mm}
\subfigure[DAGAN]{
\begin{minipage}[t]{0.13\linewidth}
\centering
\includegraphics[width=\linewidth]{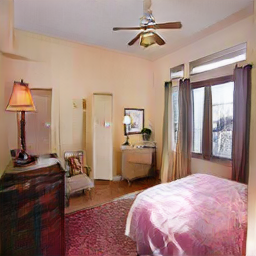}\\
\includegraphics[width=\linewidth]{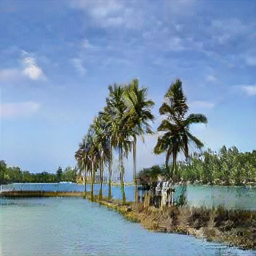}\\
\includegraphics[width=\linewidth]{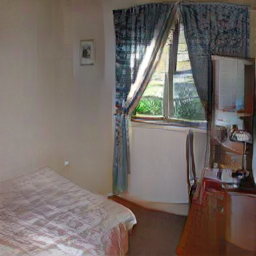}\\
\includegraphics[width=\linewidth]{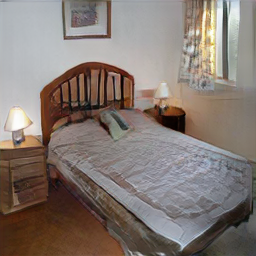}\\
\includegraphics[width=\linewidth]{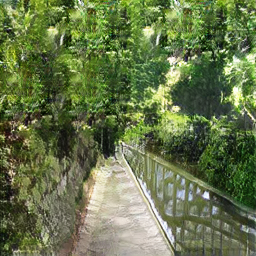}\\
\includegraphics[width=\linewidth]{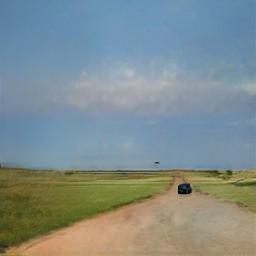}\\
\includegraphics[width=\linewidth]{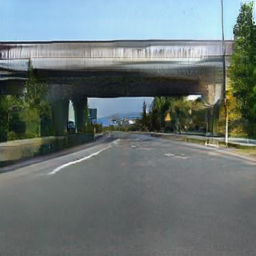}\\
\includegraphics[width=\linewidth]{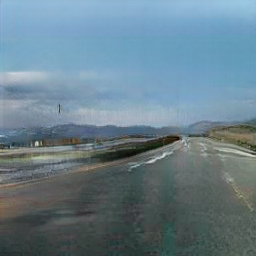}\\
\includegraphics[width=\linewidth]{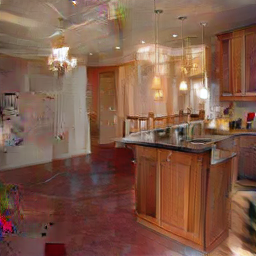}\\
\end{minipage}
} \hspace{-3mm}
\subfigure[LGGAN]{
\begin{minipage}[t]{0.13\linewidth}
\centering
\includegraphics[width=\linewidth]{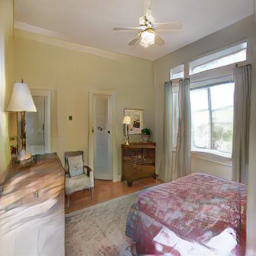}\\
\includegraphics[width=\linewidth]{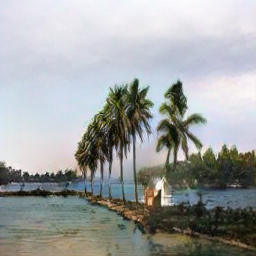}\\
\includegraphics[width=\linewidth]{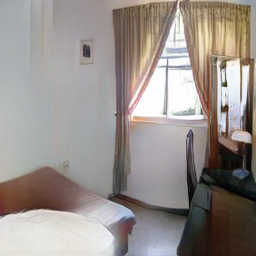}\\
\includegraphics[width=\linewidth]{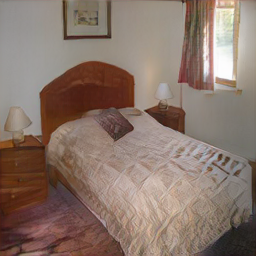}\\
\includegraphics[width=\linewidth]{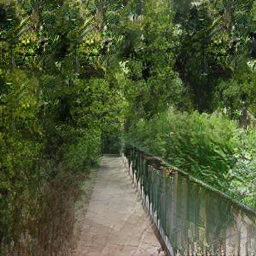}\\
\includegraphics[width=\linewidth]{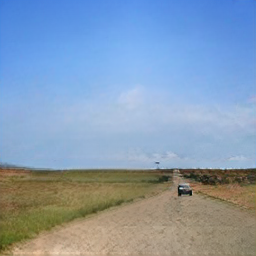}\\
\includegraphics[width=\linewidth]{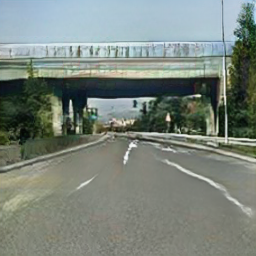}\\
\includegraphics[width=\linewidth]{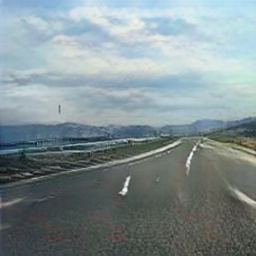}\\
\includegraphics[width=\linewidth]{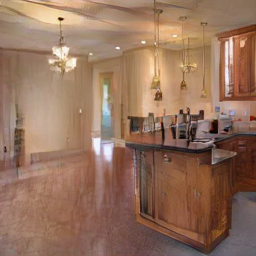}\\
\end{minipage}
} \hspace{-3mm}
\subfigure[SPADE]{
\begin{minipage}[t]{0.13\linewidth}
\centering
\includegraphics[width=\linewidth]{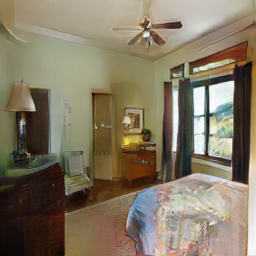}\\
\includegraphics[width=\linewidth]{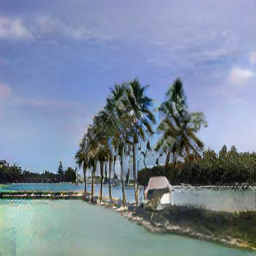}\\
\includegraphics[width=\linewidth]{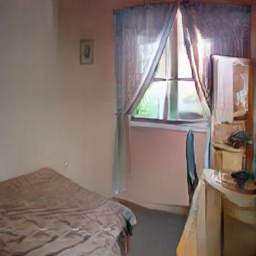}\\
\includegraphics[width=\linewidth]{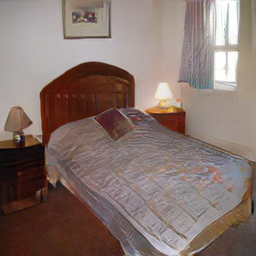}\\
\includegraphics[width=\linewidth]{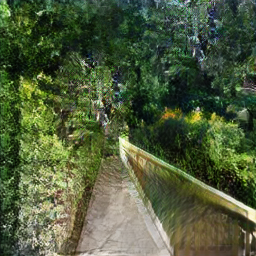}\\
\includegraphics[width=\linewidth]{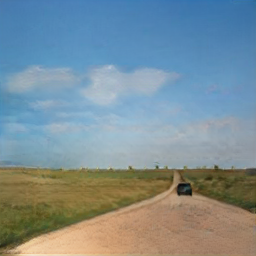}\\
\includegraphics[width=\linewidth]{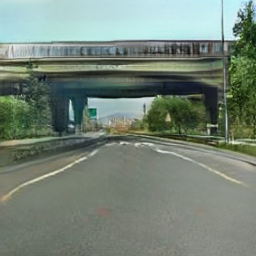}\\
\includegraphics[width=\linewidth]{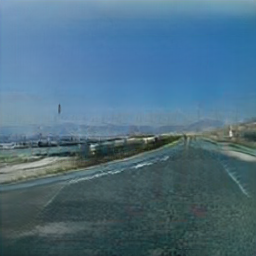}\\
\includegraphics[width=\linewidth]{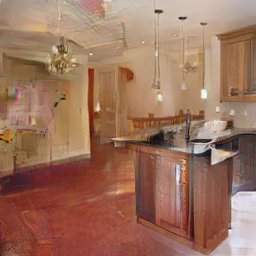}\\
\end{minipage}
} \hspace{-3mm}
\subfigure[OASIS]{
\begin{minipage}[t]{0.13\linewidth}
\centering
\includegraphics[width=\linewidth]{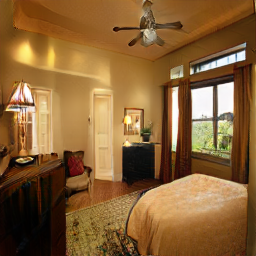}\\
\includegraphics[width=\linewidth]{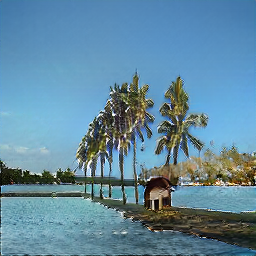}\\
\includegraphics[width=\linewidth]{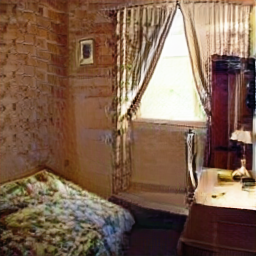}\\
\includegraphics[width=\linewidth]{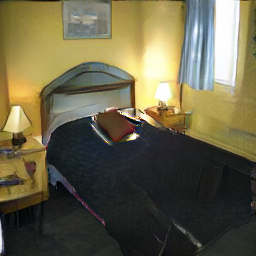}\\
\includegraphics[width=\linewidth]{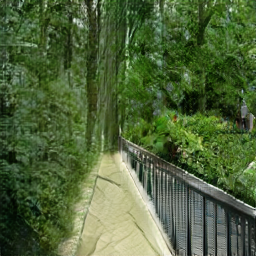}\\
\includegraphics[width=\linewidth]{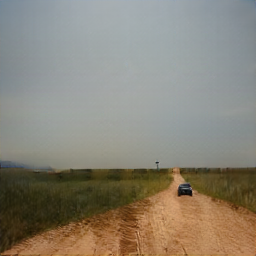}\\
\includegraphics[width=\linewidth]{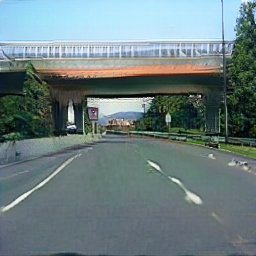}\\
\includegraphics[width=\linewidth]{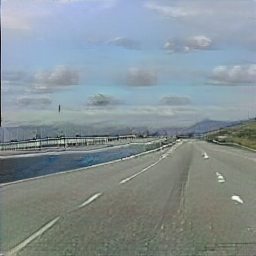}\\
\includegraphics[width=\linewidth]{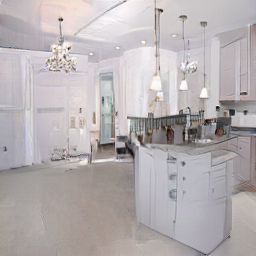}\\
\end{minipage}
} \hspace{-3mm}
\subfigure[Ours]{
\begin{minipage}[t]{0.13\linewidth}
\centering
\includegraphics[width=\linewidth]{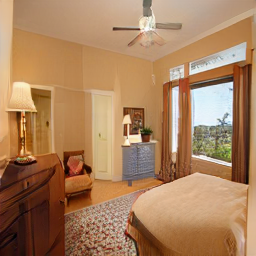}\\
\includegraphics[width=\linewidth]{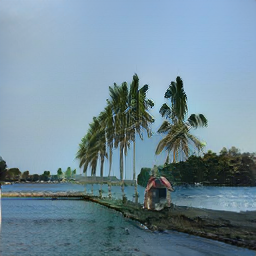}\\
\includegraphics[width=\linewidth]{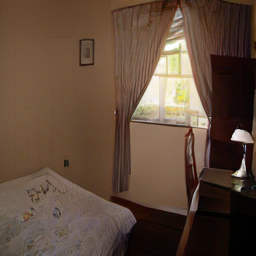}\\
\includegraphics[width=\linewidth]{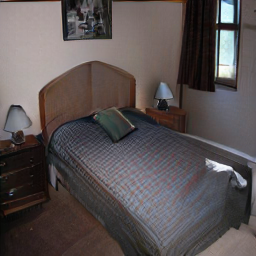}\\
\includegraphics[width=\linewidth]{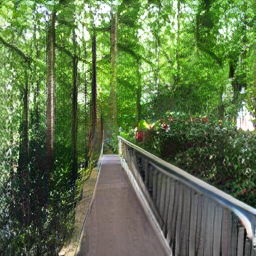}\\
\includegraphics[width=\linewidth]{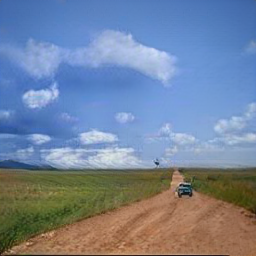}\\
\includegraphics[width=\linewidth]{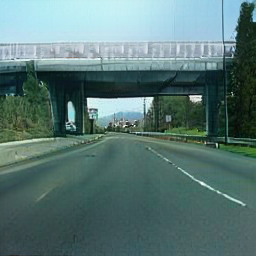}\\
\includegraphics[width=\linewidth]{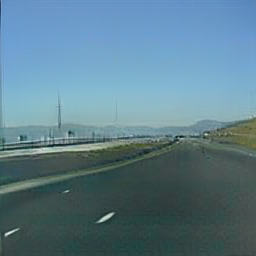}\\
\includegraphics[width=\linewidth]{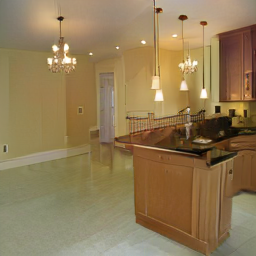}\\
\end{minipage}
} 
\caption{Qualitative results for ADE20K.}
\label{fig:qvis_cityscapes}
\end{figure*}

\begin{figure*}[htbp]
\centering
\subfigure[Label]{
\begin{minipage}[t]{0.13\linewidth}
\centering
\includegraphics[width=\linewidth]{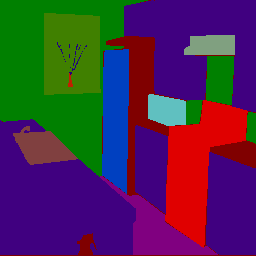}\\
\includegraphics[width=\linewidth]{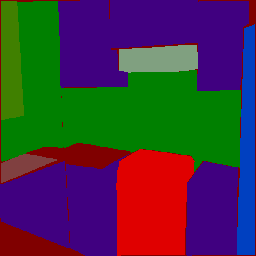}\\
\includegraphics[width=\linewidth]{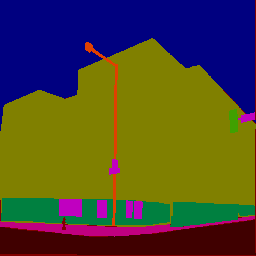}\\
\includegraphics[width=\linewidth]{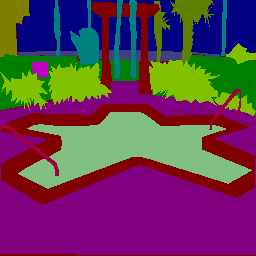}\\
\includegraphics[width=\linewidth]{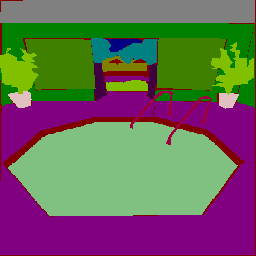}\\
\includegraphics[width=\linewidth]{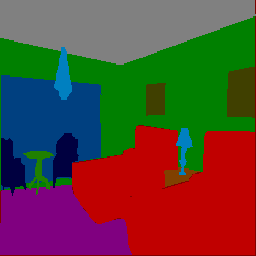}\\
\includegraphics[width=\linewidth]{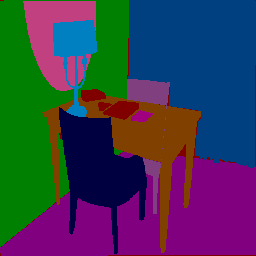}\\
\includegraphics[width=\linewidth]{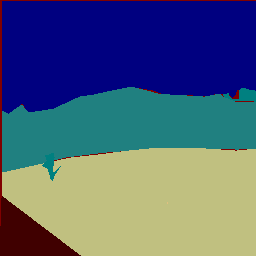}\\
\includegraphics[width=\linewidth]{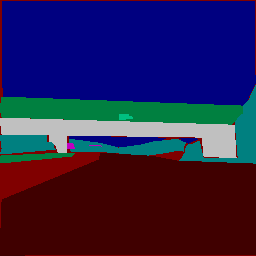}\\
\end{minipage}
}  \hspace{-3mm}
\subfigure[CC-FPSE]{
\begin{minipage}[t]{0.13\linewidth}
\centering
\includegraphics[width=\linewidth]{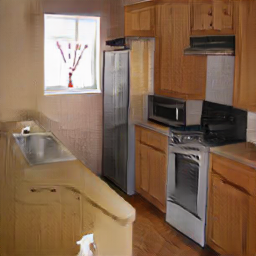}\\
\includegraphics[width=\linewidth]{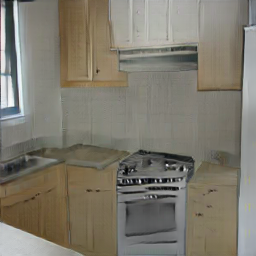}\\
\includegraphics[width=\linewidth]{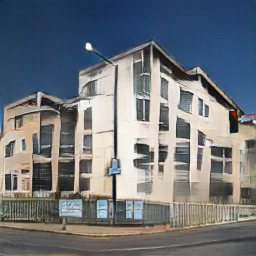}\\
\includegraphics[width=\linewidth]{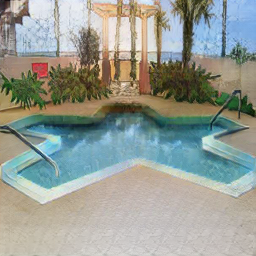}\\
\includegraphics[width=\linewidth]{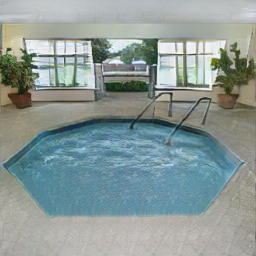}\\
\includegraphics[width=\linewidth]{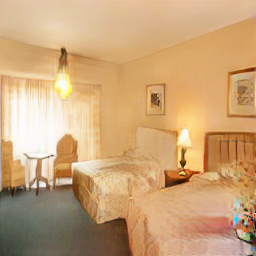}\\
\includegraphics[width=\linewidth]{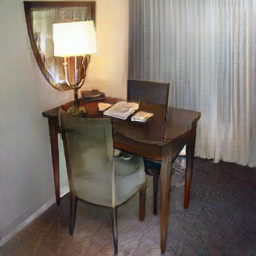}\\
\includegraphics[width=\linewidth]{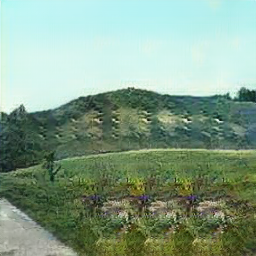}\\
\includegraphics[width=\linewidth]{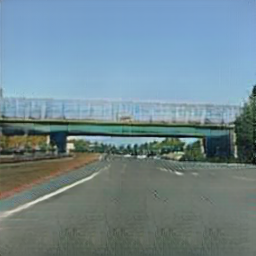}\\
\end{minipage}
} \hspace{-3mm}
\subfigure[DAGAN]{
\begin{minipage}[t]{0.13\linewidth}
\centering
\includegraphics[width=\linewidth]{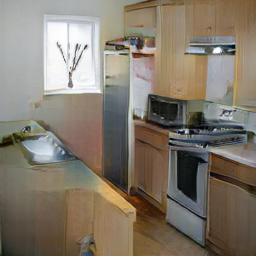}\\
\includegraphics[width=\linewidth]{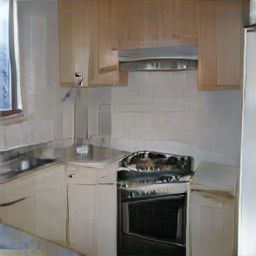}\\
\includegraphics[width=\linewidth]{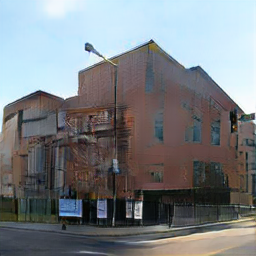}\\
\includegraphics[width=\linewidth]{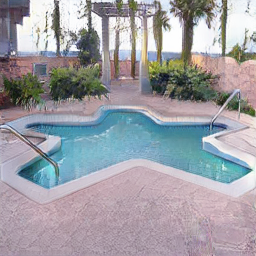}\\
\includegraphics[width=\linewidth]{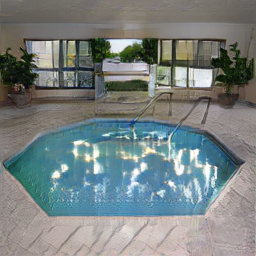}\\
\includegraphics[width=\linewidth]{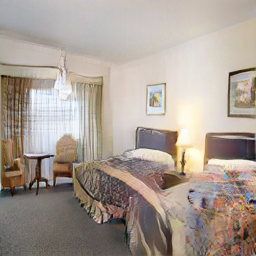}\\
\includegraphics[width=\linewidth]{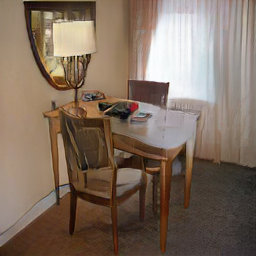}\\
\includegraphics[width=\linewidth]{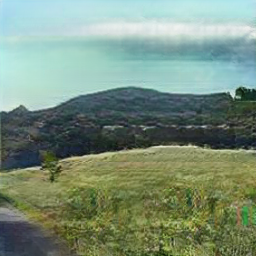}\\
\includegraphics[width=\linewidth]{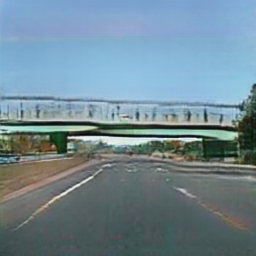}\\
\end{minipage}
} \hspace{-3mm}
\subfigure[LGGAN]{
\begin{minipage}[t]{0.13\linewidth}
\centering
\includegraphics[width=\linewidth]{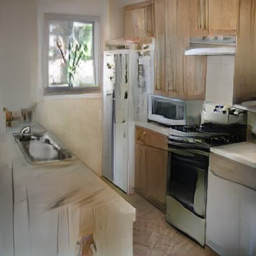}\\
\includegraphics[width=\linewidth]{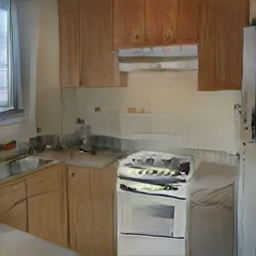}\\
\includegraphics[width=\linewidth]{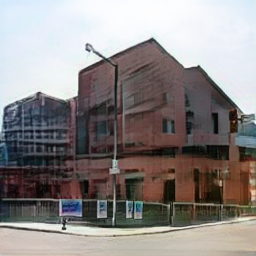}\\
\includegraphics[width=\linewidth]{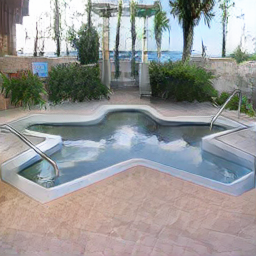}\\
\includegraphics[width=\linewidth]{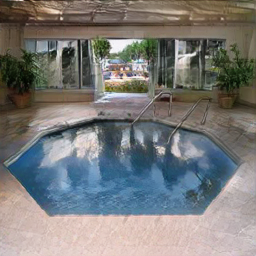}\\
\includegraphics[width=\linewidth]{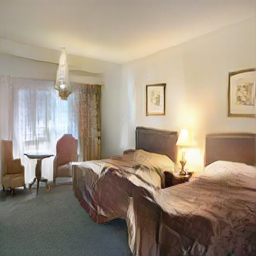}\\
\includegraphics[width=\linewidth]{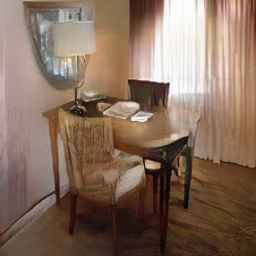}\\
\includegraphics[width=\linewidth]{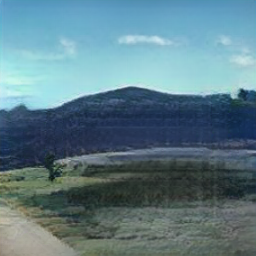}\\
\includegraphics[width=\linewidth]{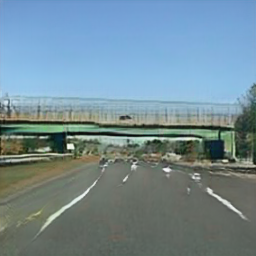}\\
\end{minipage}
} \hspace{-3mm}
\subfigure[SPADE]{
\begin{minipage}[t]{0.13\linewidth}
\centering
\includegraphics[width=\linewidth]{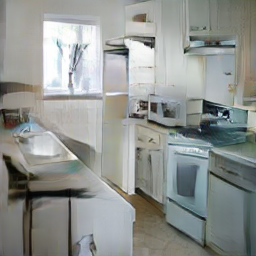}\\
\includegraphics[width=\linewidth]{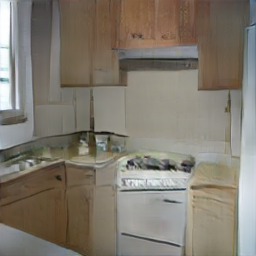}\\
\includegraphics[width=\linewidth]{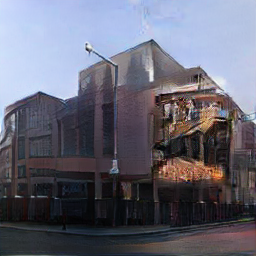}\\
\includegraphics[width=\linewidth]{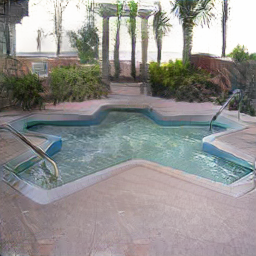}\\
\includegraphics[width=\linewidth]{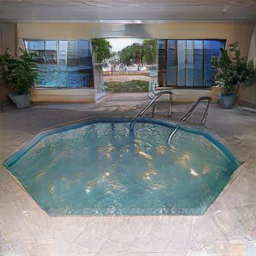}\\
\includegraphics[width=\linewidth]{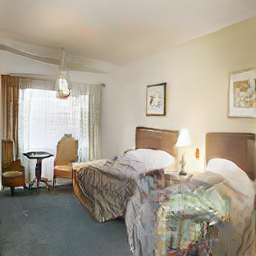}\\
\includegraphics[width=\linewidth]{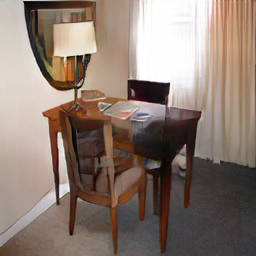}\\
\includegraphics[width=\linewidth]{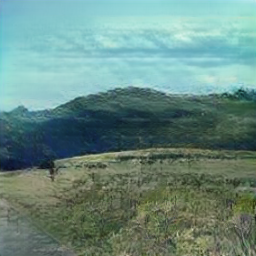}\\
\includegraphics[width=\linewidth]{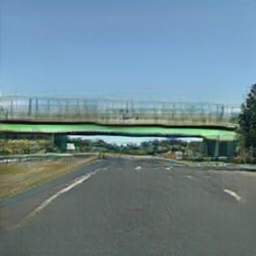}\\
\end{minipage}
} \hspace{-3mm}
\subfigure[OASIS]{
\begin{minipage}[t]{0.13\linewidth}
\centering
\includegraphics[width=\linewidth]{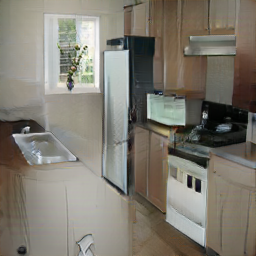}\\
\includegraphics[width=\linewidth]{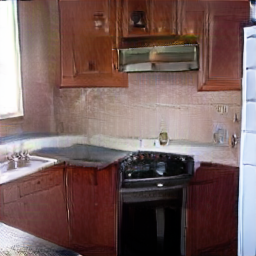}\\
\includegraphics[width=\linewidth]{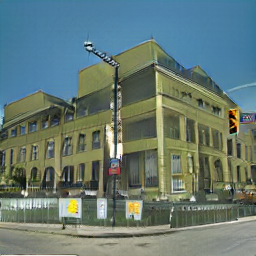}\\
\includegraphics[width=\linewidth]{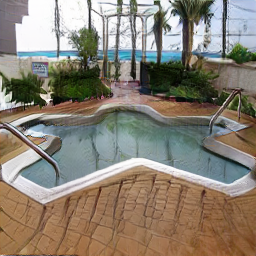}\\
\includegraphics[width=\linewidth]{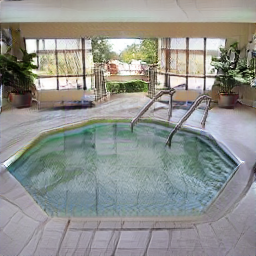}\\
\includegraphics[width=\linewidth]{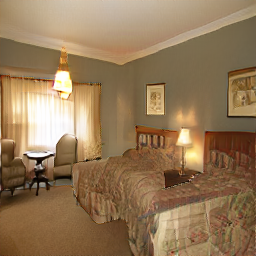}\\
\includegraphics[width=\linewidth]{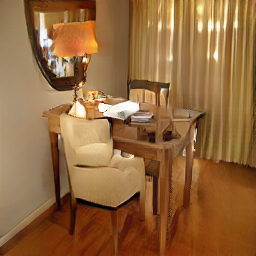}\\
\includegraphics[width=\linewidth]{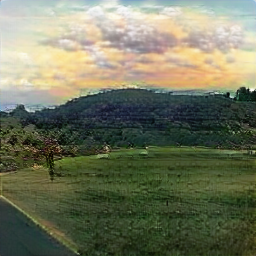}\\
\includegraphics[width=\linewidth]{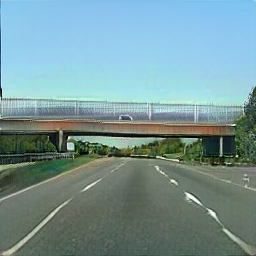}\\
\end{minipage}
} \hspace{-3mm}
\subfigure[Ours]{
\begin{minipage}[t]{0.13\linewidth}
\centering
\includegraphics[width=\linewidth]{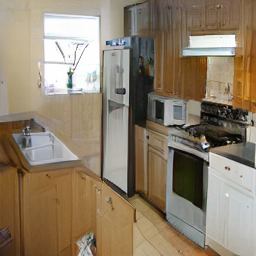}\\
\includegraphics[width=\linewidth]{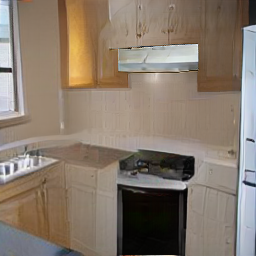}\\
\includegraphics[width=\linewidth]{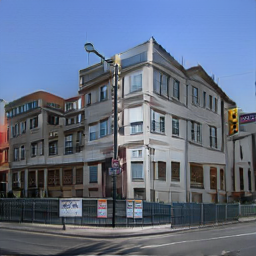}\\
\includegraphics[width=\linewidth]{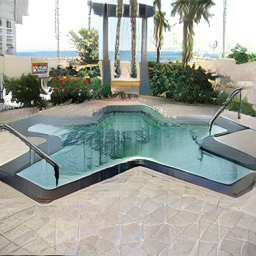}\\
\includegraphics[width=\linewidth]{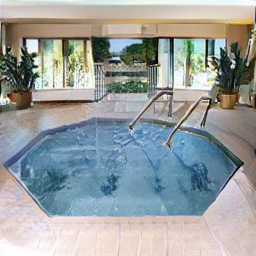}\\
\includegraphics[width=\linewidth]{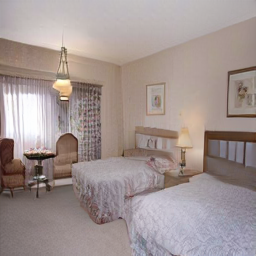}\\
\includegraphics[width=\linewidth]{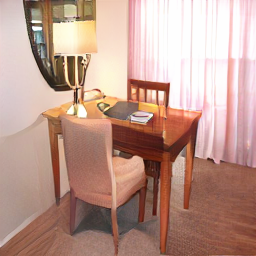}\\
\includegraphics[width=\linewidth]{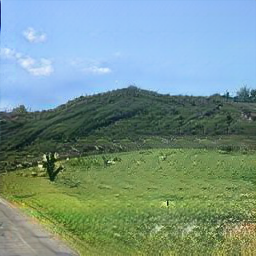}\\
\includegraphics[width=\linewidth]{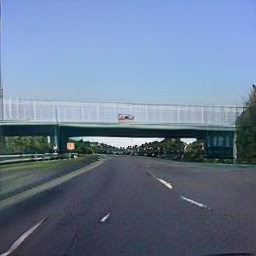}\\
\end{minipage}
} \hspace*{-2mm}
\caption{Qualitative results for ADE20K.}
\label{fig:qvis_cityscapes}
\end{figure*}

\begin{figure*}[htbp]
\centering
\subfigure[Label]{
\begin{minipage}[t]{0.13\linewidth}
\centering
\includegraphics[width=\linewidth]{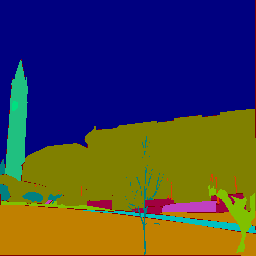}\\
\includegraphics[width=\linewidth]{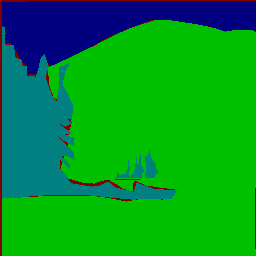}\\
\includegraphics[width=\linewidth]{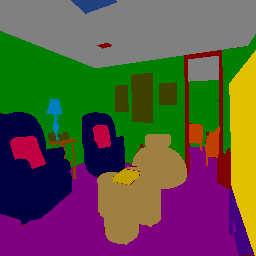}\\
\includegraphics[width=\linewidth]{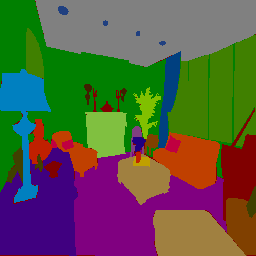}\\
\includegraphics[width=\linewidth]{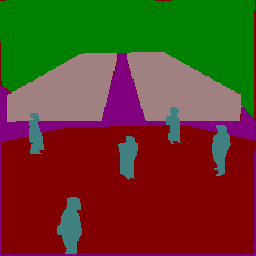}\\
\includegraphics[width=\linewidth]{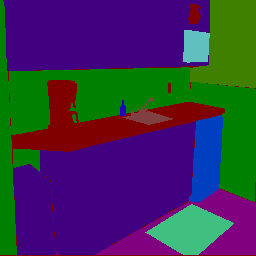}\\
\includegraphics[width=\linewidth]{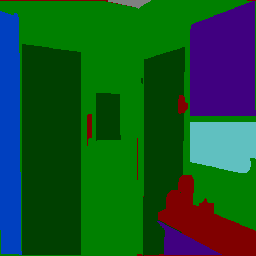}\\
\includegraphics[width=\linewidth]{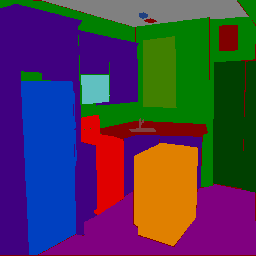}\\
\includegraphics[width=\linewidth]{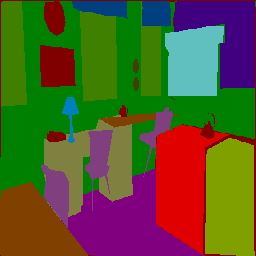}\\
\end{minipage}
}  \hspace{-3mm}
\subfigure[CC-FPSE]{
\begin{minipage}[t]{0.13\linewidth}
\centering
\includegraphics[width=\linewidth]{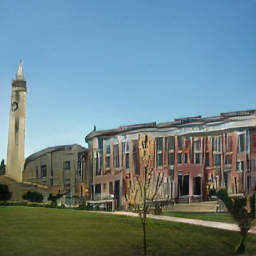}\\
\includegraphics[width=\linewidth]{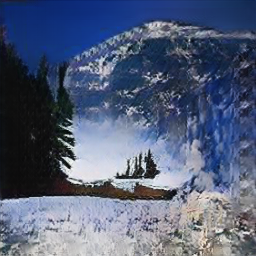}\\
\includegraphics[width=\linewidth]{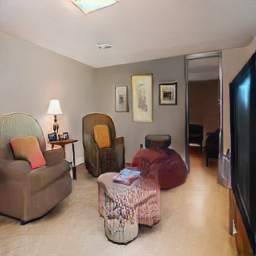}\\
\includegraphics[width=\linewidth]{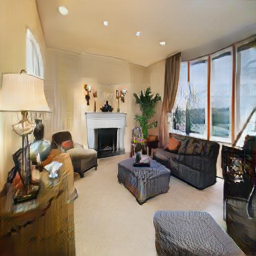}\\
\includegraphics[width=\linewidth]{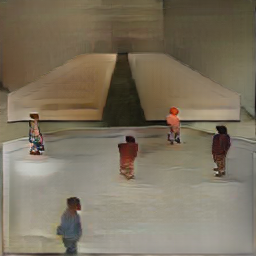}\\
\includegraphics[width=\linewidth]{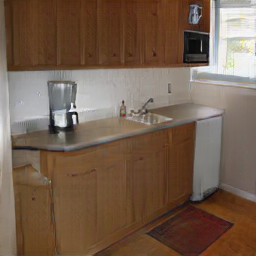}\\
\includegraphics[width=\linewidth]{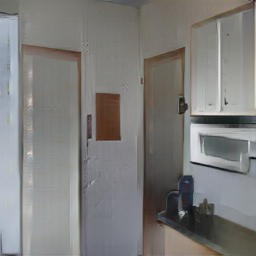}\\
\includegraphics[width=\linewidth]{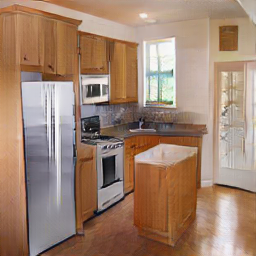}\\
\includegraphics[width=\linewidth]{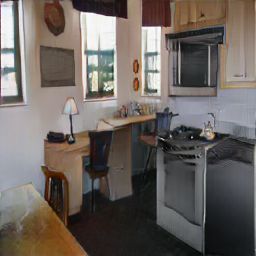}\\
\end{minipage}
} \hspace{-3mm}
\subfigure[DAGAN]{
\begin{minipage}[t]{0.13\linewidth}
\centering
\includegraphics[width=\linewidth]{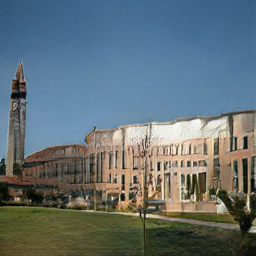}\\
\includegraphics[width=\linewidth]{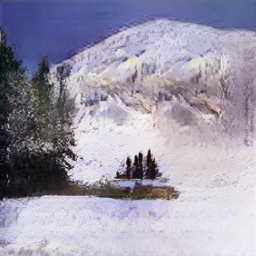}\\
\includegraphics[width=\linewidth]{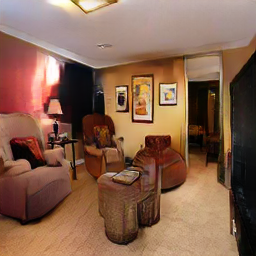}\\
\includegraphics[width=\linewidth]{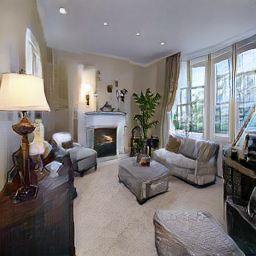}\\
\includegraphics[width=\linewidth]{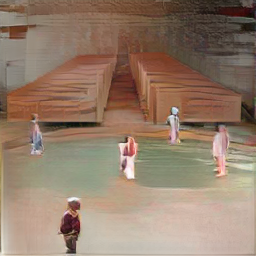}\\
\includegraphics[width=\linewidth]{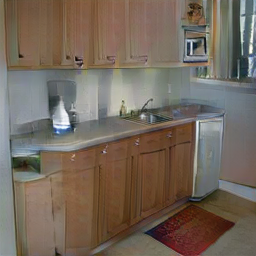}\\
\includegraphics[width=\linewidth]{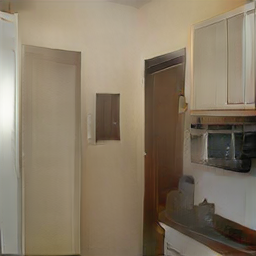}\\
\includegraphics[width=\linewidth]{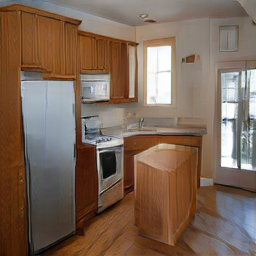}\\
\includegraphics[width=\linewidth]{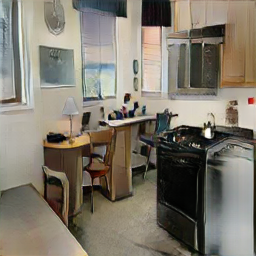}\\
\end{minipage}
} \hspace{-3mm}
\subfigure[LGGAN]{
\begin{minipage}[t]{0.13\linewidth}
\centering
\includegraphics[width=\linewidth]{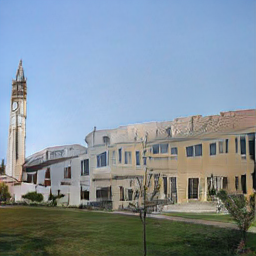}\\
\includegraphics[width=\linewidth]{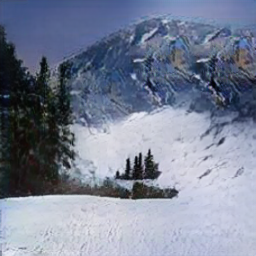}\\
\includegraphics[width=\linewidth]{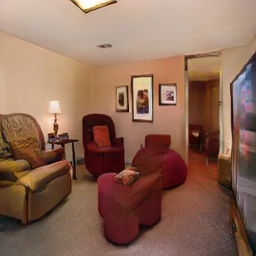}\\
\includegraphics[width=\linewidth]{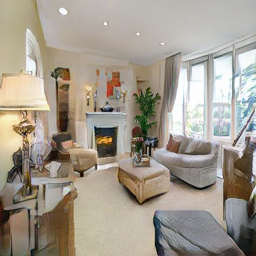}\\
\includegraphics[width=\linewidth]{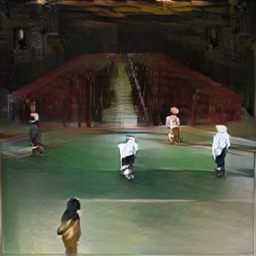}\\
\includegraphics[width=\linewidth]{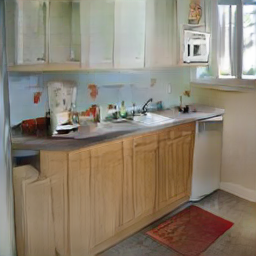}\\
\includegraphics[width=\linewidth]{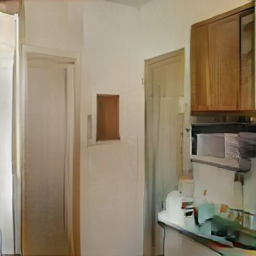}\\
\includegraphics[width=\linewidth]{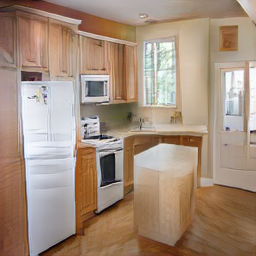}\\
\includegraphics[width=\linewidth]{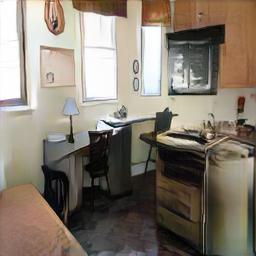}\\
\end{minipage}
} \hspace{-3mm}
\subfigure[SPADE]{
\begin{minipage}[t]{0.13\linewidth}
\centering
\includegraphics[width=\linewidth]{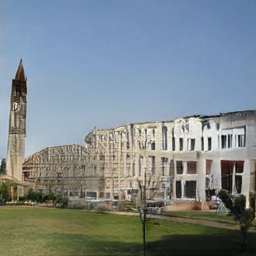}\\
\includegraphics[width=\linewidth]{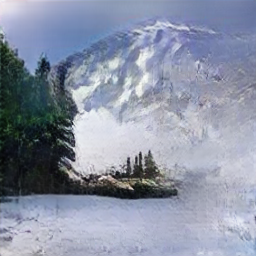}\\
\includegraphics[width=\linewidth]{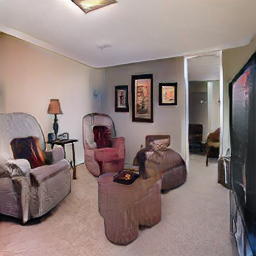}\\
\includegraphics[width=\linewidth]{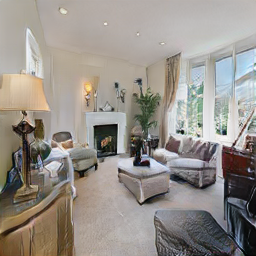}\\
\includegraphics[width=\linewidth]{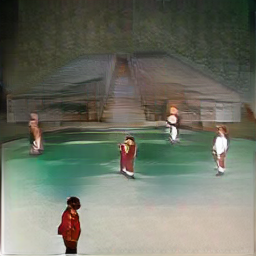}\\
\includegraphics[width=\linewidth]{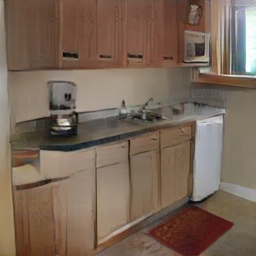}\\
\includegraphics[width=\linewidth]{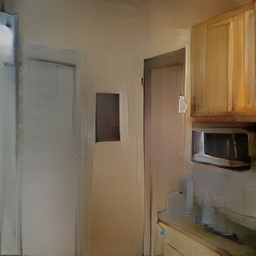}\\
\includegraphics[width=\linewidth]{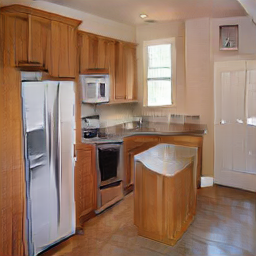}\\
\includegraphics[width=\linewidth]{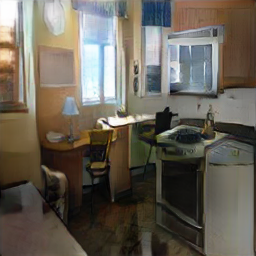}\\
\end{minipage}
} \hspace{-3mm}
\subfigure[OASIS]{
\begin{minipage}[t]{0.13\linewidth}
\centering
\includegraphics[width=\linewidth]{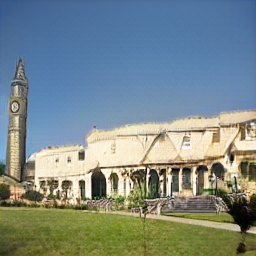}\\
\includegraphics[width=\linewidth]{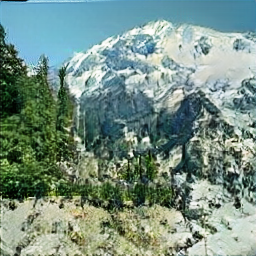}\\
\includegraphics[width=\linewidth]{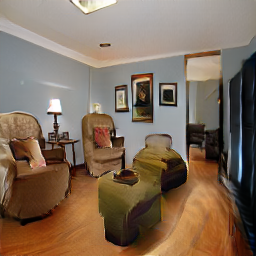}\\
\includegraphics[width=\linewidth]{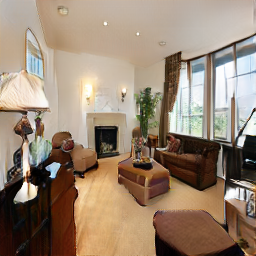}\\
\includegraphics[width=\linewidth]{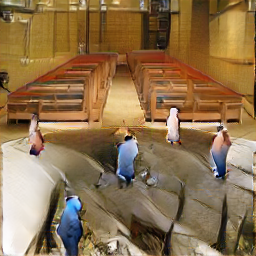}\\
\includegraphics[width=\linewidth]{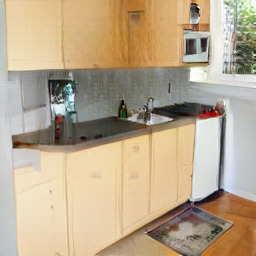}\\
\includegraphics[width=\linewidth]{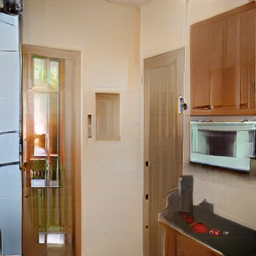}\\
\includegraphics[width=\linewidth]{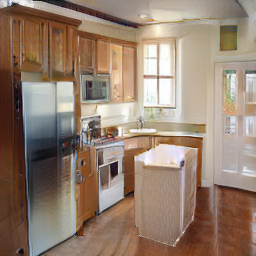}\\
\includegraphics[width=\linewidth]{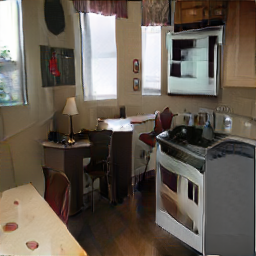}\\
\end{minipage}
} \hspace{-3mm}
\subfigure[Ours]{
\begin{minipage}[t]{0.13\linewidth}
\centering
\includegraphics[width=\linewidth]{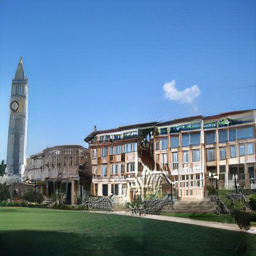}\\
\includegraphics[width=\linewidth]{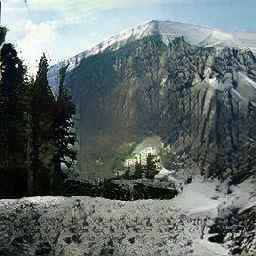}\\
\includegraphics[width=\linewidth]{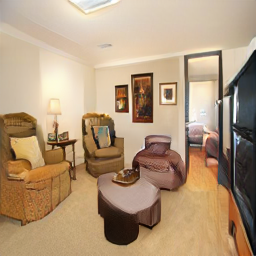}\\
\includegraphics[width=\linewidth]{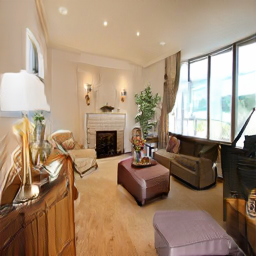}\\
\includegraphics[width=\linewidth]{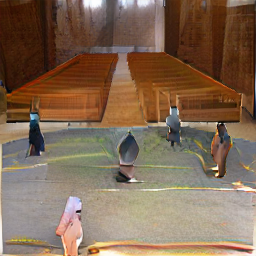}\\
\includegraphics[width=\linewidth]{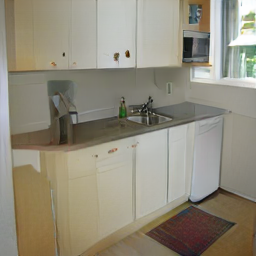}\\
\includegraphics[width=\linewidth]{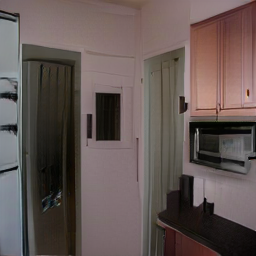}\\
\includegraphics[width=\linewidth]{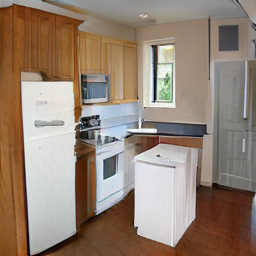}\\
\includegraphics[width=\linewidth]{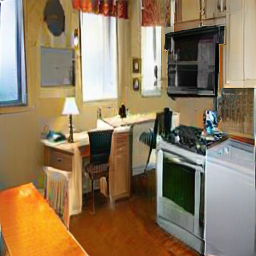}\\
\end{minipage}
} 
\caption{Qualitative results for ADE20K.}
\label{fig:qvis_cityscapes}
\end{figure*}

\begin{figure*}[htbp]
\centering
\subfigure[Label]{
\begin{minipage}[t]{0.13\linewidth}
\centering
\includegraphics[width=\linewidth]{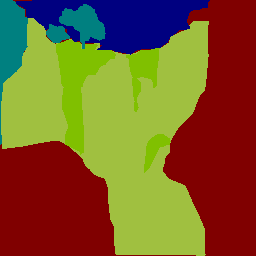}\\
\includegraphics[width=\linewidth]{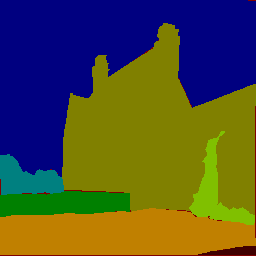}\\
\includegraphics[width=\linewidth]{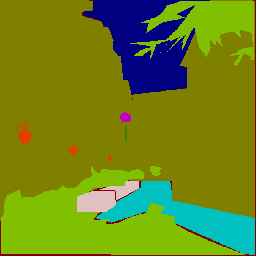}\\
\includegraphics[width=\linewidth]{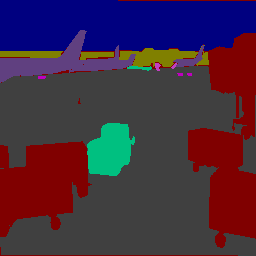}\\
\includegraphics[width=\linewidth]{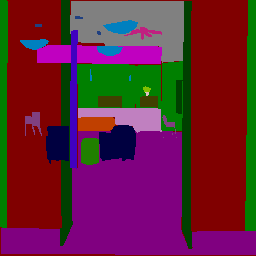}\\
\includegraphics[width=\linewidth]{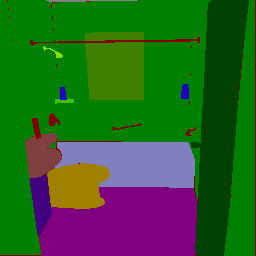}\\
\includegraphics[width=\linewidth]{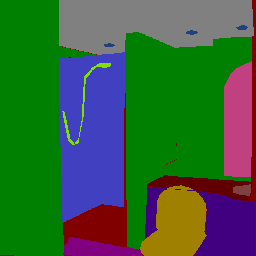}\\
\includegraphics[width=\linewidth]{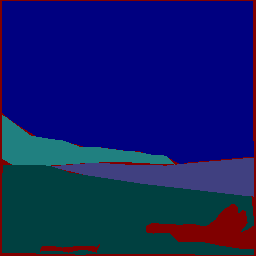}\\
\includegraphics[width=\linewidth]{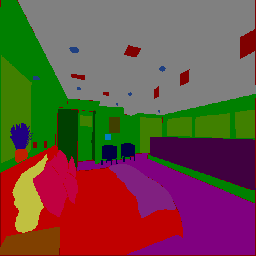}\\
\end{minipage}
}  \hspace{-3mm}
\subfigure[CC-FPSE]{
\begin{minipage}[t]{0.13\linewidth}
\centering
\includegraphics[width=\linewidth]{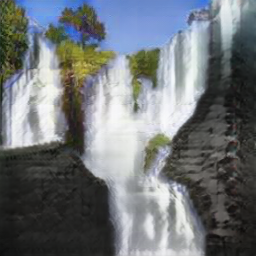}\\
\includegraphics[width=\linewidth]{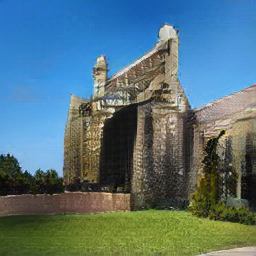}\\
\includegraphics[width=\linewidth]{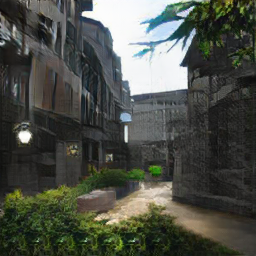}\\
\includegraphics[width=\linewidth]{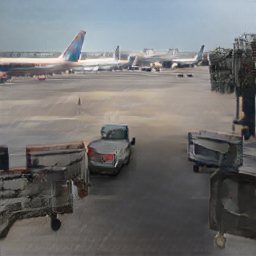}\\
\includegraphics[width=\linewidth]{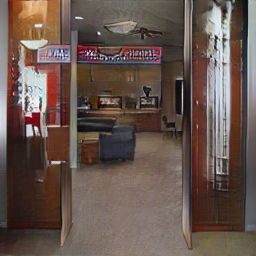}\\
\includegraphics[width=\linewidth]{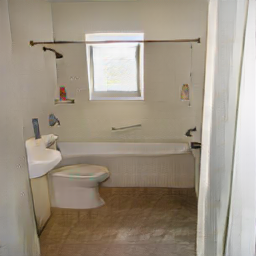}\\
\includegraphics[width=\linewidth]{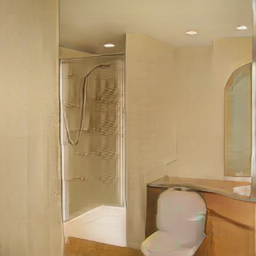}\\
\includegraphics[width=\linewidth]{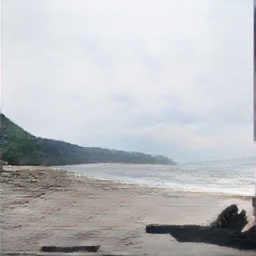}\\
\includegraphics[width=\linewidth]{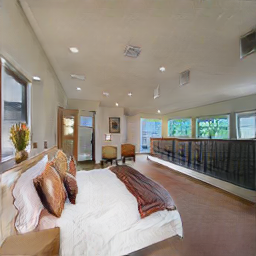}\\
\end{minipage}
} \hspace{-3mm}
\subfigure[DAGAN]{
\begin{minipage}[t]{0.13\linewidth}
\centering
\includegraphics[width=\linewidth]{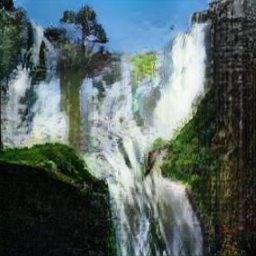}\\
\includegraphics[width=\linewidth]{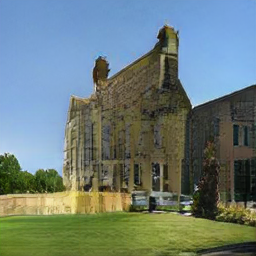}\\
\includegraphics[width=\linewidth]{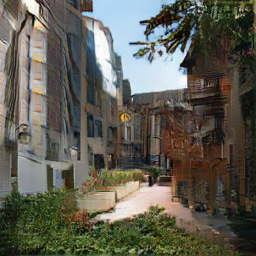}\\
\includegraphics[width=\linewidth]{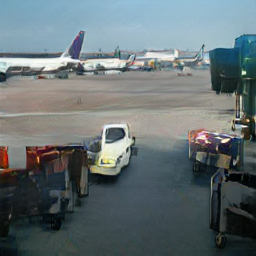}\\
\includegraphics[width=\linewidth]{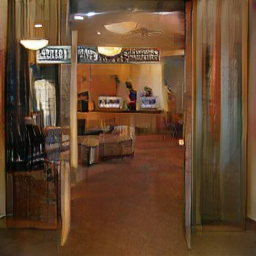}\\
\includegraphics[width=\linewidth]{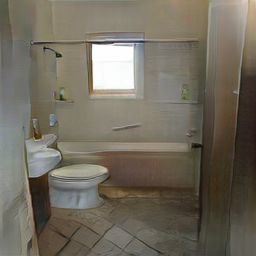}\\
\includegraphics[width=\linewidth]{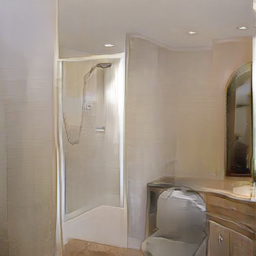}\\
\includegraphics[width=\linewidth]{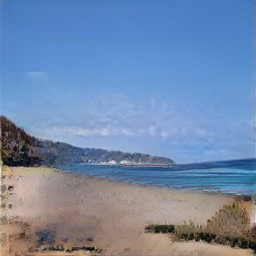}\\
\includegraphics[width=\linewidth]{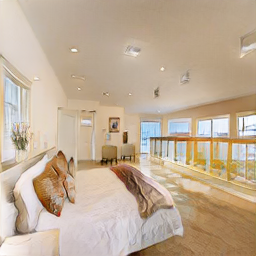}\\
\end{minipage}
} \hspace{-3mm}
\subfigure[LGGAN]{
\begin{minipage}[t]{0.13\linewidth}
\centering
\includegraphics[width=\linewidth]{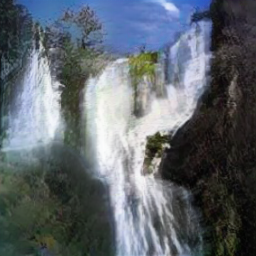}\\
\includegraphics[width=\linewidth]{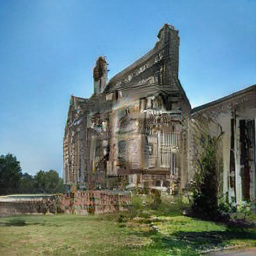}\\
\includegraphics[width=\linewidth]{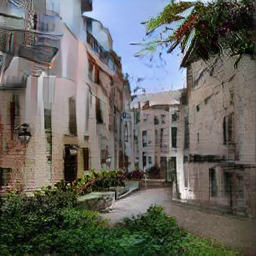}\\
\includegraphics[width=\linewidth]{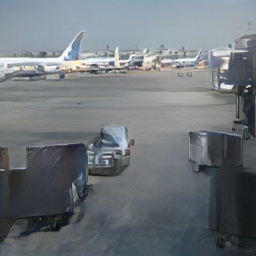}\\
\includegraphics[width=\linewidth]{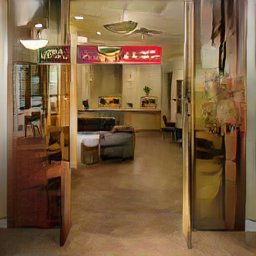}\\
\includegraphics[width=\linewidth]{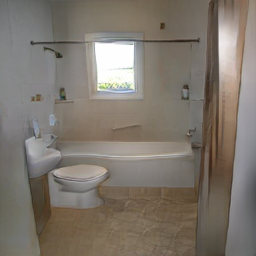}\\
\includegraphics[width=\linewidth]{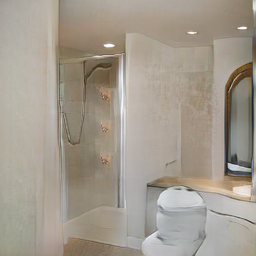}\\
\includegraphics[width=\linewidth]{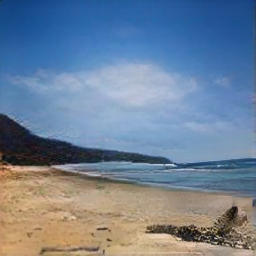}\\
\includegraphics[width=\linewidth]{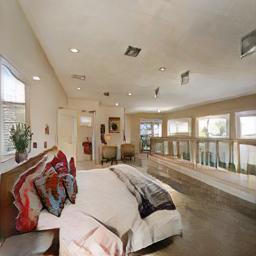}\\
\end{minipage}
} \hspace{-3mm}
\subfigure[SPADE]{
\begin{minipage}[t]{0.13\linewidth}
\centering
\includegraphics[width=\linewidth]{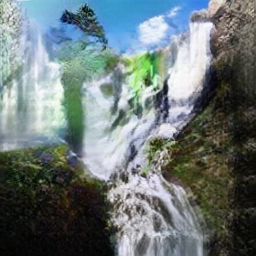}\\
\includegraphics[width=\linewidth]{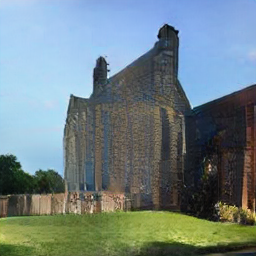}\\
\includegraphics[width=\linewidth]{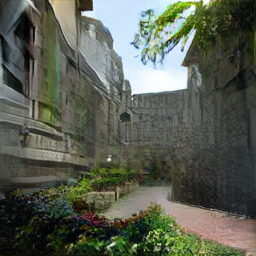}\\
\includegraphics[width=\linewidth]{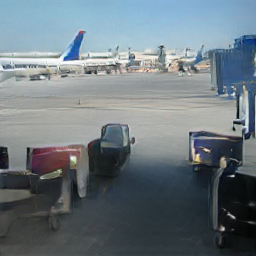}\\
\includegraphics[width=\linewidth]{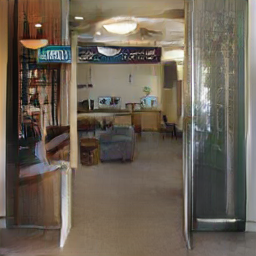}\\
\includegraphics[width=\linewidth]{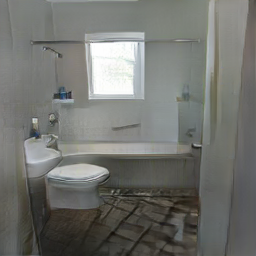}\\
\includegraphics[width=\linewidth]{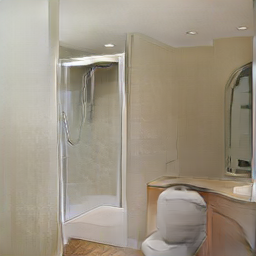}\\
\includegraphics[width=\linewidth]{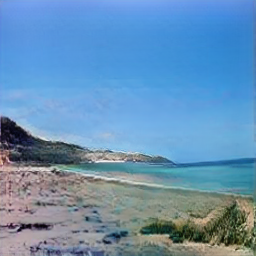}\\
\includegraphics[width=\linewidth]{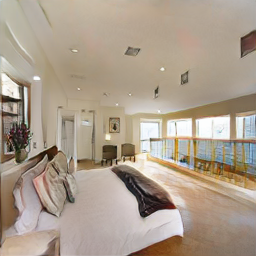}\\
\end{minipage}
} \hspace{-3mm}
\subfigure[OASIS]{
\begin{minipage}[t]{0.13\linewidth}
\centering
\includegraphics[width=\linewidth]{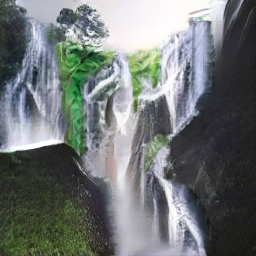}\\
\includegraphics[width=\linewidth]{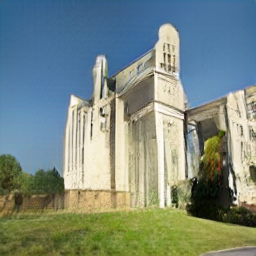}\\
\includegraphics[width=\linewidth]{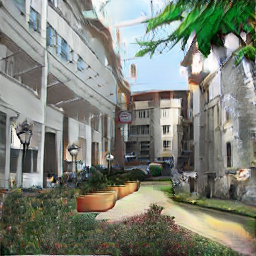}\\
\includegraphics[width=\linewidth]{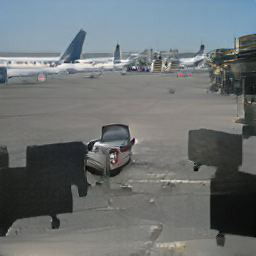}\\
\includegraphics[width=\linewidth]{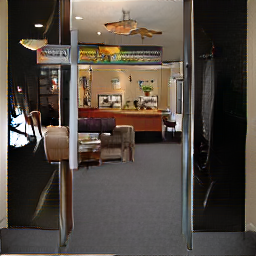}\\
\includegraphics[width=\linewidth]{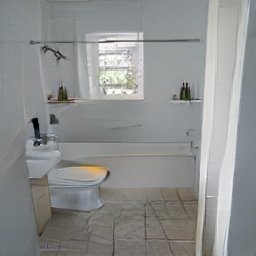}\\
\includegraphics[width=\linewidth]{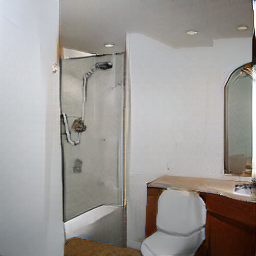}\\
\includegraphics[width=\linewidth]{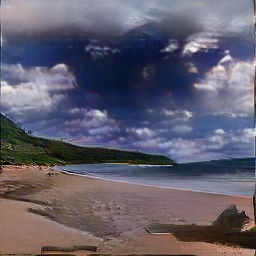}\\
\includegraphics[width=\linewidth]{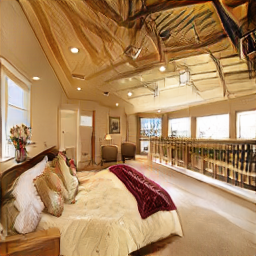}\\
\end{minipage}
} \hspace{-3mm}
\subfigure[Ours]{
\begin{minipage}[t]{0.13\linewidth}
\centering
\includegraphics[width=\linewidth]{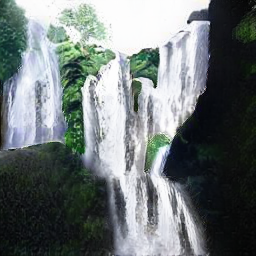}\\
\includegraphics[width=\linewidth]{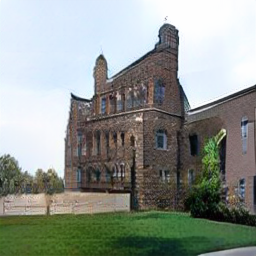}\\
\includegraphics[width=\linewidth]{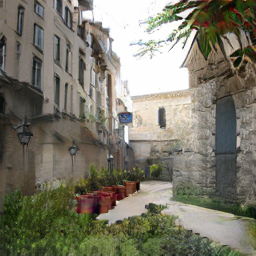}\\
\includegraphics[width=\linewidth]{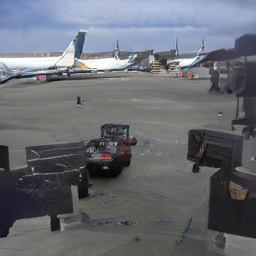}\\
\includegraphics[width=\linewidth]{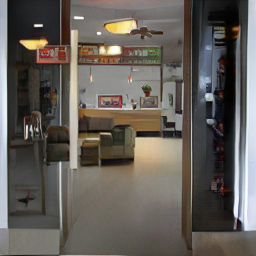}\\
\includegraphics[width=\linewidth]{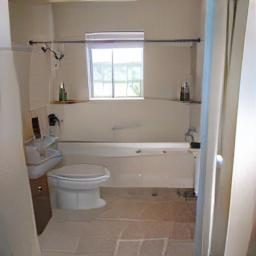}\\
\includegraphics[width=\linewidth]{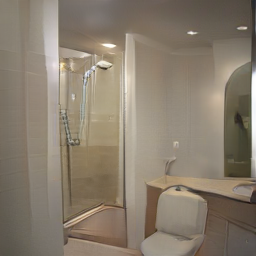}\\
\includegraphics[width=\linewidth]{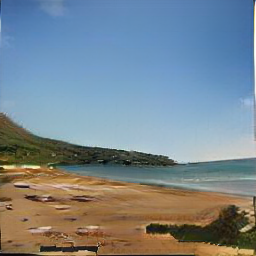}\\
\includegraphics[width=\linewidth]{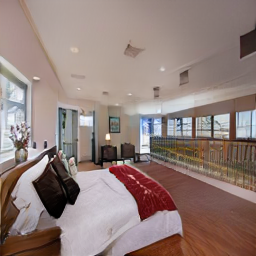}\\
\end{minipage}
} 
\caption{Qualitative results for ADE20K.}
\label{fig:qvis_cityscapes}
\end{figure*}

\begin{figure*}[htbp]
\centering
\subfigure[Label]{
\begin{minipage}[t]{0.13\linewidth}
\centering
\includegraphics[width=\linewidth,height=1.3cm]{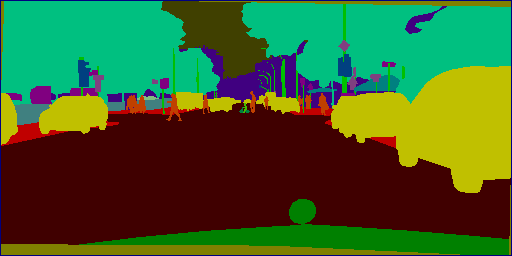}\\
\includegraphics[width=\linewidth,height=1.3cm]{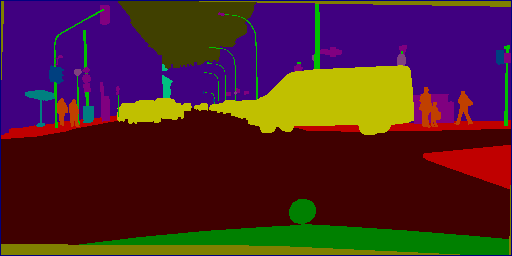}\\
\includegraphics[width=\linewidth,height=1.3cm]{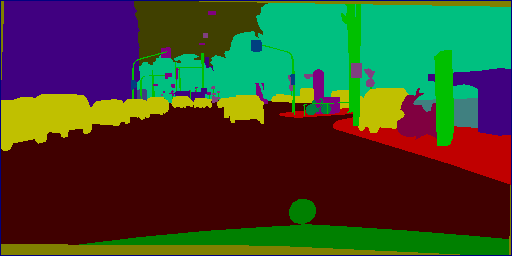}\\
\includegraphics[width=\linewidth,height=1.3cm]{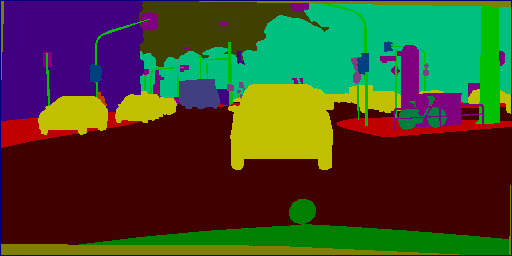}\\
\includegraphics[width=\linewidth,height=1.3cm]{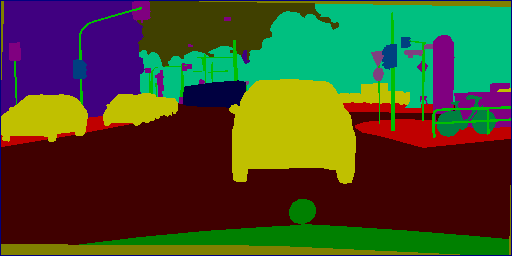}\\
\includegraphics[width=\linewidth,height=1.3cm]{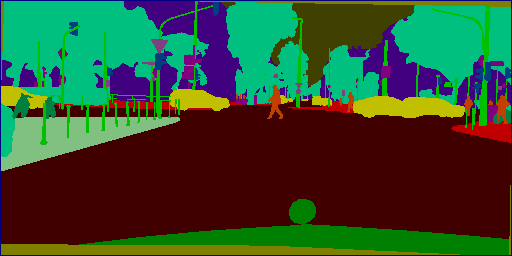}\\
\includegraphics[width=\linewidth,height=1.3cm]{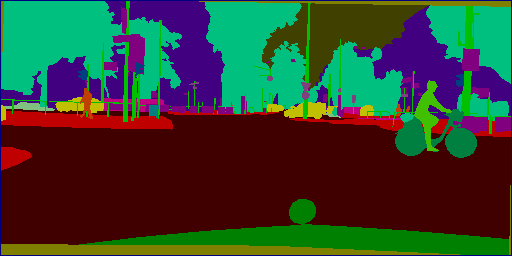}\\
\includegraphics[width=\linewidth,height=1.3cm]{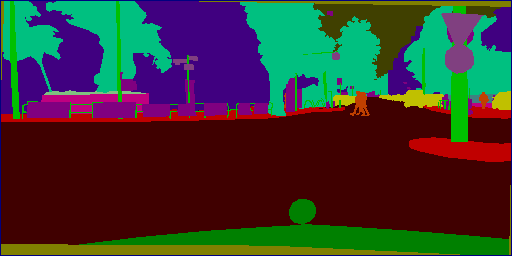}\\
\includegraphics[width=\linewidth,height=1.3cm]{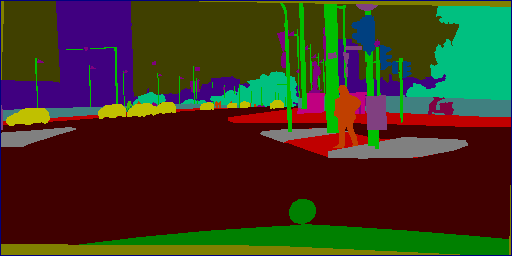}\\
\includegraphics[width=\linewidth,height=1.3cm]{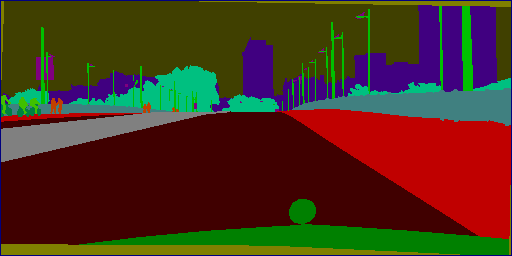}\\
\includegraphics[width=\linewidth,height=1.3cm]{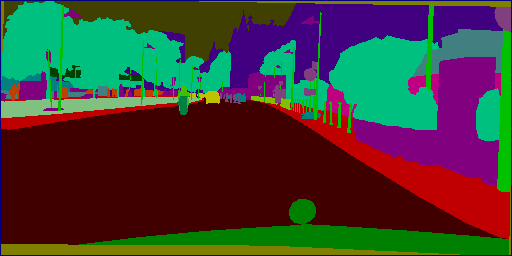}\\
\includegraphics[width=\linewidth,height=1.3cm]{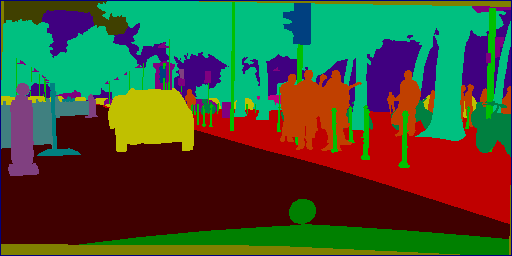}\\
\includegraphics[width=\linewidth,height=1.3cm]{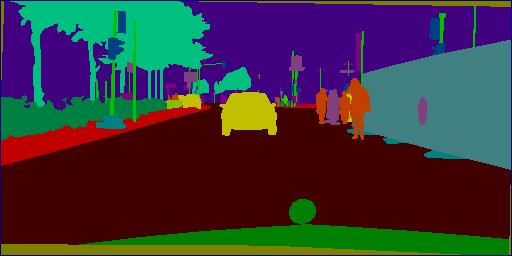}\\
\includegraphics[width=\linewidth,height=1.3cm]{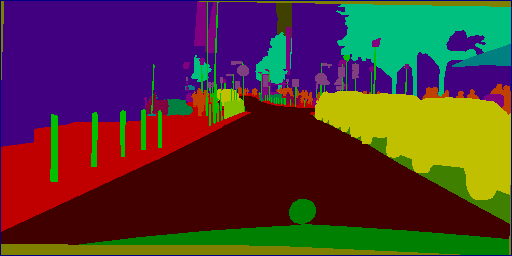}\\
\end{minipage}
}  \hspace*{-3mm}
\subfigure[CC-FPSE]{
\begin{minipage}[t]{0.13\linewidth}
\centering
\includegraphics[width=\linewidth,height=1.3cm]{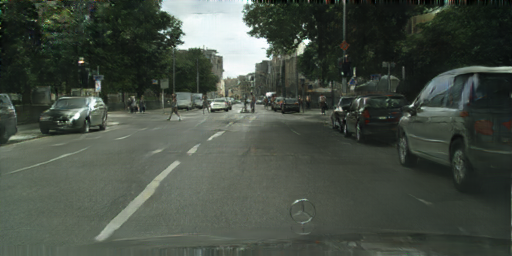}\\
\includegraphics[width=\linewidth,height=1.3cm]{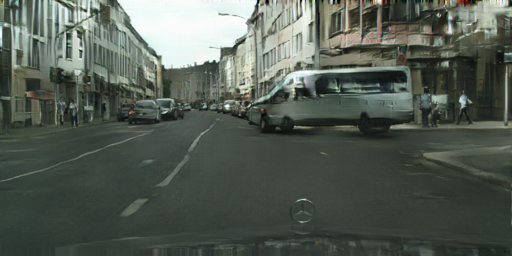}\\
\includegraphics[width=\linewidth,height=1.3cm]{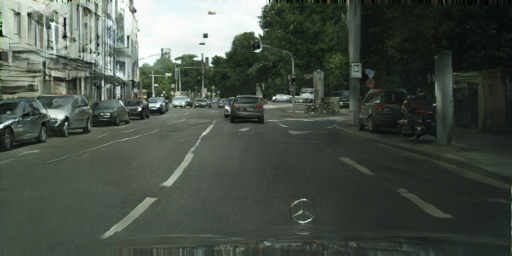}\\
\includegraphics[width=\linewidth,height=1.3cm]{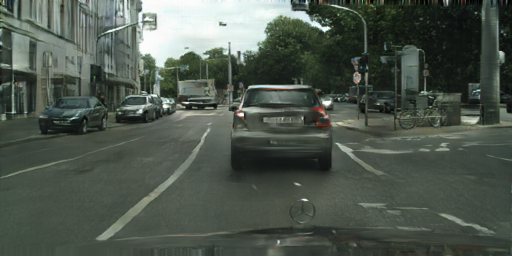}\\
\includegraphics[width=\linewidth,height=1.3cm]{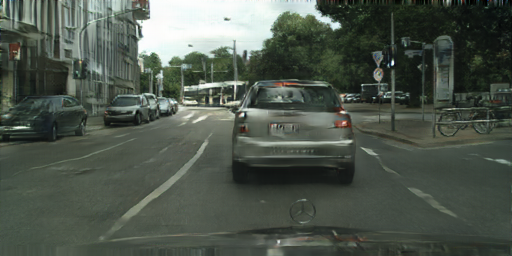}\\
\includegraphics[width=\linewidth,height=1.3cm]{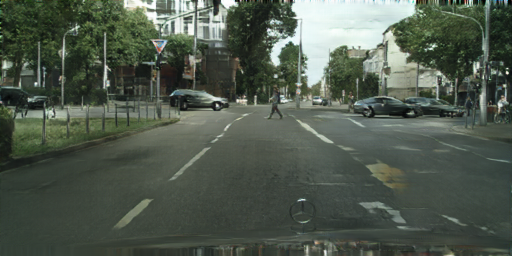}\\
\includegraphics[width=\linewidth,height=1.3cm]{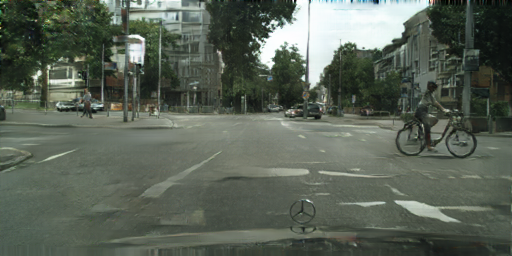}\\
\includegraphics[width=\linewidth,height=1.3cm]{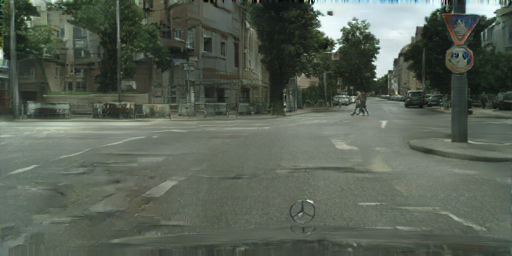}\\
\includegraphics[width=\linewidth,height=1.3cm]{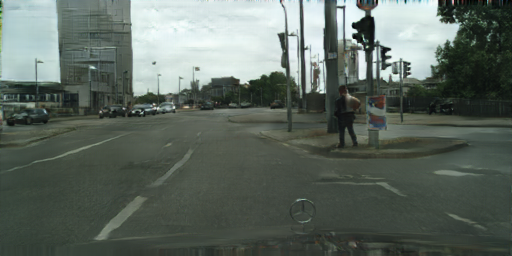}\\
\includegraphics[width=\linewidth,height=1.3cm]{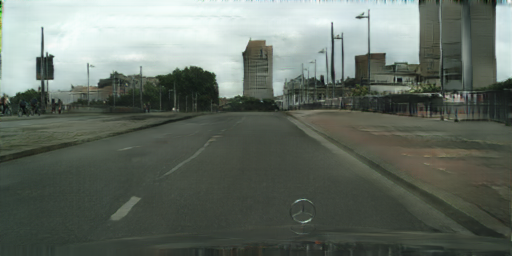}\\
\includegraphics[width=\linewidth,height=1.3cm]{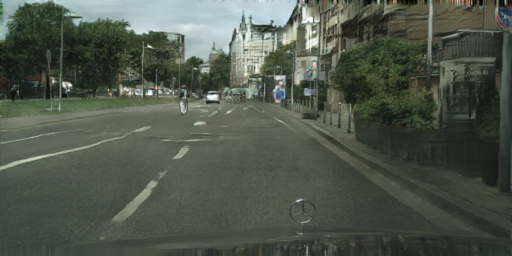}\\
\includegraphics[width=\linewidth,height=1.3cm]{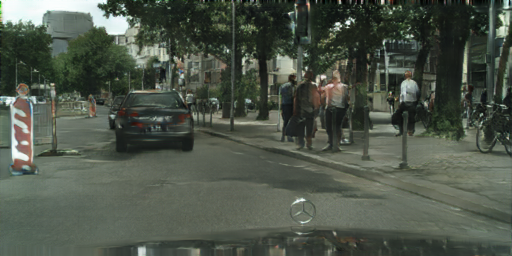}\\
\includegraphics[width=\linewidth,height=1.3cm]{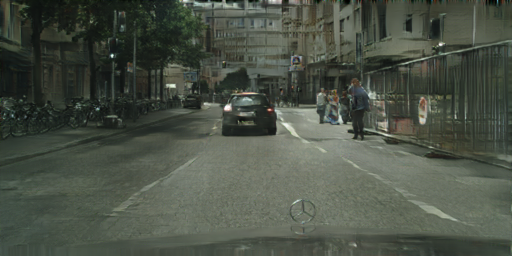}\\
\includegraphics[width=\linewidth,height=1.3cm]{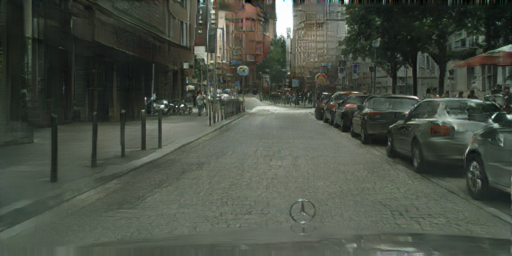}\\
\end{minipage}
} \hspace*{-3mm}
\subfigure[DAGAN]{
\begin{minipage}[t]{0.13\linewidth}
\centering
\includegraphics[width=\linewidth,height=1.3cm]{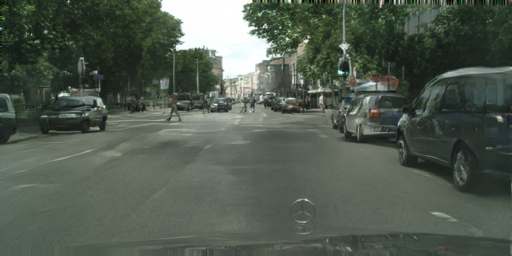}\\
\includegraphics[width=\linewidth,height=1.3cm]{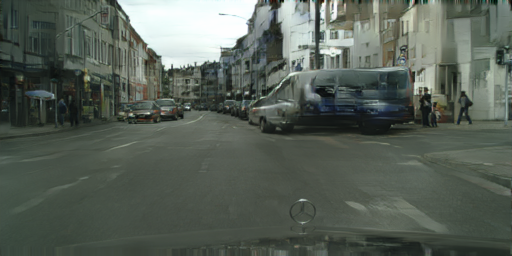}\\
\includegraphics[width=\linewidth,height=1.3cm]{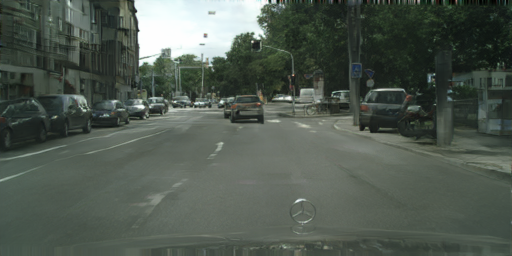}\\
\includegraphics[width=\linewidth,height=1.3cm]{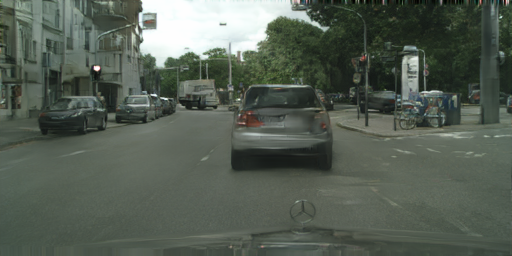}\\
\includegraphics[width=\linewidth,height=1.3cm]{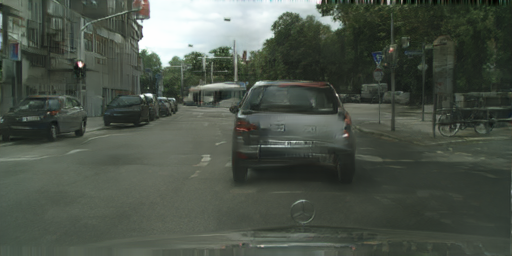}\\
\includegraphics[width=\linewidth,height=1.3cm]{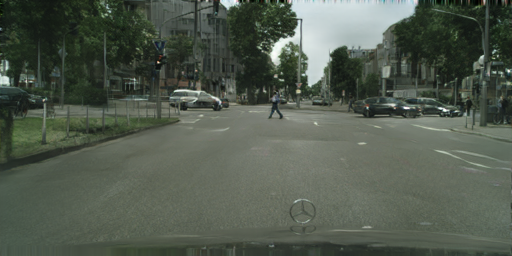}\\
\includegraphics[width=\linewidth,height=1.3cm]{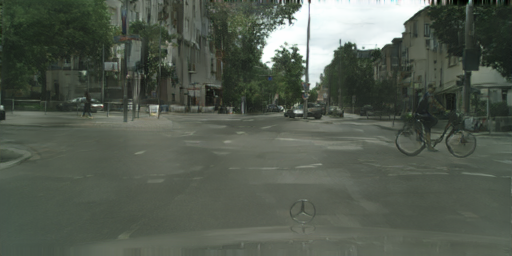}\\
\includegraphics[width=\linewidth,height=1.3cm]{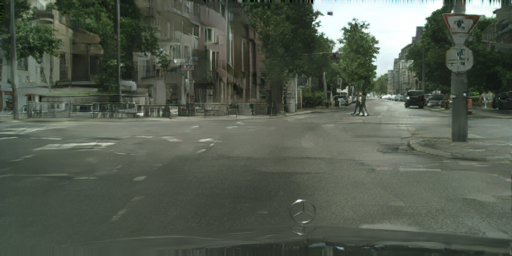}\\
\includegraphics[width=\linewidth,height=1.3cm]{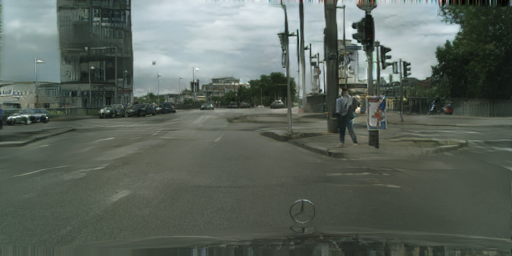}\\
\includegraphics[width=\linewidth,height=1.3cm]{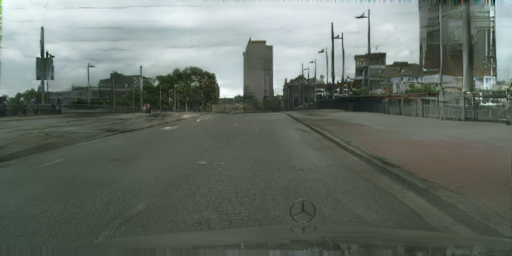}\\
\includegraphics[width=\linewidth,height=1.3cm]{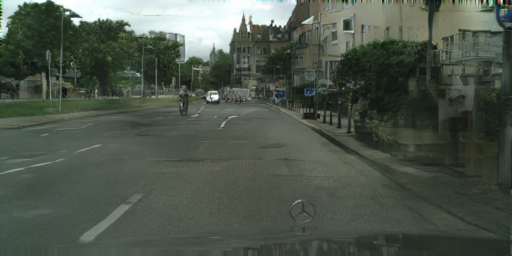}\\
\includegraphics[width=\linewidth,height=1.3cm]{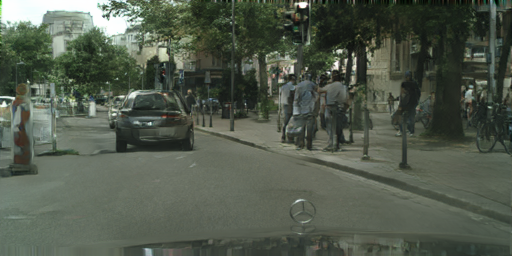}\\
\includegraphics[width=\linewidth,height=1.3cm]{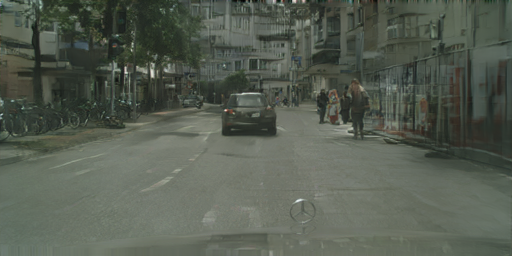}\\
\includegraphics[width=\linewidth,height=1.3cm]{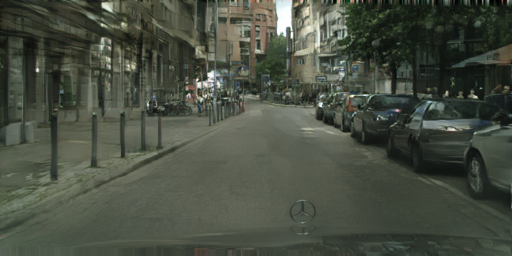}\\
\end{minipage}
} \hspace*{-3mm}
\subfigure[LGGAN]{
\begin{minipage}[t]{0.13\linewidth}
\centering
\includegraphics[width=\linewidth,height=1.3cm]{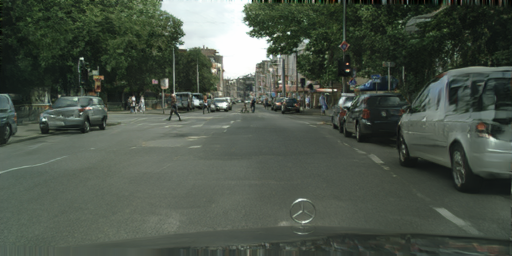}\\
\includegraphics[width=\linewidth,height=1.3cm]{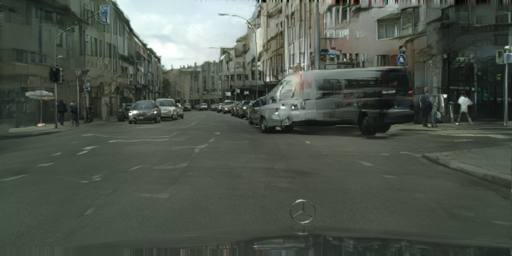}\\
\includegraphics[width=\linewidth,height=1.3cm]{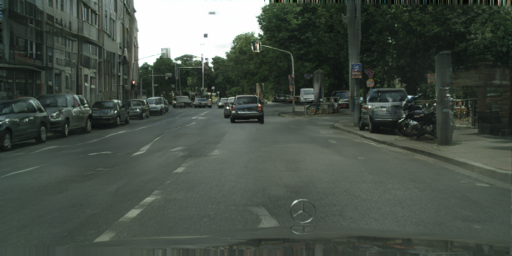}\\
\includegraphics[width=\linewidth,height=1.3cm]{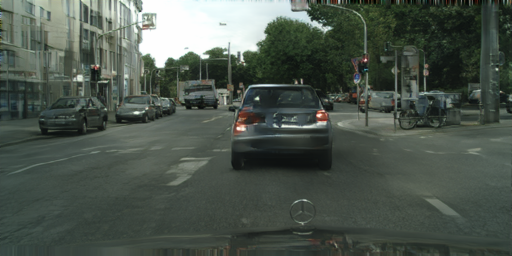}\\
\includegraphics[width=\linewidth,height=1.3cm]{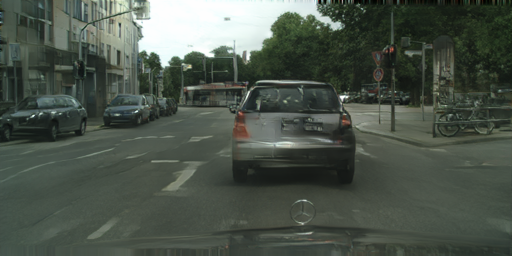}\\
\includegraphics[width=\linewidth,height=1.3cm]{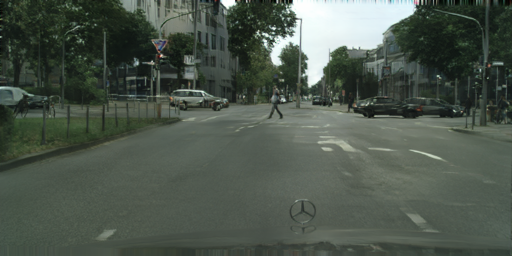}\\
\includegraphics[width=\linewidth,height=1.3cm]{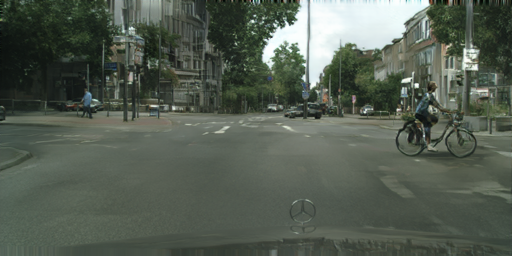}\\
\includegraphics[width=\linewidth,height=1.3cm]{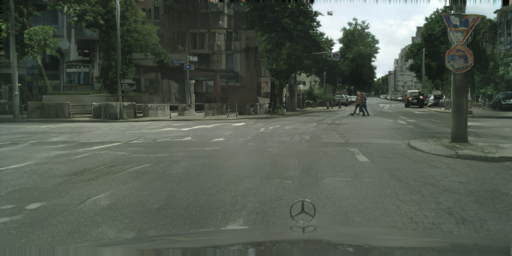}\\
\includegraphics[width=\linewidth,height=1.3cm]{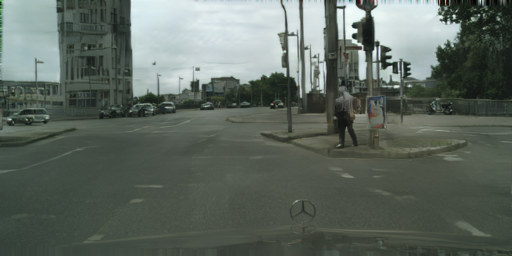}\\
\includegraphics[width=\linewidth,height=1.3cm]{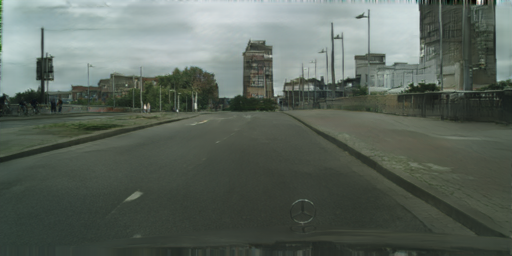}\\
\includegraphics[width=\linewidth,height=1.3cm]{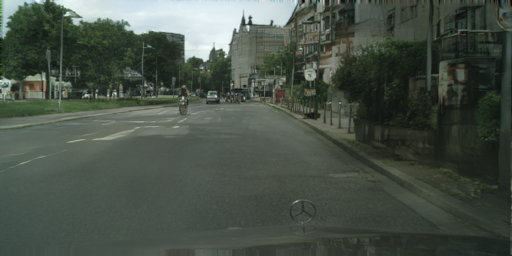}\\
\includegraphics[width=\linewidth,height=1.3cm]{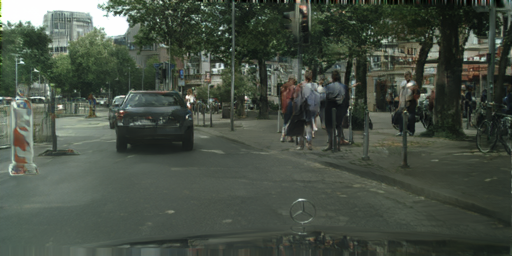}\\
\includegraphics[width=\linewidth,height=1.3cm]{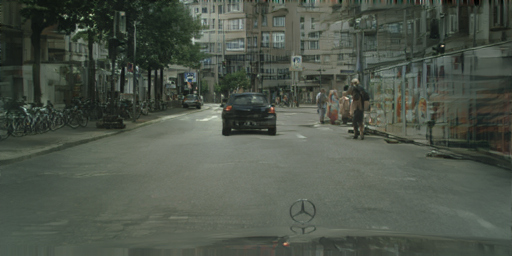}\\
\includegraphics[width=\linewidth,height=1.3cm]{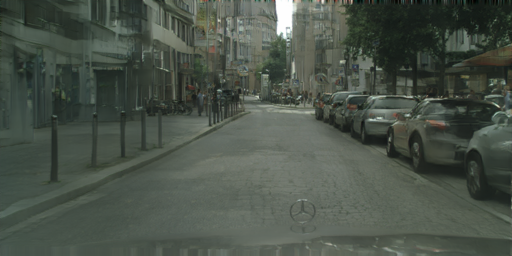}\\
\end{minipage}
} \hspace*{-3mm}
\subfigure[SPADE]{
\begin{minipage}[t]{0.13\linewidth}
\centering
\includegraphics[width=\linewidth,height=1.3cm]{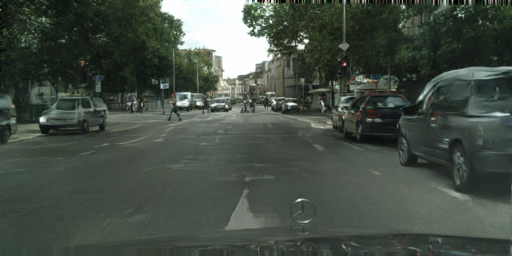}\\
\includegraphics[width=\linewidth,height=1.3cm]{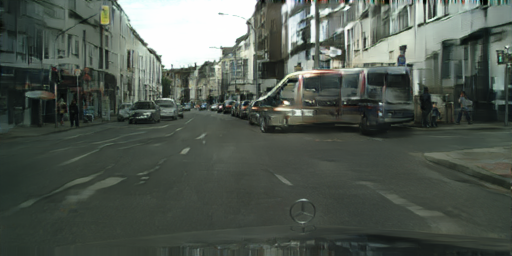}\\
\includegraphics[width=\linewidth,height=1.3cm]{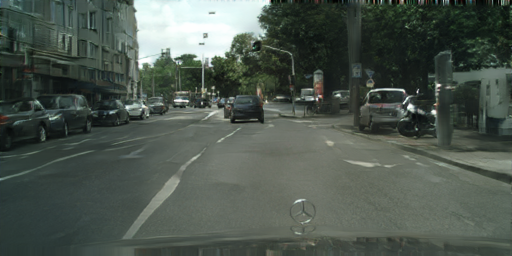}\\
\includegraphics[width=\linewidth,height=1.3cm]{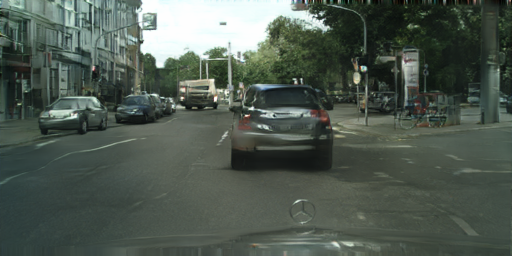}\\
\includegraphics[width=\linewidth,height=1.3cm]{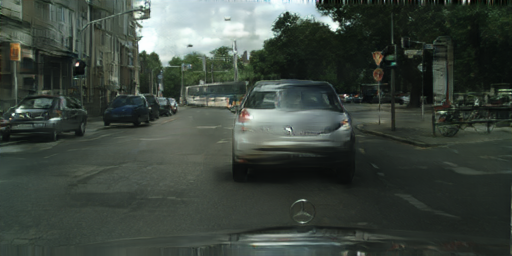}\\
\includegraphics[width=\linewidth,height=1.3cm]{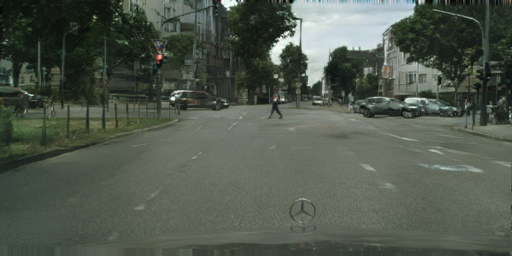}\\
\includegraphics[width=\linewidth,height=1.3cm]{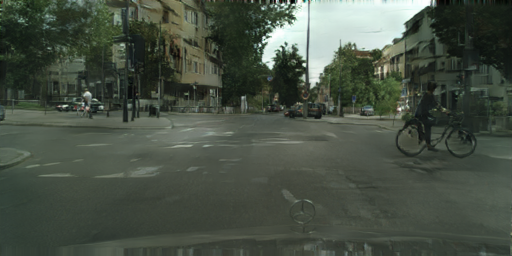}\\
\includegraphics[width=\linewidth,height=1.3cm]{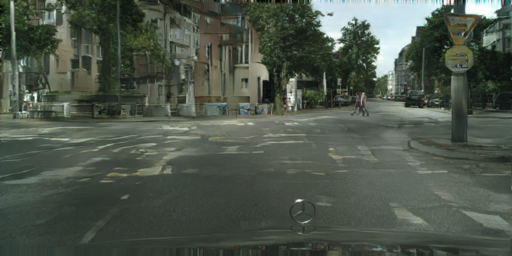}\\
\includegraphics[width=\linewidth,height=1.3cm]{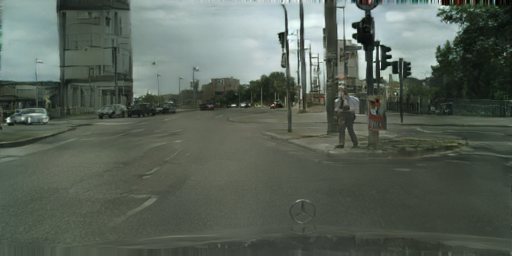}\\
\includegraphics[width=\linewidth,height=1.3cm]{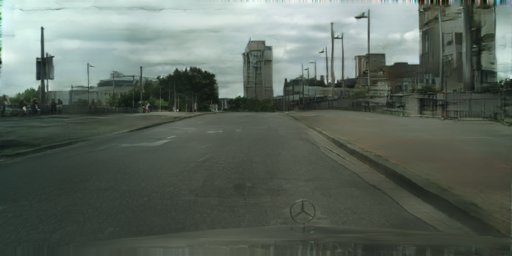}\\
\includegraphics[width=\linewidth,height=1.3cm]{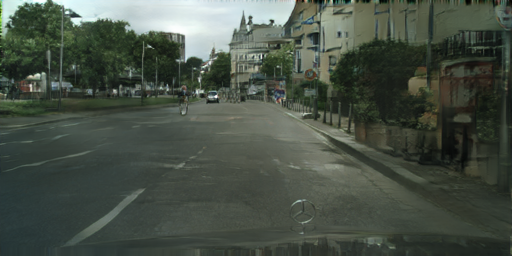}\\
\includegraphics[width=\linewidth,height=1.3cm]{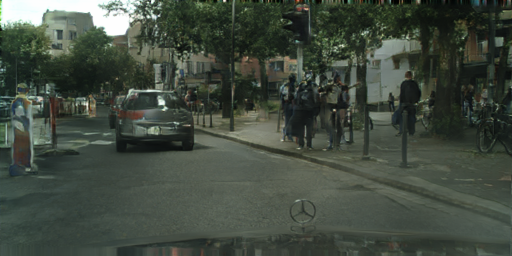}\\
\includegraphics[width=\linewidth,height=1.3cm]{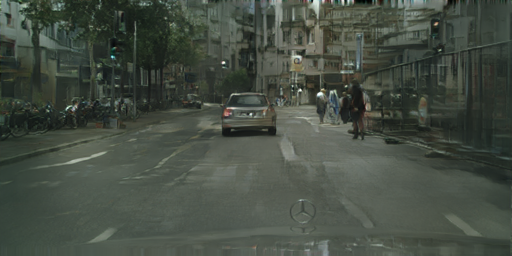}\\
\includegraphics[width=\linewidth,height=1.3cm]{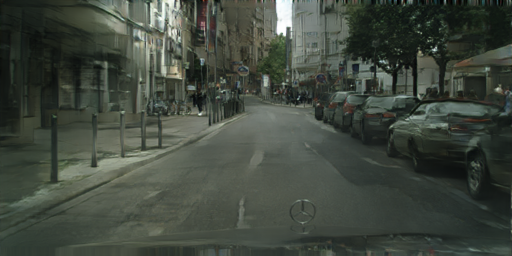}\\
\end{minipage}
} \hspace*{-3mm}
\subfigure[OASIS]{
\begin{minipage}[t]{0.13\linewidth}
\centering
\includegraphics[width=\linewidth,height=1.3cm]{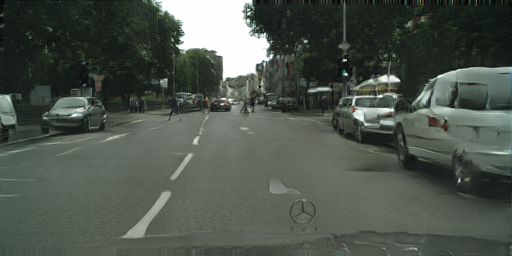}\\
\includegraphics[width=\linewidth,height=1.3cm]{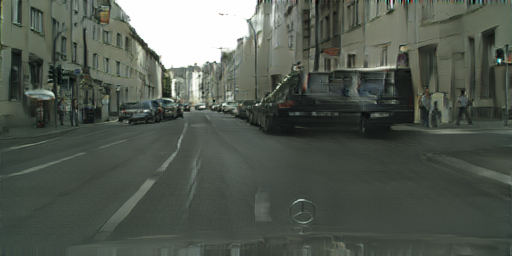}\\
\includegraphics[width=\linewidth,height=1.3cm]{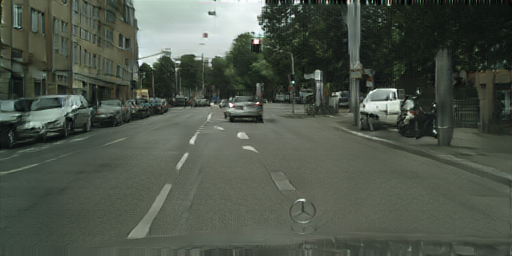}\\
\includegraphics[width=\linewidth,height=1.3cm]{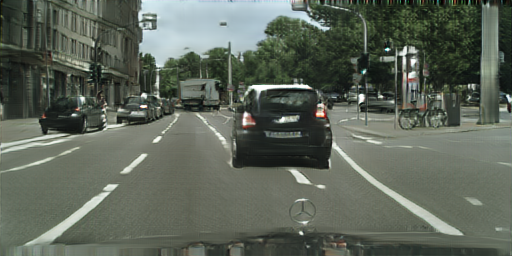}\\
\includegraphics[width=\linewidth,height=1.3cm]{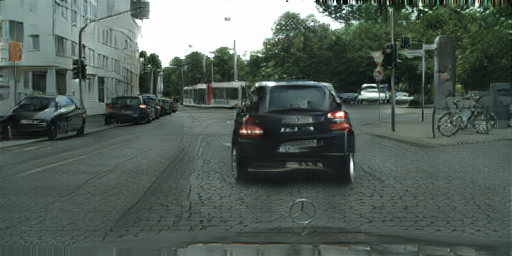}\\
\includegraphics[width=\linewidth,height=1.3cm]{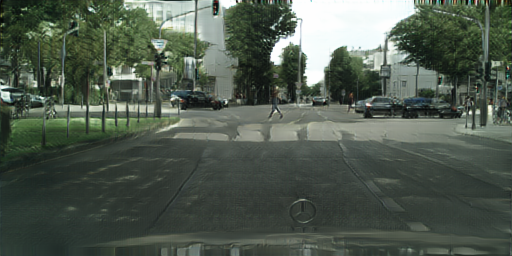}\\
\includegraphics[width=\linewidth,height=1.3cm]{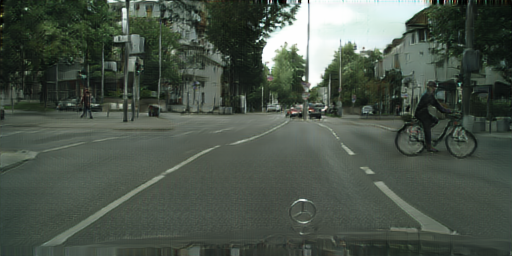}\\
\includegraphics[width=\linewidth,height=1.3cm]{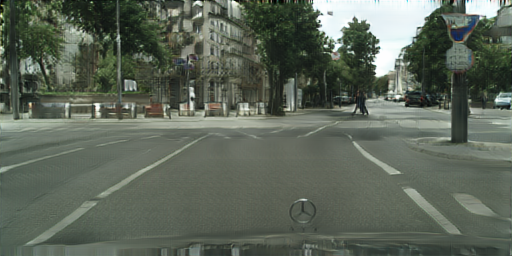}\\
\includegraphics[width=\linewidth,height=1.3cm]{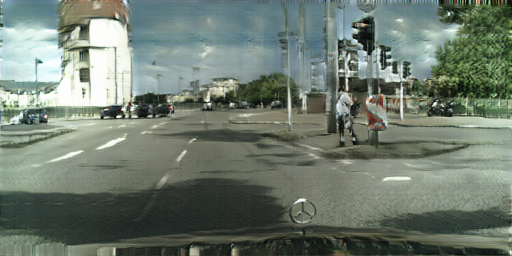}\\
\includegraphics[width=\linewidth,height=1.3cm]{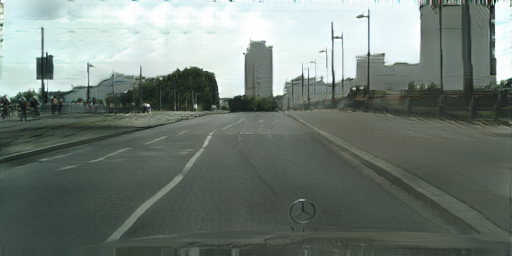}\\
\includegraphics[width=\linewidth,height=1.3cm]{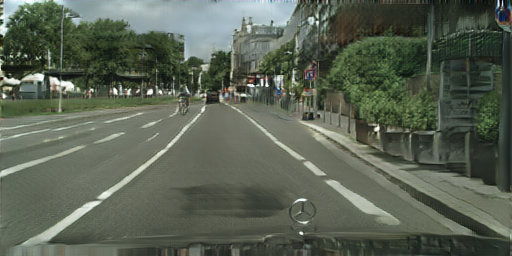}\\
\includegraphics[width=\linewidth,height=1.3cm]{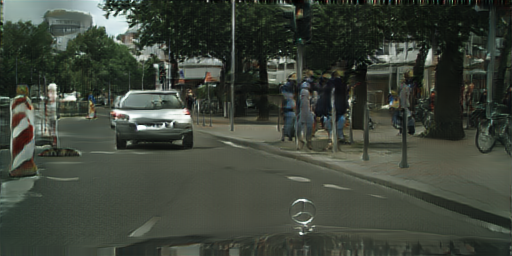}\\
\includegraphics[width=\linewidth,height=1.3cm]{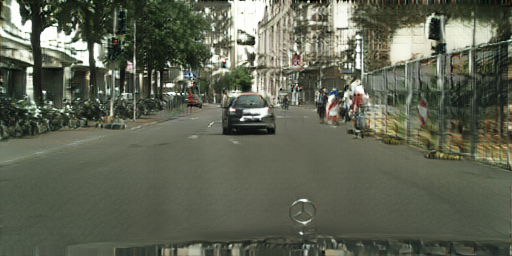}\\
\includegraphics[width=\linewidth,height=1.3cm]{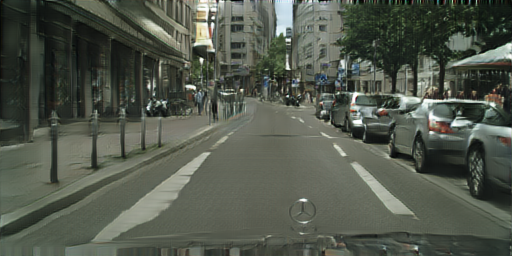}\\
\end{minipage}
} \hspace*{-3mm}
\subfigure[Ours]{
\begin{minipage}[t]{0.13\linewidth}
\centering
\includegraphics[width=\linewidth,height=1.3cm]{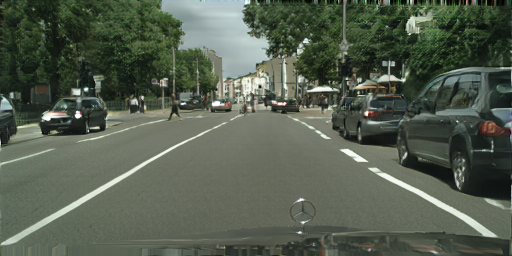}\\
\includegraphics[width=\linewidth,height=1.3cm]{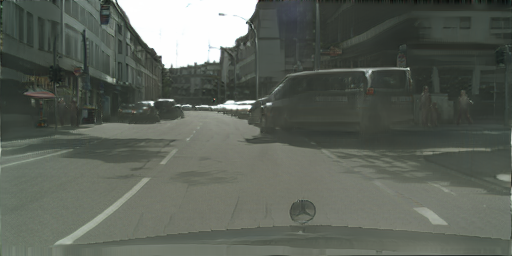}\\
\includegraphics[width=\linewidth,height=1.3cm]{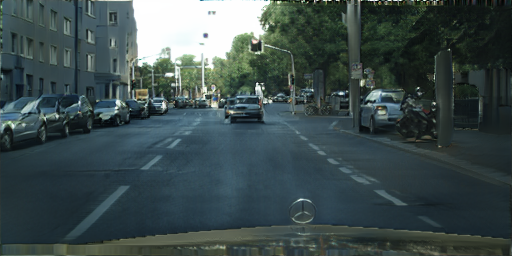}\\
\includegraphics[width=\linewidth,height=1.3cm]{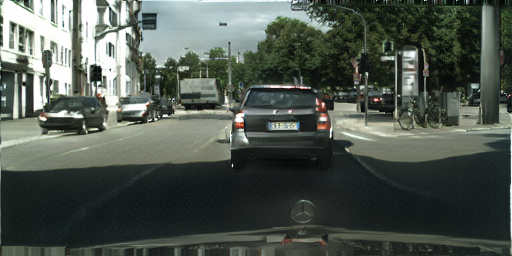}\\
\includegraphics[width=\linewidth,height=1.3cm]{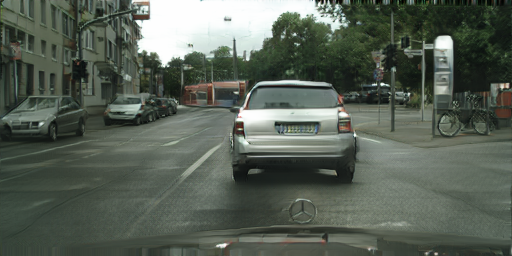}\\
\includegraphics[width=\linewidth,height=1.3cm]{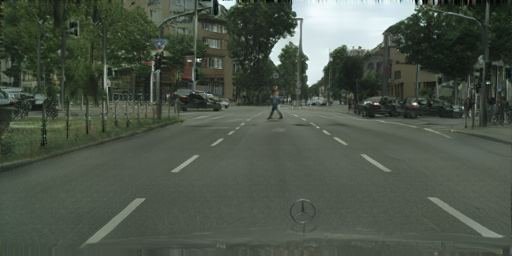}\\
\includegraphics[width=\linewidth,height=1.3cm]{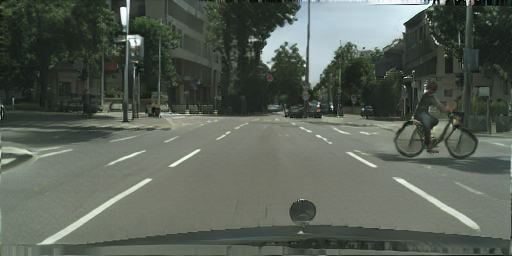}\\
\includegraphics[width=\linewidth,height=1.3cm]{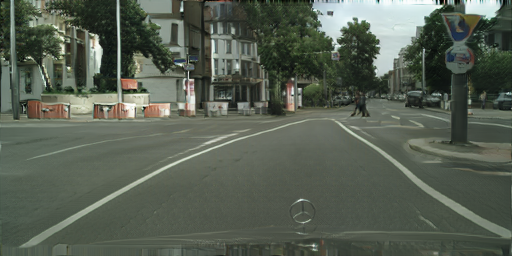}\\
\includegraphics[width=\linewidth,height=1.3cm]{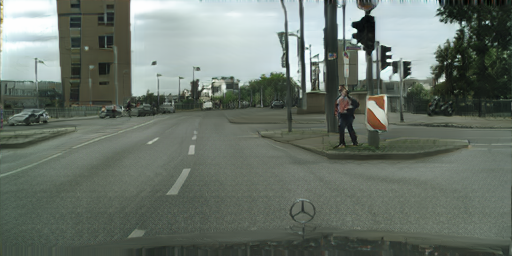}\\
\includegraphics[width=\linewidth,height=1.3cm]{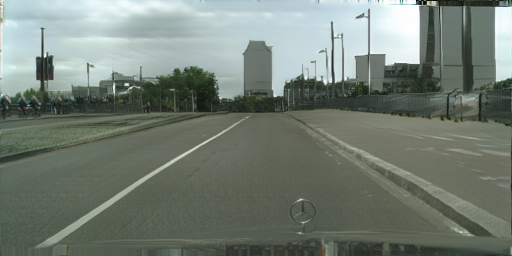}\\
\includegraphics[width=\linewidth,height=1.3cm]{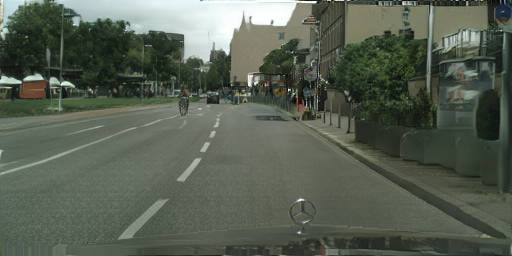}\\
\includegraphics[width=\linewidth,height=1.3cm]{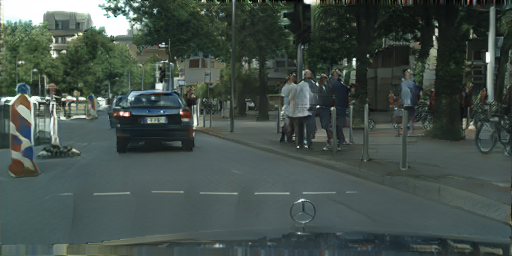}\\
\includegraphics[width=\linewidth,height=1.3cm]{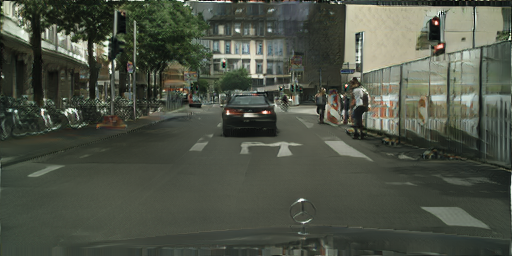}\\
\includegraphics[width=\linewidth,height=1.3cm]{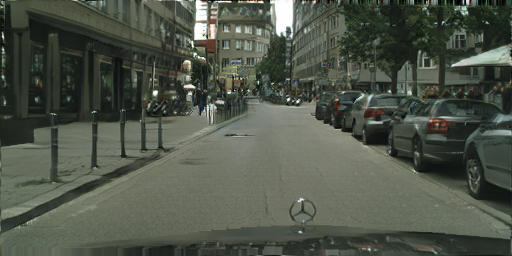}\\
\end{minipage}
} 
\caption{Qualitative results for Cityscapes.}
\label{fig:qvis_cityscapes}
\end{figure*}

\begin{figure*}[htbp]
\centering
\subfigure[Label]{
\begin{minipage}[t]{0.13\linewidth}
\centering
\includegraphics[width=\linewidth,height=1.3cm]{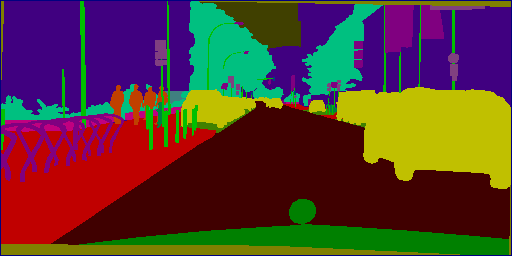}\\
\includegraphics[width=\linewidth,height=1.3cm]{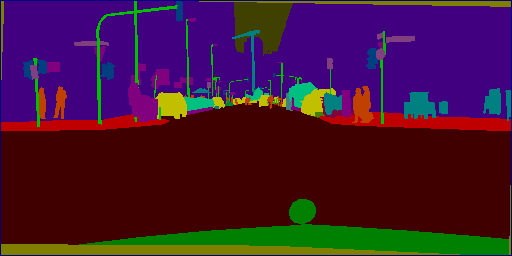}\\
\includegraphics[width=\linewidth,height=1.3cm]{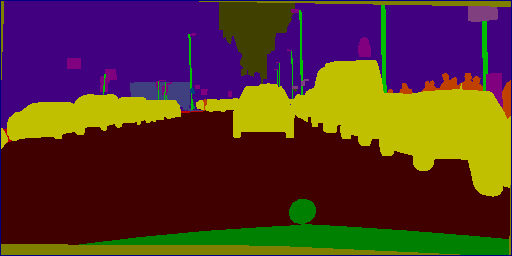}\\
\includegraphics[width=\linewidth,height=1.3cm]{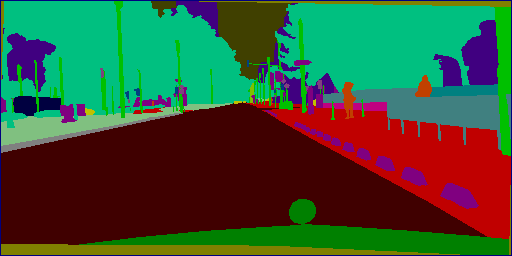}\\
\includegraphics[width=\linewidth,height=1.3cm]{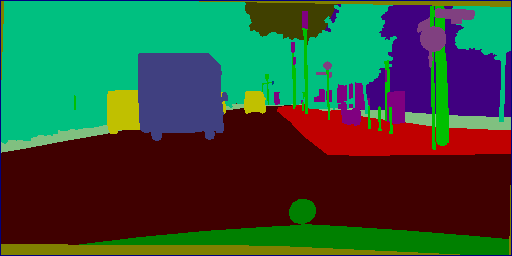}\\
\includegraphics[width=\linewidth,height=1.3cm]{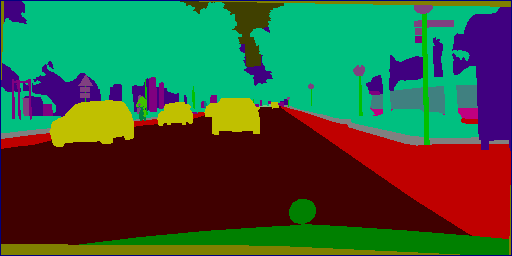}\\
\includegraphics[width=\linewidth,height=1.3cm]{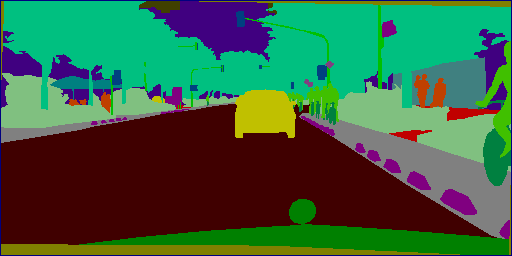}\\
\includegraphics[width=\linewidth,height=1.3cm]{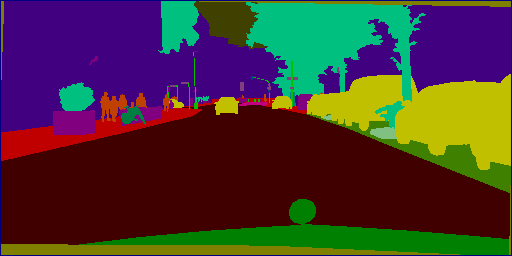}\\
\includegraphics[width=\linewidth,height=1.3cm]{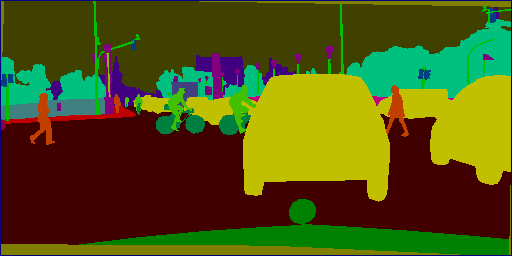}\\
\includegraphics[width=\linewidth,height=1.3cm]{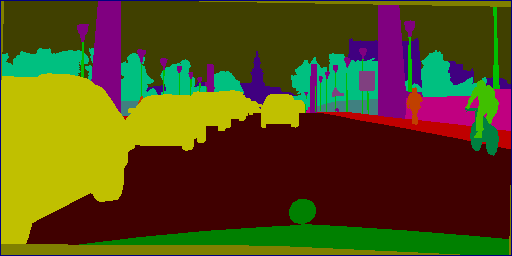}\\
\includegraphics[width=\linewidth,height=1.3cm]{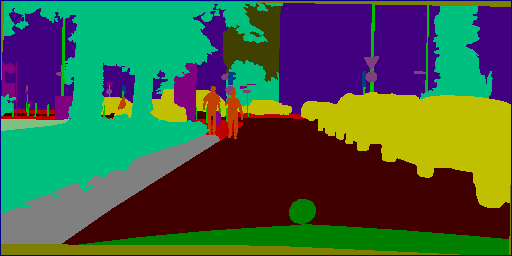}\\
\includegraphics[width=\linewidth,height=1.3cm]{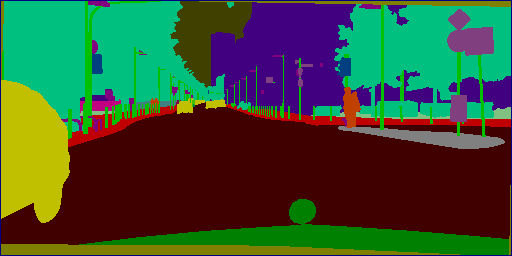}\\
\includegraphics[width=\linewidth,height=1.3cm]{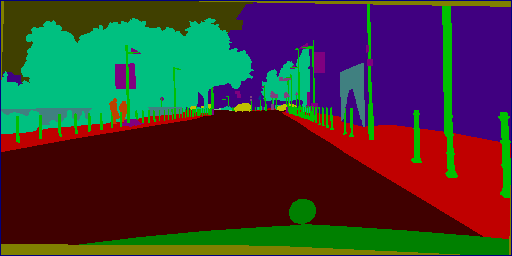}\\
\includegraphics[width=\linewidth,height=1.3cm]{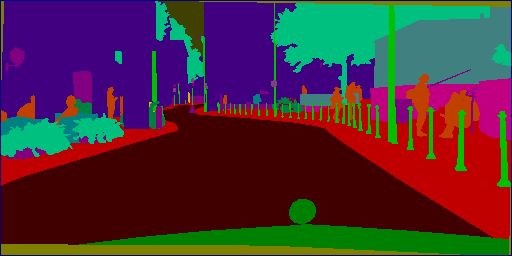}\\
\end{minipage}
}  \hspace*{-3mm}
\subfigure[CC-FPSE]{
\begin{minipage}[t]{0.13\linewidth}
\centering
\includegraphics[width=\linewidth,height=1.3cm]{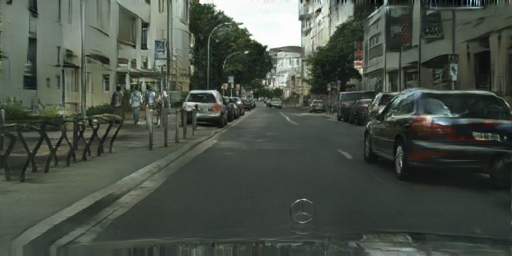}\\
\includegraphics[width=\linewidth,height=1.3cm]{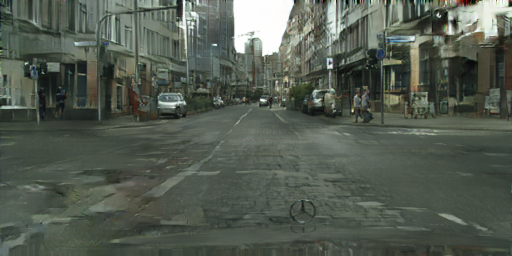}\\
\includegraphics[width=\linewidth,height=1.3cm]{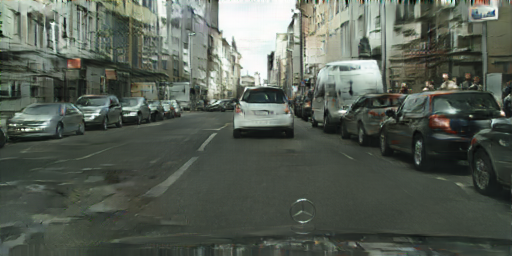}\\
\includegraphics[width=\linewidth,height=1.3cm]{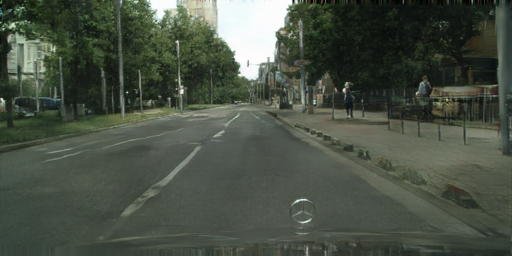}\\
\includegraphics[width=\linewidth,height=1.3cm]{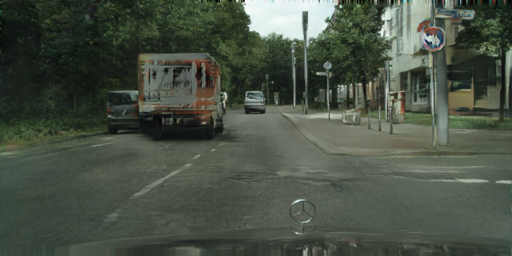}\\
\includegraphics[width=\linewidth,height=1.3cm]{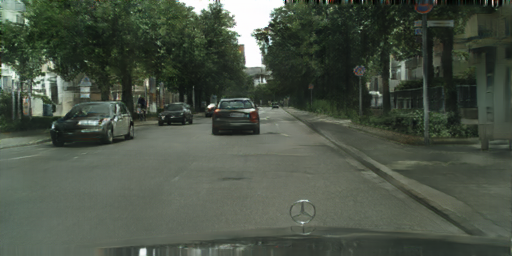}\\
\includegraphics[width=\linewidth,height=1.3cm]{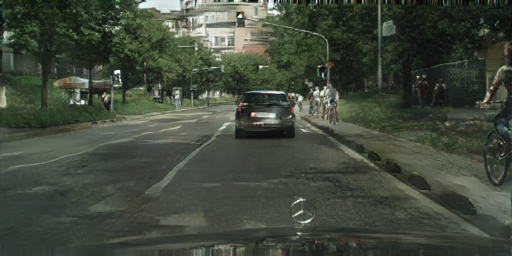}\\
\includegraphics[width=\linewidth,height=1.3cm]{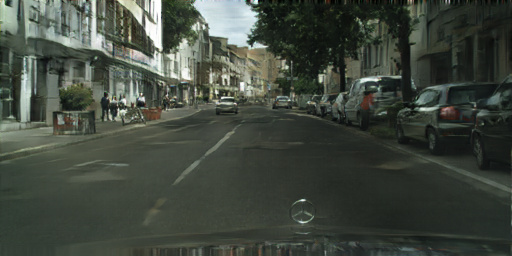}\\
\includegraphics[width=\linewidth,height=1.3cm]{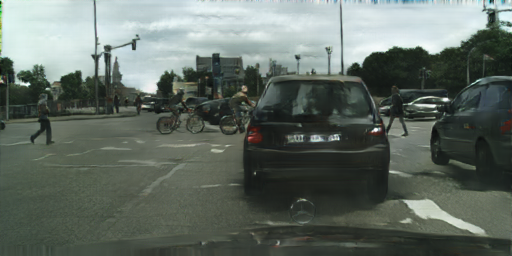}\\
\includegraphics[width=\linewidth,height=1.3cm]{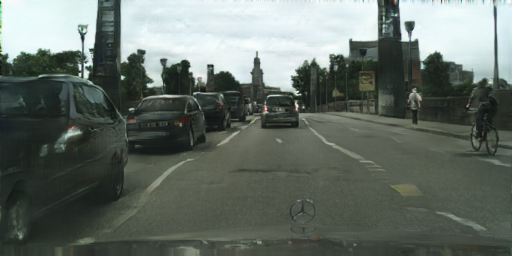}\\
\includegraphics[width=\linewidth,height=1.3cm]{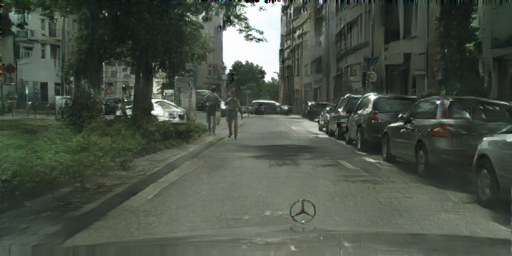}\\
\includegraphics[width=\linewidth,height=1.3cm]{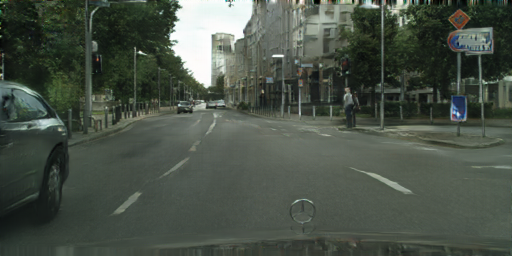}\\
\includegraphics[width=\linewidth,height=1.3cm]{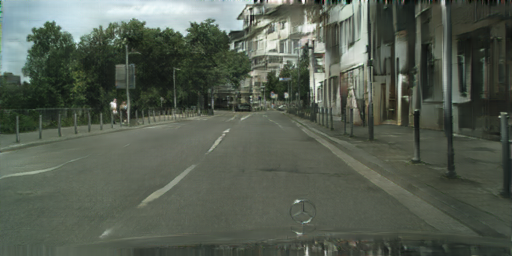}\\
\includegraphics[width=\linewidth,height=1.3cm]{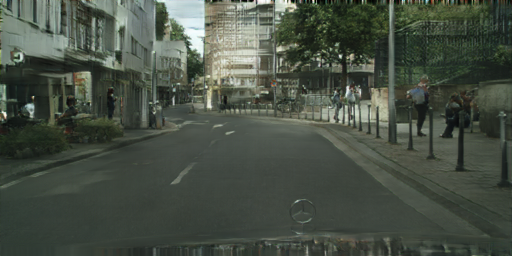}\\
\end{minipage}
} \hspace*{-3mm}
\subfigure[DAGAN]{
\begin{minipage}[t]{0.13\linewidth}
\centering
\includegraphics[width=\linewidth,height=1.3cm]{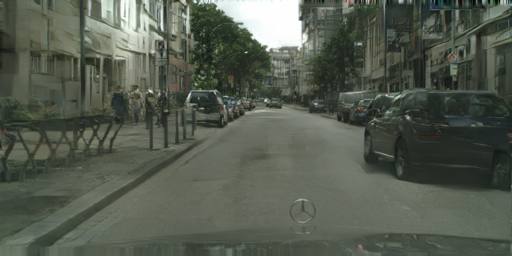}\\
\includegraphics[width=\linewidth,height=1.3cm]{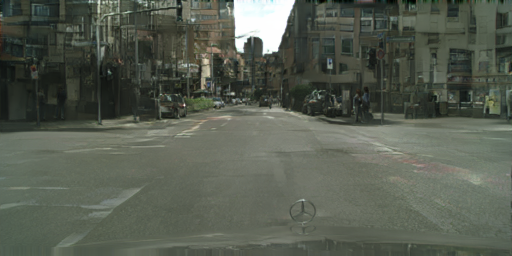}\\
\includegraphics[width=\linewidth,height=1.3cm]{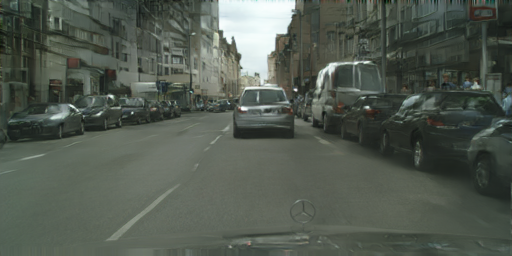}\\
\includegraphics[width=\linewidth,height=1.3cm]{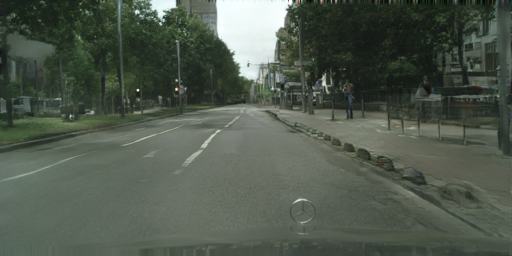}\\
\includegraphics[width=\linewidth,height=1.3cm]{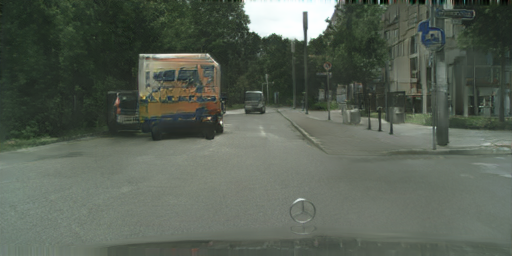}\\
\includegraphics[width=\linewidth,height=1.3cm]{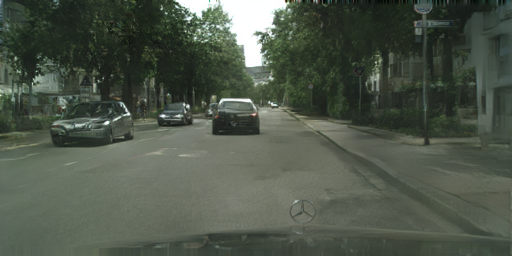}\\
\includegraphics[width=\linewidth,height=1.3cm]{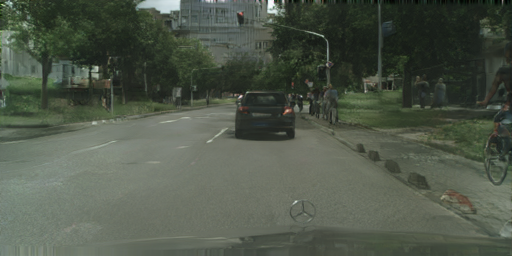}\\
\includegraphics[width=\linewidth,height=1.3cm]{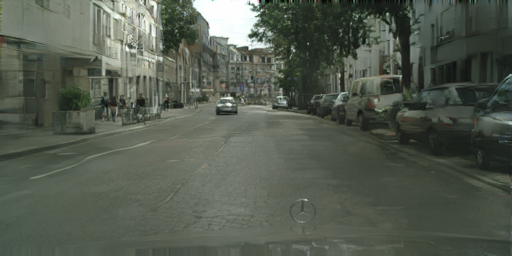}\\
\includegraphics[width=\linewidth,height=1.3cm]{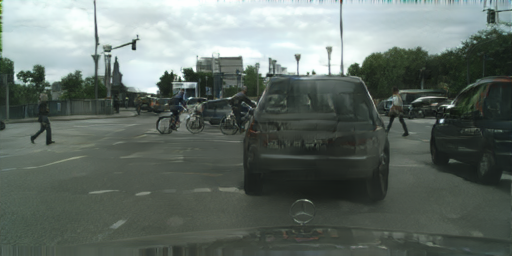}\\
\includegraphics[width=\linewidth,height=1.3cm]{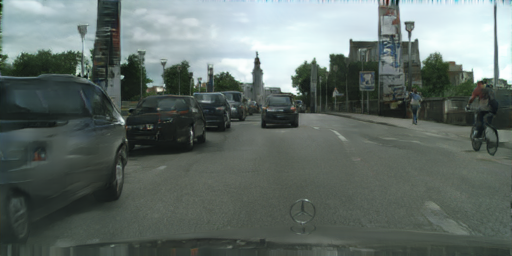}\\
\includegraphics[width=\linewidth,height=1.3cm]{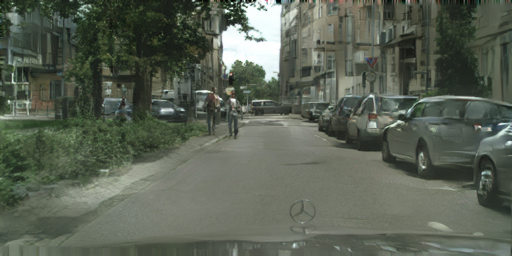}\\
\includegraphics[width=\linewidth,height=1.3cm]{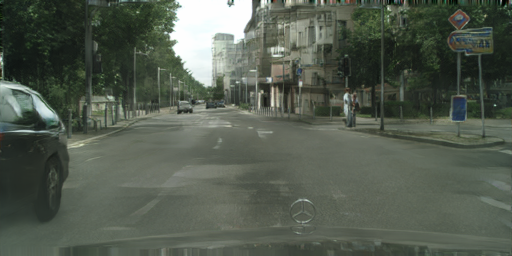}\\
\includegraphics[width=\linewidth,height=1.3cm]{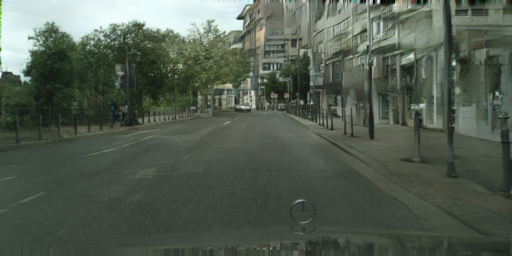}\\
\includegraphics[width=\linewidth,height=1.3cm]{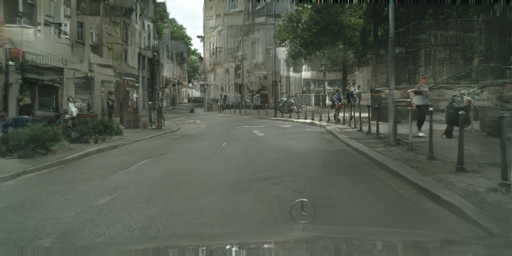}\\
\end{minipage}
} \hspace*{-3mm}
\subfigure[LGGAN]{
\begin{minipage}[t]{0.13\linewidth}
\centering
\includegraphics[width=\linewidth,height=1.3cm]{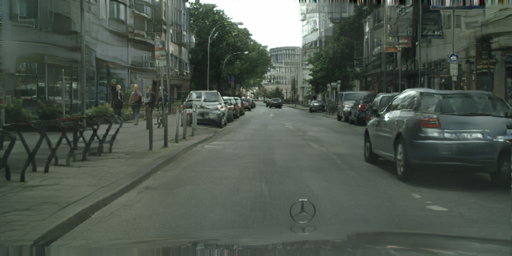}\\
\includegraphics[width=\linewidth,height=1.3cm]{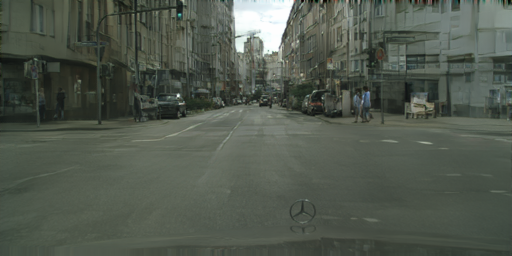}\\
\includegraphics[width=\linewidth,height=1.3cm]{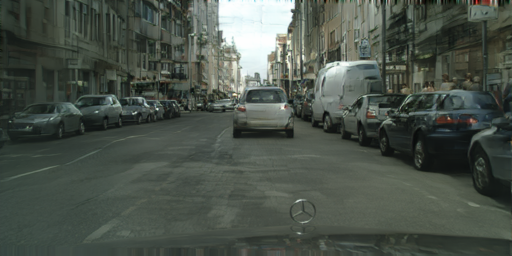}\\
\includegraphics[width=\linewidth,height=1.3cm]{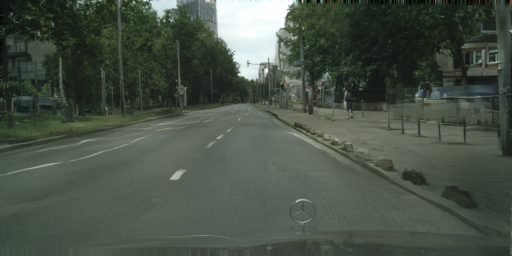}\\
\includegraphics[width=\linewidth,height=1.3cm]{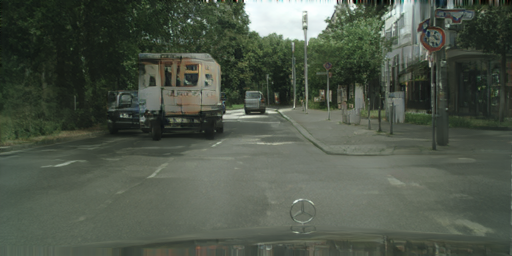}\\
\includegraphics[width=\linewidth,height=1.3cm]{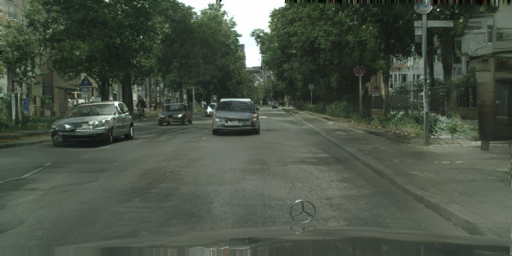}\\
\includegraphics[width=\linewidth,height=1.3cm]{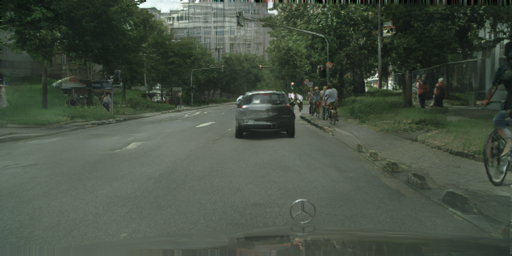}\\
\includegraphics[width=\linewidth,height=1.3cm]{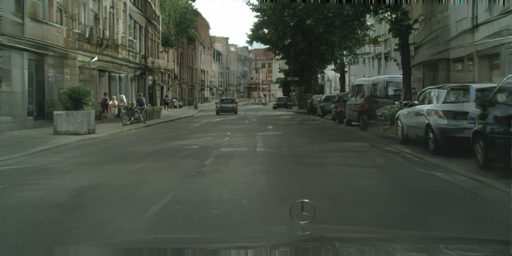}\\
\includegraphics[width=\linewidth,height=1.3cm]{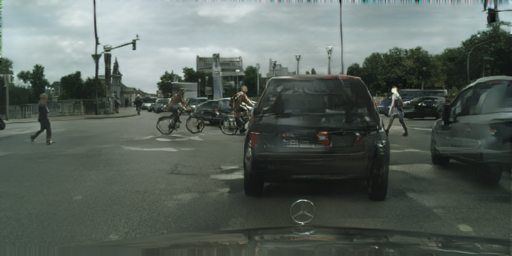}\\
\includegraphics[width=\linewidth,height=1.3cm]{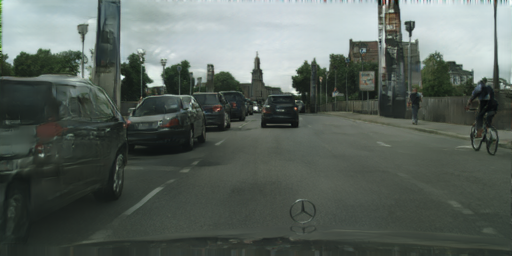}\\
\includegraphics[width=\linewidth,height=1.3cm]{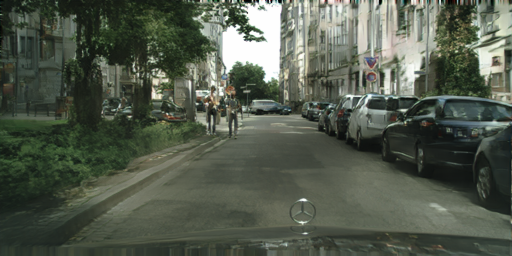}\\
\includegraphics[width=\linewidth,height=1.3cm]{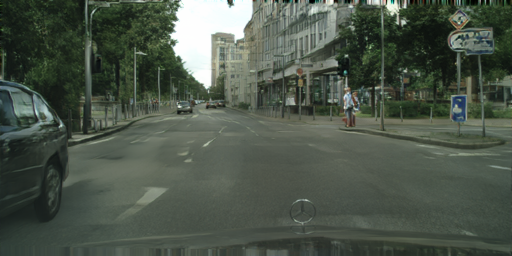}\\
\includegraphics[width=\linewidth,height=1.3cm]{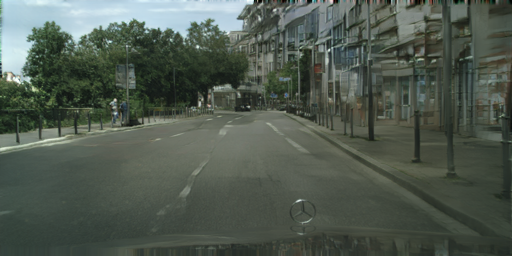}\\
\includegraphics[width=\linewidth,height=1.3cm]{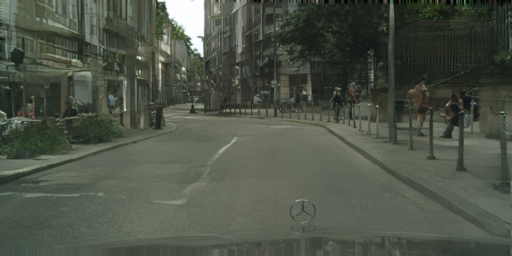}\\
\end{minipage}
} \hspace*{-3mm}
\subfigure[SPADE]{
\begin{minipage}[t]{0.13\linewidth}
\centering
\includegraphics[width=\linewidth,height=1.3cm]{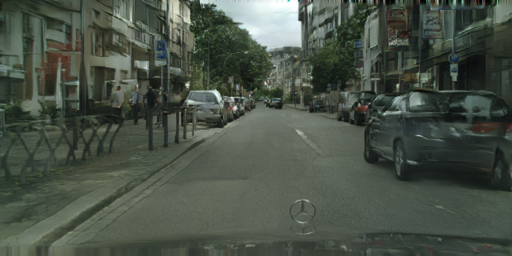}\\
\includegraphics[width=\linewidth,height=1.3cm]{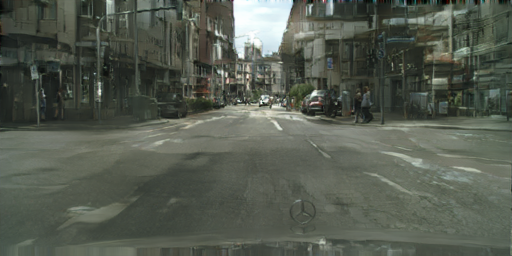}\\
\includegraphics[width=\linewidth,height=1.3cm]{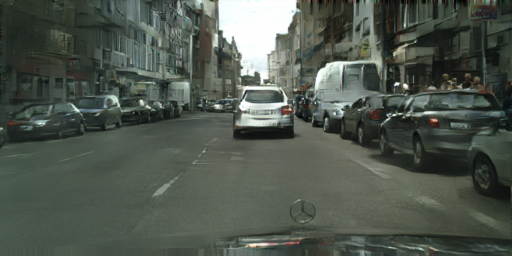}\\
\includegraphics[width=\linewidth,height=1.3cm]{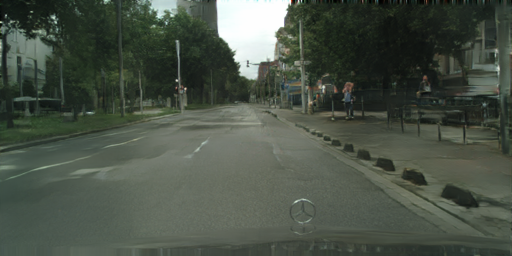}\\
\includegraphics[width=\linewidth,height=1.3cm]{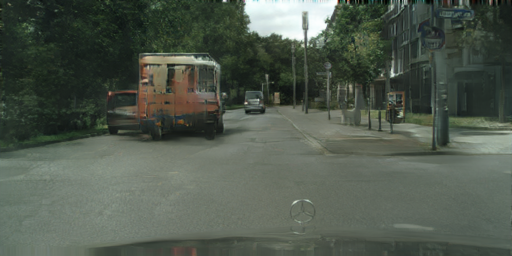}\\
\includegraphics[width=\linewidth,height=1.3cm]{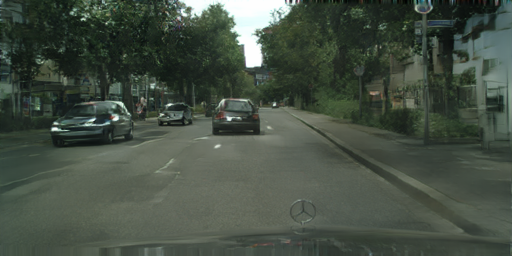}\\
\includegraphics[width=\linewidth,height=1.3cm]{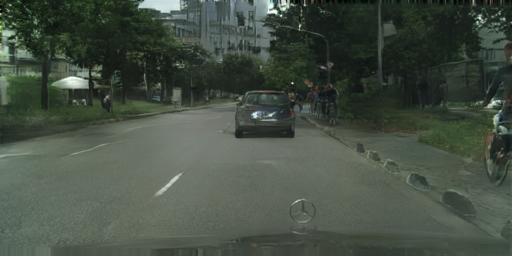}\\
\includegraphics[width=\linewidth,height=1.3cm]{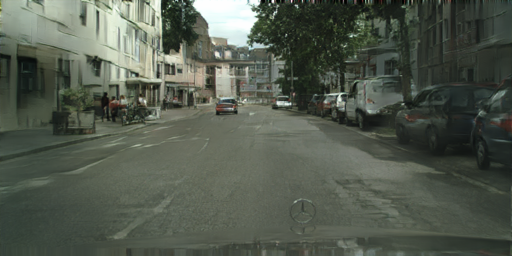}\\
\includegraphics[width=\linewidth,height=1.3cm]{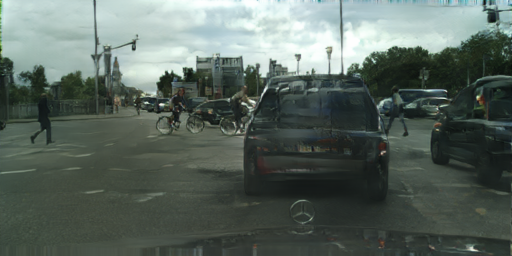}\\
\includegraphics[width=\linewidth,height=1.3cm]{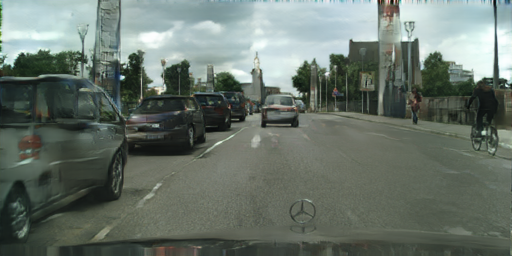}\\
\includegraphics[width=\linewidth,height=1.3cm]{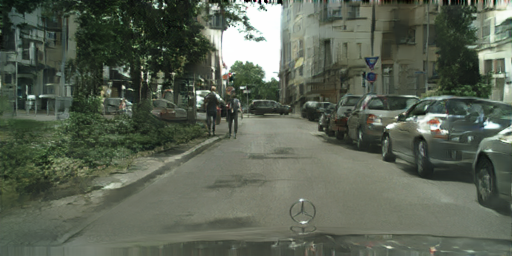}\\
\includegraphics[width=\linewidth,height=1.3cm]{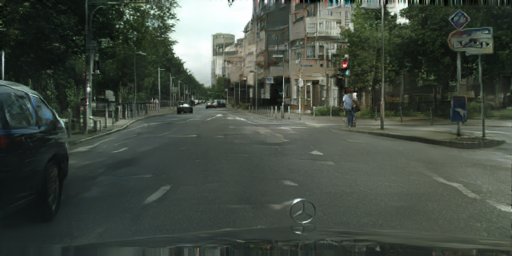}\\
\includegraphics[width=\linewidth,height=1.3cm]{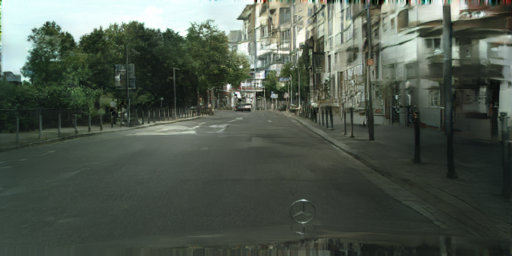}\\
\includegraphics[width=\linewidth,height=1.3cm]{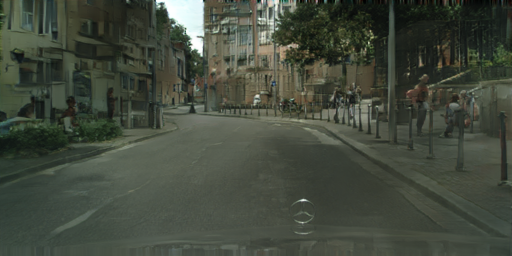}\\
\end{minipage}
} \hspace*{-3mm}
\subfigure[OASIS]{
\begin{minipage}[t]{0.13\linewidth}
\centering
\includegraphics[width=\linewidth,height=1.3cm]{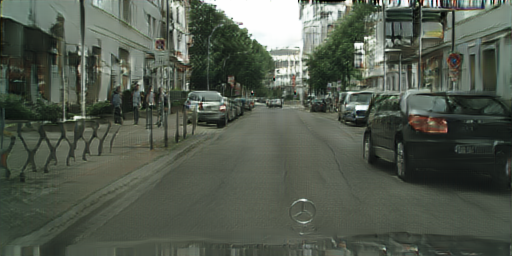}\\
\includegraphics[width=\linewidth,height=1.3cm]{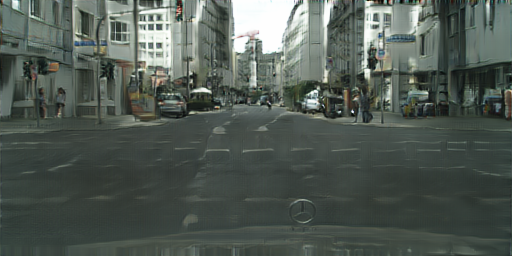}\\
\includegraphics[width=\linewidth,height=1.3cm]{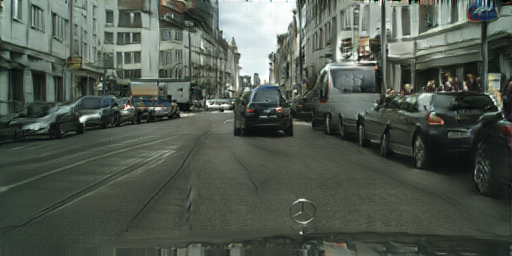}\\
\includegraphics[width=\linewidth,height=1.3cm]{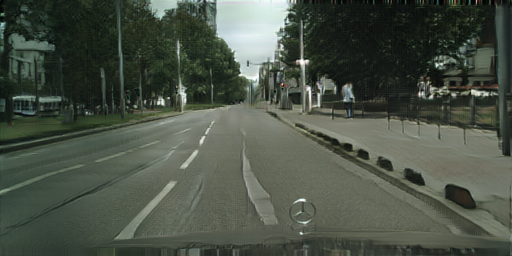}\\
\includegraphics[width=\linewidth,height=1.3cm]{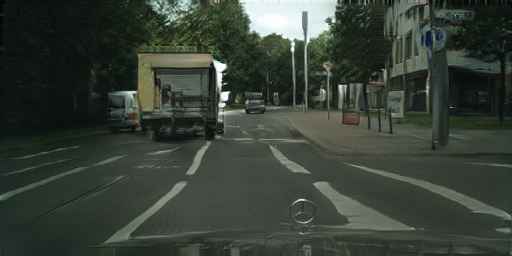}\\
\includegraphics[width=\linewidth,height=1.3cm]{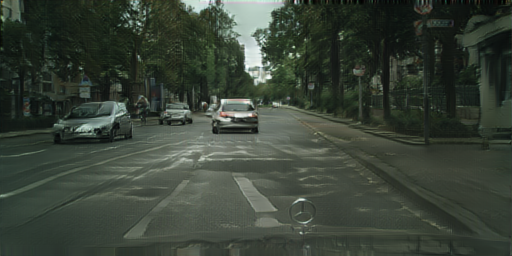}\\
\includegraphics[width=\linewidth,height=1.3cm]{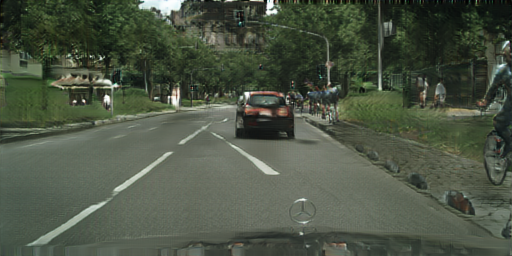}\\
\includegraphics[width=\linewidth,height=1.3cm]{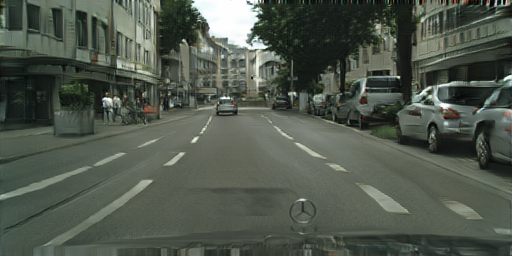}\\
\includegraphics[width=\linewidth,height=1.3cm]{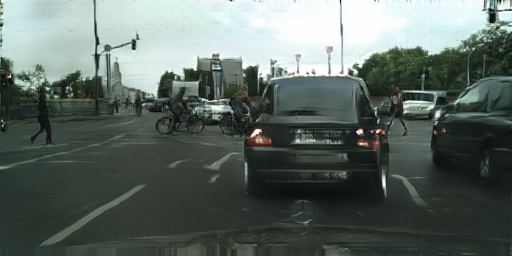}\\
\includegraphics[width=\linewidth,height=1.3cm]{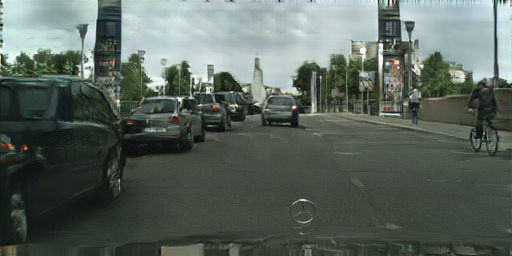}\\
\includegraphics[width=\linewidth,height=1.3cm]{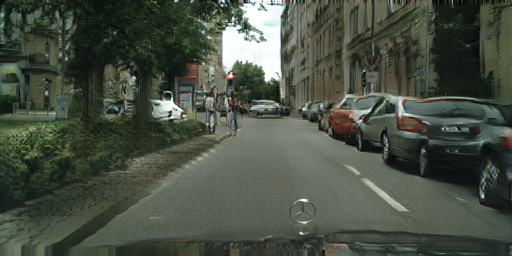}\\
\includegraphics[width=\linewidth,height=1.3cm]{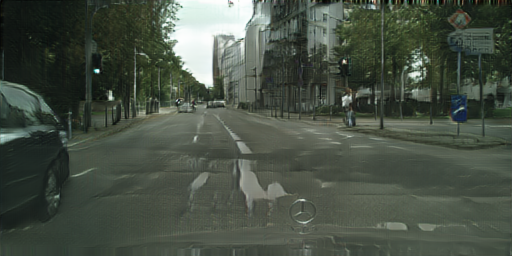}\\
\includegraphics[width=\linewidth,height=1.3cm]{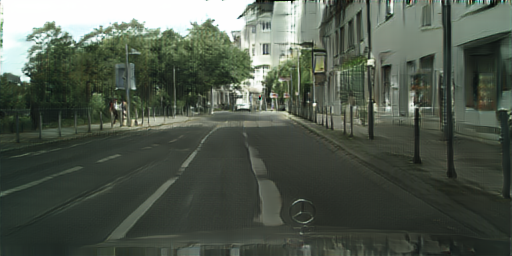}\\
\includegraphics[width=\linewidth,height=1.3cm]{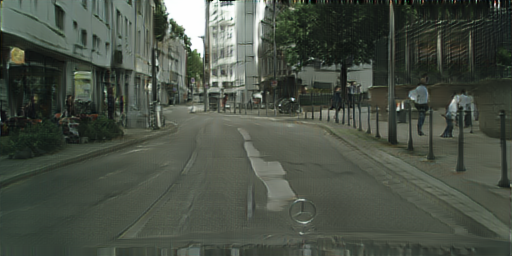}\\
\end{minipage}
} \hspace*{-3mm}
\subfigure[Ours]{
\begin{minipage}[t]{0.13\linewidth}
\centering
\includegraphics[width=\linewidth,height=1.3cm]{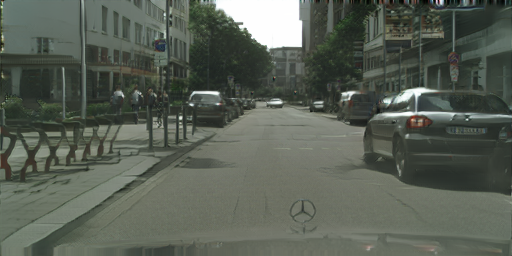}\\
\includegraphics[width=\linewidth,height=1.3cm]{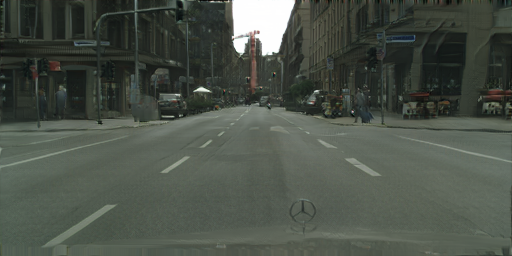}\\
\includegraphics[width=\linewidth,height=1.3cm]{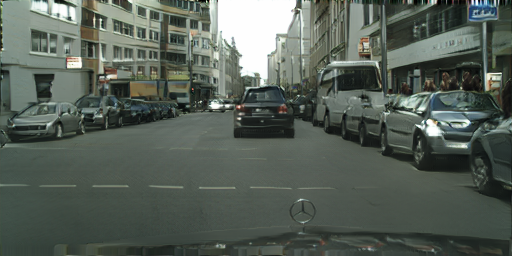}\\
\includegraphics[width=\linewidth,height=1.3cm]{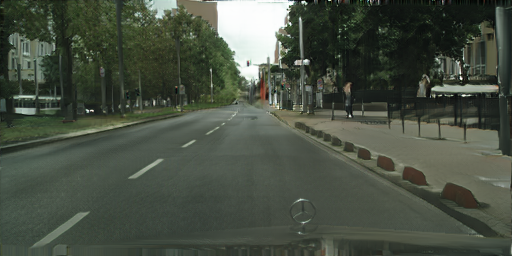}\\
\includegraphics[width=\linewidth,height=1.3cm]{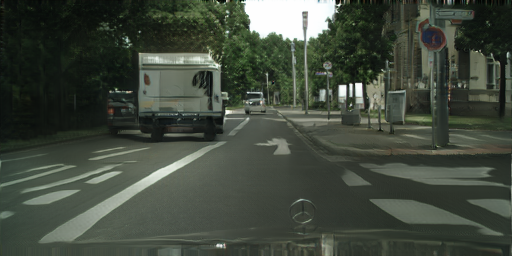}\\
\includegraphics[width=\linewidth,height=1.3cm]{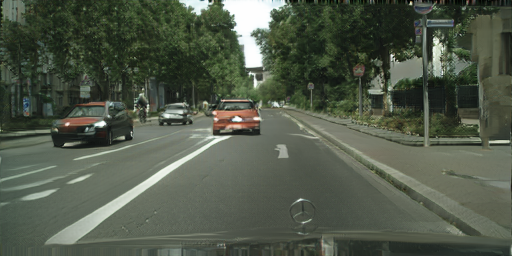}\\
\includegraphics[width=\linewidth,height=1.3cm]{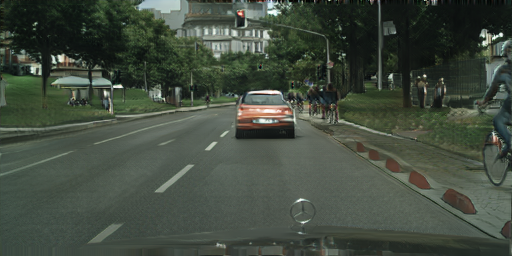}\\
\includegraphics[width=\linewidth,height=1.3cm]{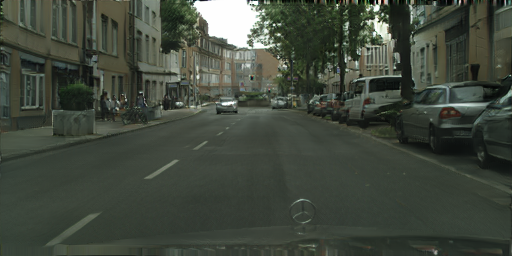}\\
\includegraphics[width=\linewidth,height=1.3cm]{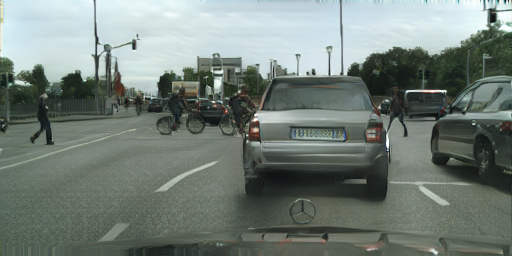}\\
\includegraphics[width=\linewidth,height=1.3cm]{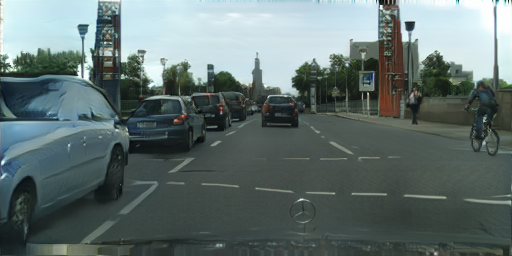}\\
\includegraphics[width=\linewidth,height=1.3cm]{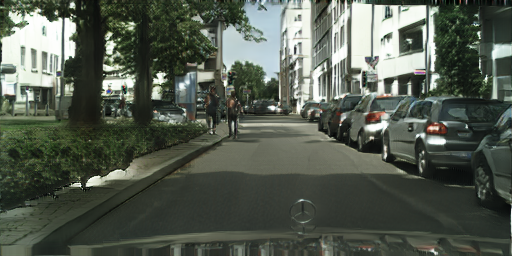}\\
\includegraphics[width=\linewidth,height=1.3cm]{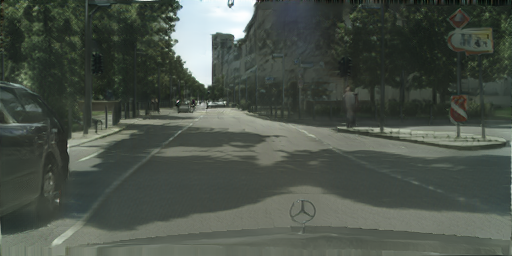}\\
\includegraphics[width=\linewidth,height=1.3cm]{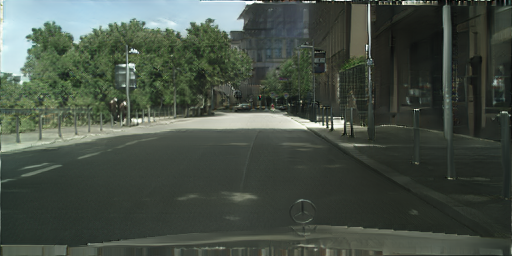}\\
\includegraphics[width=\linewidth,height=1.3cm]{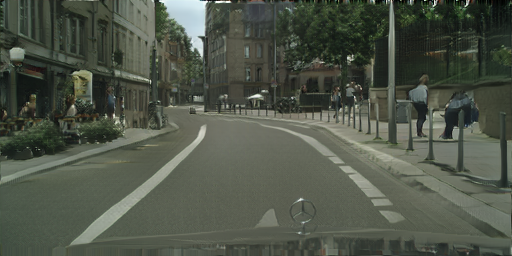}\\
\end{minipage}
} 
\caption{Qualitative results for Cityscapes.}
\label{fig:qvis_cityscapes}
\end{figure*}

\bibliography{egbib}